**Министерство науки и высшего образовагия Российской Федерации**
**Федеральное государственное бюджетное образовательное учреждение**
**высшего образования**
**Московский государственный технический университет**
**имени Н.Э. Баумана**
**(национальный исследовательский университет)**
**(МГТУ им. Баумана)**

ФАКУЛЬТЕТ СПЕЦИАЛЬНОЕ МАШИНОСТРОЕНИЕ

КАФЕДРА СМРОБОТОТЕХНИЧЕСКИЕ СИТЕМЫ И МЕХАТРОНИКА

# НАУЧНО-ИССЛЕДОВАТЕЛЬСКАЯ РАБОТА СТУДЕНТА

НА ТЕМУ:

***Манипулятор***

***для помощи людям с ограниченным возможностями***

Студент **Хуан Бинкунь**

Руководитель ВКР **Котов Е. А.**

Руководитель Кафедры **Ющенко А. С.**

*2024 г*

# РЕФЕРАТ

Расчетно-пояснительная записка XX страницы, XX рисунков, XX таблиц, XX источников.

Выпускная квалификационная работа выполнена по теме: Робот-помощник для людей с ограниченными возможностями.

Цель выпускной квалификационной работы является разработка манипулятора и его системы управления для людей с ограниченными возможностями, который может представить научный и практический интерес.

В данной расчетно-пояснительной записке приведены:

1. Проектирование и конструирование манипулятора
2. Определение структуры системы управления
3. Энергетический расчет манипулятора
4. Кинематический расчет манипулятора
5. Моделирование задач по кинематике и динамике
6. Исследование модели привода конечного исполнителя
7. Разработка системы технического зрения
8. Разработка программного обеспечения в ROS
9. Изготовление макета робота

Ключевые слова: манипулятор, система управления, моделирование, программное обеспечение в ROS, реальный макет.

# СОДЕРЖАНИЕ

# ВВЕДЕНИЕ

Выбор темы данной выпускной квалификационной работы обусловлен значительной актуальностью разработки роботизированных систем, призванных помогать людям с ограниченными возможностями. Развитие технологий в области робототехники и автоматизации открывает новые перспективы для создания устройств, способных значительно улучшить качество жизни этой категории людей. В этом контексте, создание манипулятора с системой управления, адаптированной под нужды людей с ограниченными физическими возможностями, представляет собой значимую научную и практическую задачу.

Проблема, которую предстоит решить в рамках данной работы, заключается в разработке и создании манипулятора с четырьмя степенями свободы, адаптированного для работы в реальных условиях. Решение этой проблемы требует комплексного подхода, включающего проектирование механической структуры манипулятора, разработку системы управления, а также интеграцию с системами технического зрения и программным обеспечением на базе ROS (Robot Operating System).

При выполнении дипломного проекта использовать следующий программы: Microsoft Word, Компас 3D, SolidWorks, Matlab, VMware Fusion.

Объектом исследования является робот-манипулятор с интегрированной системой управления. Методы исследования включают использование программного обеспечения для 3D моделирования, анализа динамики и кинематики, а также разработки на платформе ROS для реализации функций управления и взаимодействия с пользователем.

# 1. Обзор проекта

## 1.1 Актуальность, объект и предметы исследования

Актуальность данной выпускной квалификационной работы заключается в необходимости проектирования, разработки и внедрения манипулятора для помощи людям с ограниченным возможностям и его системы управления, включая двигатели, машинное зрение и планирование движения. В результате будет создан управляемый макет на основе исследования.

В первую очередь, данный манипулятор для кормления был разработаны для улучшения качества жизни и независимости. Они особенно полезны для людей, которые испытывают трудности в повседневной жизни из-за возраста, болезни или инвалидности. Например, люди, страдающие атрофией мышц, последствиями инсульта или другими невромышечными заболеваниями, могут обнаружить, что им трудно использовать обычные столовые приборы.

Используя манипулятор для кормления, эти пользователи могут есть более независимо, не так сильно полагаясь на сотрудников по уходу или членов семьи. Это не только повышает их самооценку и уверенность в себе, но и снижает нагрузку на ухаживающих. Кроме того, эти роботы обычно оснащены передовыми технологиями, такими как системы интеллектуального обнаружения и регулируемые механические руки, которые обеспечивают безопасную доставку пищи, уменьшая трудности и риски при еде.

## 1.2 Анализ существующих манипуляторов для кормления

На данный момент примеров реального внедрения в производство манипуляторов для кормления довольно мало, но анализ этих существующих примеров позволит оценить текущее состояние данной области и определить основные характеристики, преимущества и недостатки этих роботов.

**Obi** - это американское революционное устройство адаптивного питания для людей с ограниченной силой и подвижностью верхних конечностей. Этот робот позволяет пользователям контролировать, что и когда они едят, благодаря настраиваемым переключателям доступности. Obi повышает независимость, социальное взаимодействие, удовольствие от еды и общее благополучие. Устройство включает в себя множество функций, таких как многоразовые ложки и тарелки из безопасного материала, вес, сопоставимый с ноутбуком, перезаряжаемую батарею с продолжительностью работы 3-4 часа, а также функции обнаружения столкновений и водонепроницаемости.

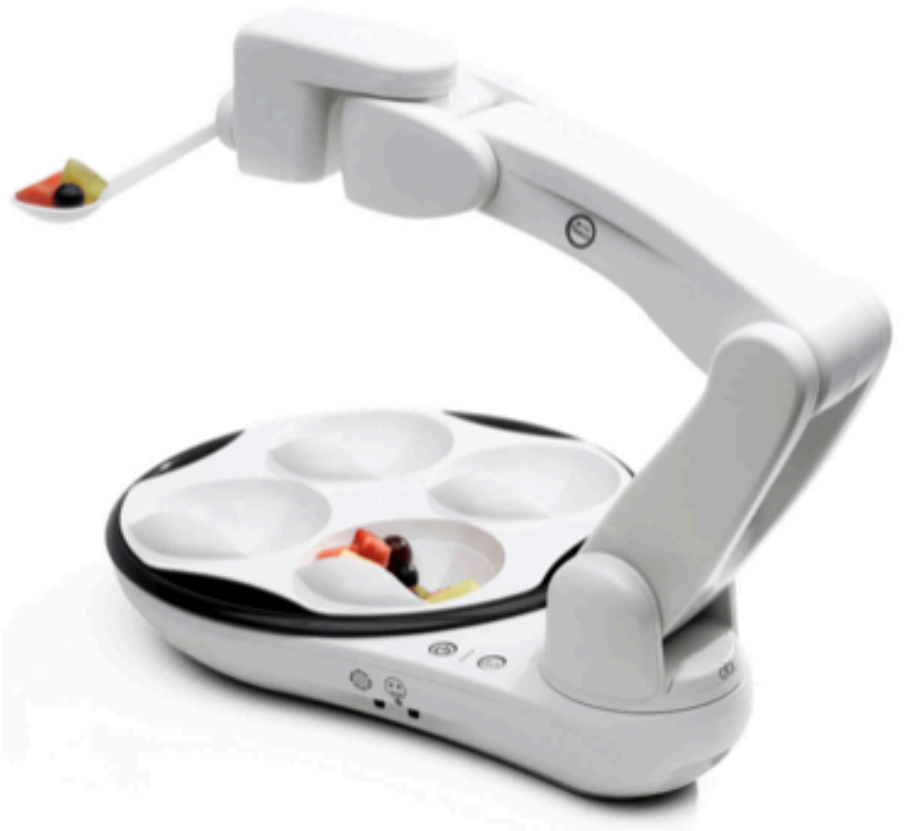

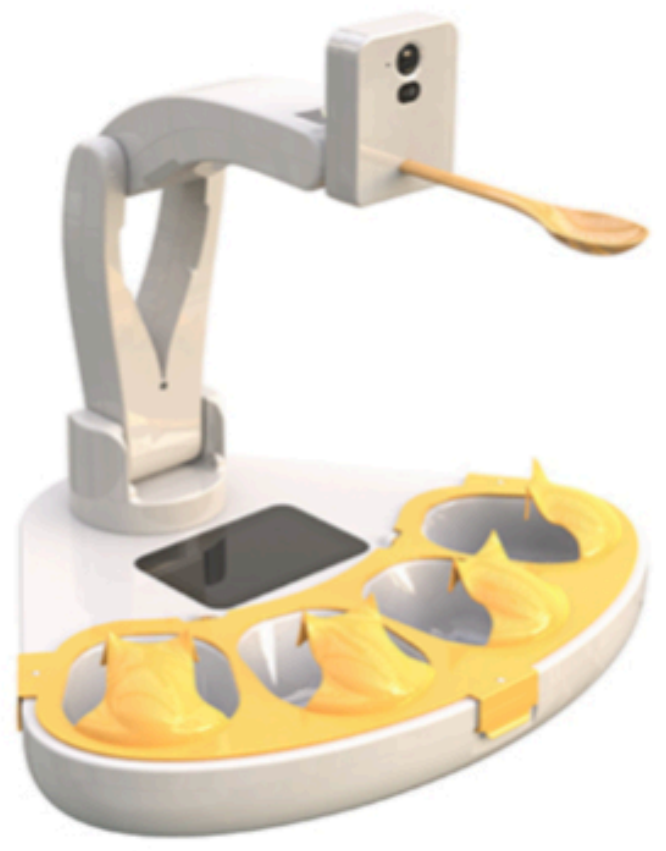

Рис. 1 - Роботы для кормления Obi и Zuowei

**Zuowei** - Китайский робот для кормления, который может автоматически захватывать пищу и подносить ее к рту пользователя, а ложка робота оснащена датчиками и отступает, чтобы избежать травмирования пользователя. Такой робот-помощник в питании также способен захватывать мягкую пищу, такую как тофу, и мелкие продукты, например, рисовые зерна. Использование этого робота для помощи в питании может эффективно помогать пациентам с параличом и пожилым людям с ограниченной подвижностью самостоятельно питаться. Робот для кормления подходит для большинства людей с заболеваниями, влияющими на функцию верхних конечностей. Пользователи должны иметь возможность безопасно и успешно управлять роботом для помощи в питании, а также обладать способностью жевать и глотать без посторонней помощи.

Не трудно заметить, что механические конструкции и способы механической передачи двух манипуляторов для кормления очень похожи, все элементы используют последовательное соединение, а на конечных исполнительных устройствах используются устройства, похожие на ложки. Obi использует тактильные датчики, в то время как Zuowei использует визуальные датчики для позиционирования. Благодаря разумному использованию механизма замедления, робот Obi может быть обучен человеком и решать задачи обратной кинематики. Однако цена этих двух роботов очень высока, около 5000 долларов.

# 2. Комплексное проектирование и анализ

## 2.1 Постановка задачи

В выпускной квалификационной работе спроектированы манипулятор по имени **RoboBK** для помощи людям с ограниченным возможностям, который имеет главную функцию кормления, и его автоматизированная система управления. Выполнение данного проекта может значительно помочь студенту укрепить и углубить знания, полученные ранее в течении бакалавра, включая, но не ограничиваясь: Проектирование и конструирование роботов, Электрические исполнительные системы РТК, Управление в технических системах, Управление роботами. Выпускная квалификационная работа, который может представить научный и практический, также поможет студенту подготовится к профессиональной деятельности на предприятиях.

Для достижения цели в работе поставлены следующие задачи:

- Проектирование и конструирование манипулятора
- Определение структуры системы управления
- Энергетический расчет манипулятора
- Кинематический расчет манипулятора
- Моделирование задач по кинематике и динамике
- Исследование модели привода конечного исполнителя
- Разработка системы технического зрения
- Разработка программного обеспечения в ROS
- Изготовление макета робота

## 2.2 Первичное проектирование манипулятора

Предлагаемый робот RoboBK для помощи людям с ограниченным возможностями должен обладать следующими функциями:

- кормление
- автоматическое ослеживание лица в системе видеонаблюдения
- разные программируемые задачи

В зависимости от комплектации, робот будет снабжаться разным количеством приводов и иметь разное количество степеней свободы.

Проектируемый робот может работать в автономном режиме (выполнять операции, заранее заложенные в его УУ), ручном режиме (выполнять команды оператора) или в полуавтоматическом режиме. Каждый из данных режимов должен быть снабжен своей программой в виде программного кода. Также для полуавтоматического режима и ручного режима необходимо наличие персонального компьютера/ноутбука на базе операционной системы Windows, Linux, MacOS на базе операционной системы ROS и Android. При этом он будет получать необходимые команды как по кабелю-USB

Структуру проектируемого робота RoboBK предлагается организовать по нижной схеме, представленной на рисунке 2.

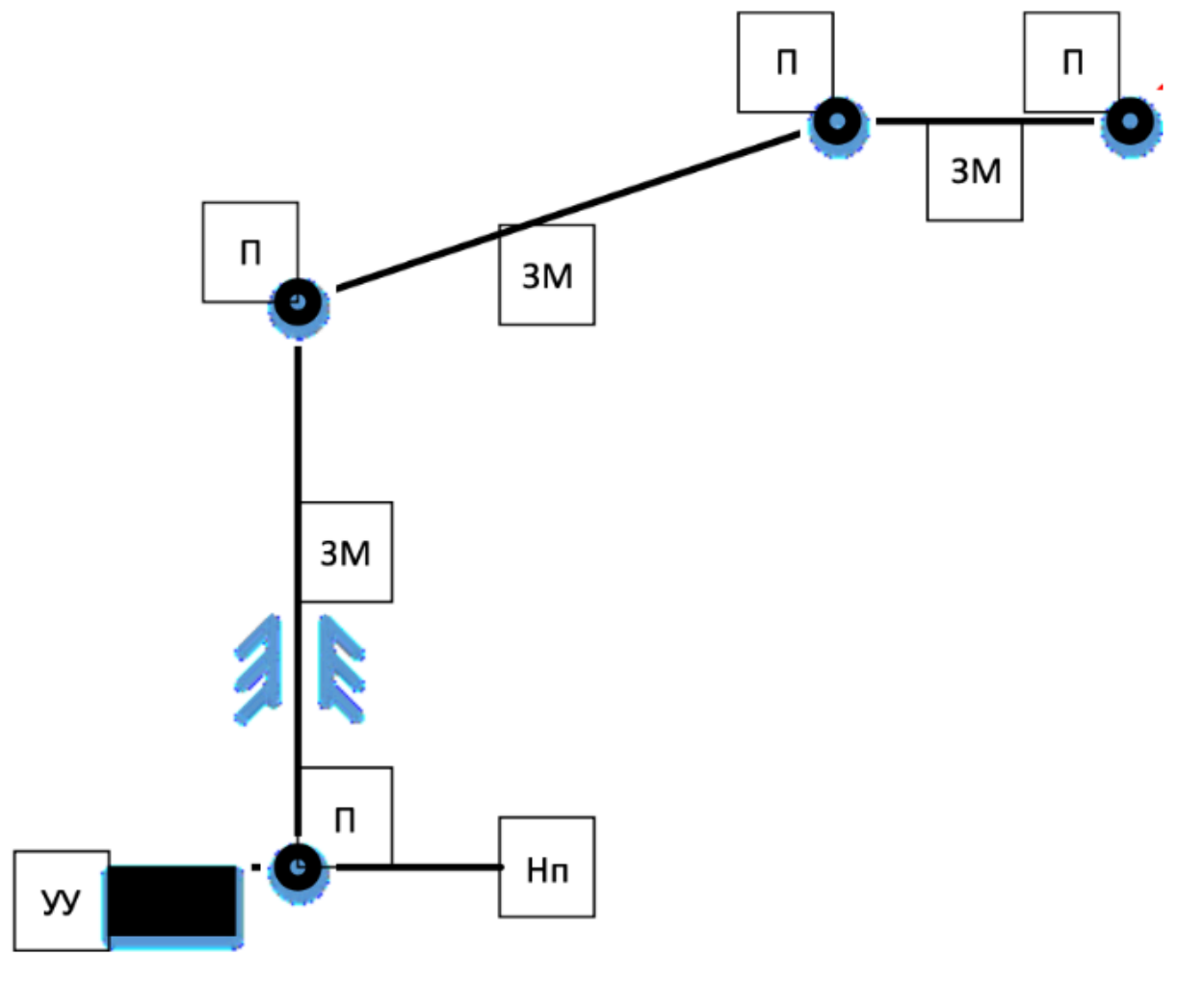

Рис. 2 - Структурная схема манипулятора

Где:

УУ – устройства управления;

Нп – неподвижная платформа;

П – привод;

ЗМ – звено манипулятора;

Сам манипулятор робота крепится на неподвижной платформе (НП). Непосредственно в платформе установлено устройство управления (УУ).

В нашем проекте RoboBK принимается механическая последовательная структура роботизированной руки соединяет несколько жестких элементов (звеньев) с помощью суставов (шарниров) (рис.3).

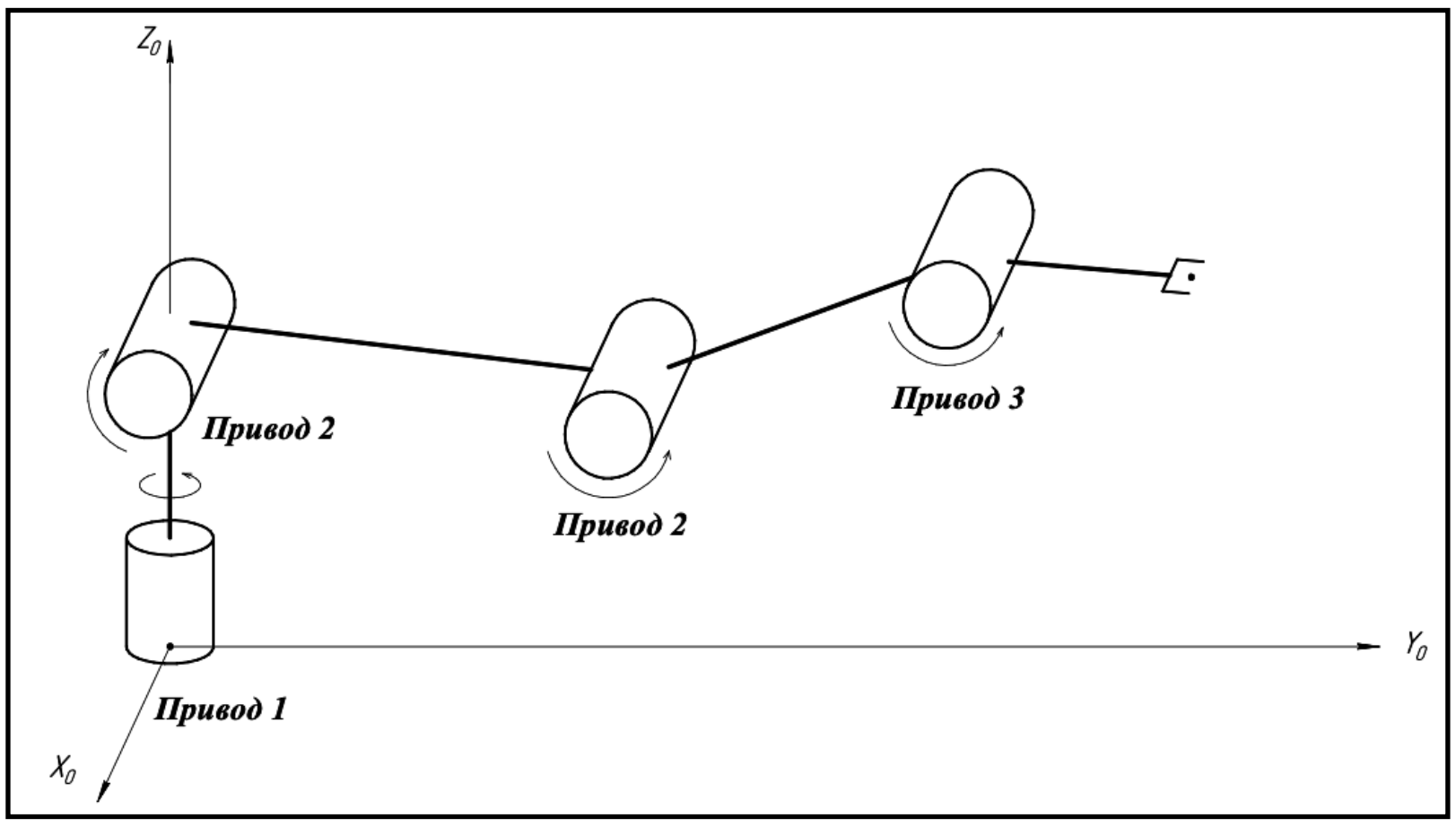

Рис. 3 - Упрощенная схема структуры манипулятора

В результате, как указано в Рис. 3, для приводов 1, 2 и 3 принимаются ШД, а для провода 3 принимется мотор-редуктор постоянного тока.

## 2.2 Разработка функциональной схемы управления роботом

Промышленный роботом является автоматически управляемый, перепрограммируемый, многоцелевой манипулятор, программируемый по трем и более осям. Он может быть либо зафиксирован в заданном месте, либо может иметь возможность передвижения для выполнения промышленных задач по автоматизации.

Рассмотрим основные элементы функциональной схемы системы управления промышленным роботом (манипулятор)(Рис. 4) :

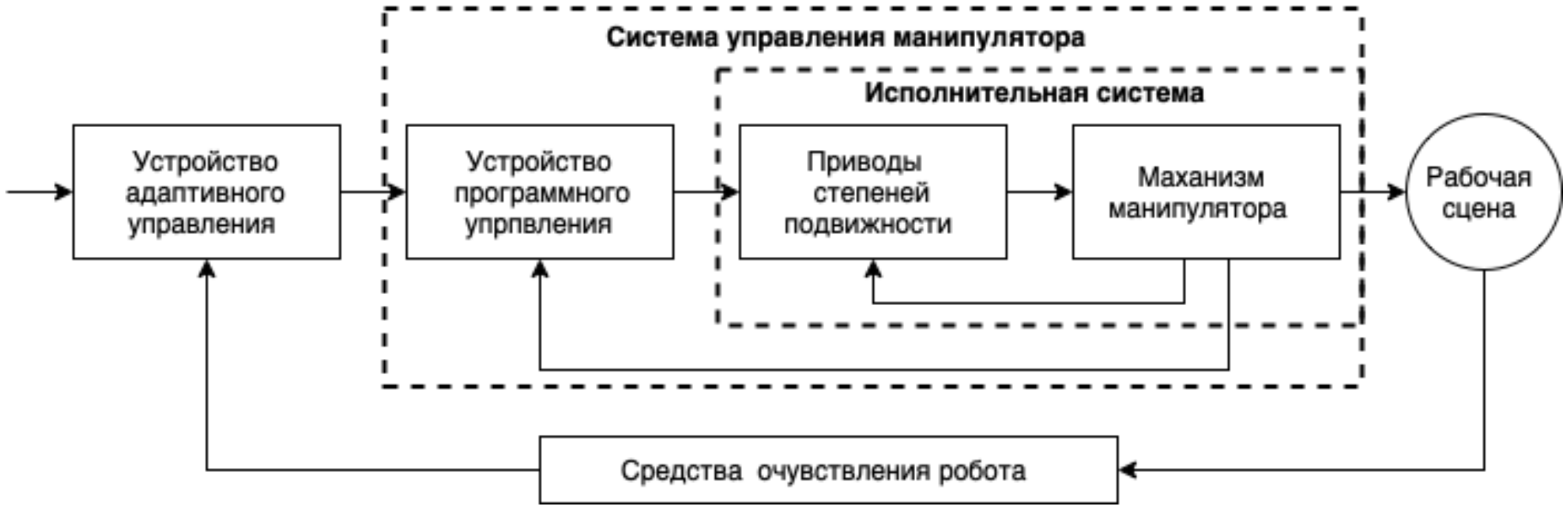

Рис. 4 - Функциональной схемы системы управления

## 2.3 Разработка структурной схемы системы управления

Система управления манипулятора RoboBK состоит из трех уровней.

**Верхний уровень:** Такой уровень состоит из управляющего компьютера. Он решает задачу со изображением зрения, которое получается с камеры и управляет приводами через микроконтроллера при процессе требуемого движения.

**Средний уровень:** Такой уровень состоит из микроконтроллера. Он выполняет связь между приводами и управляющим компьютером.

**Нижний уровень:** Такой уровень состоит из двух модулей - модуль приводов и модуль зрения. Модуль приводом состоит из 3 шаговых двигателя, мотор-редуктор постоянного тока, датчик обратной связи и их драйверов двигателей. Модуль зрения состоит из самой камеры и монитор, на котором изображается информация.

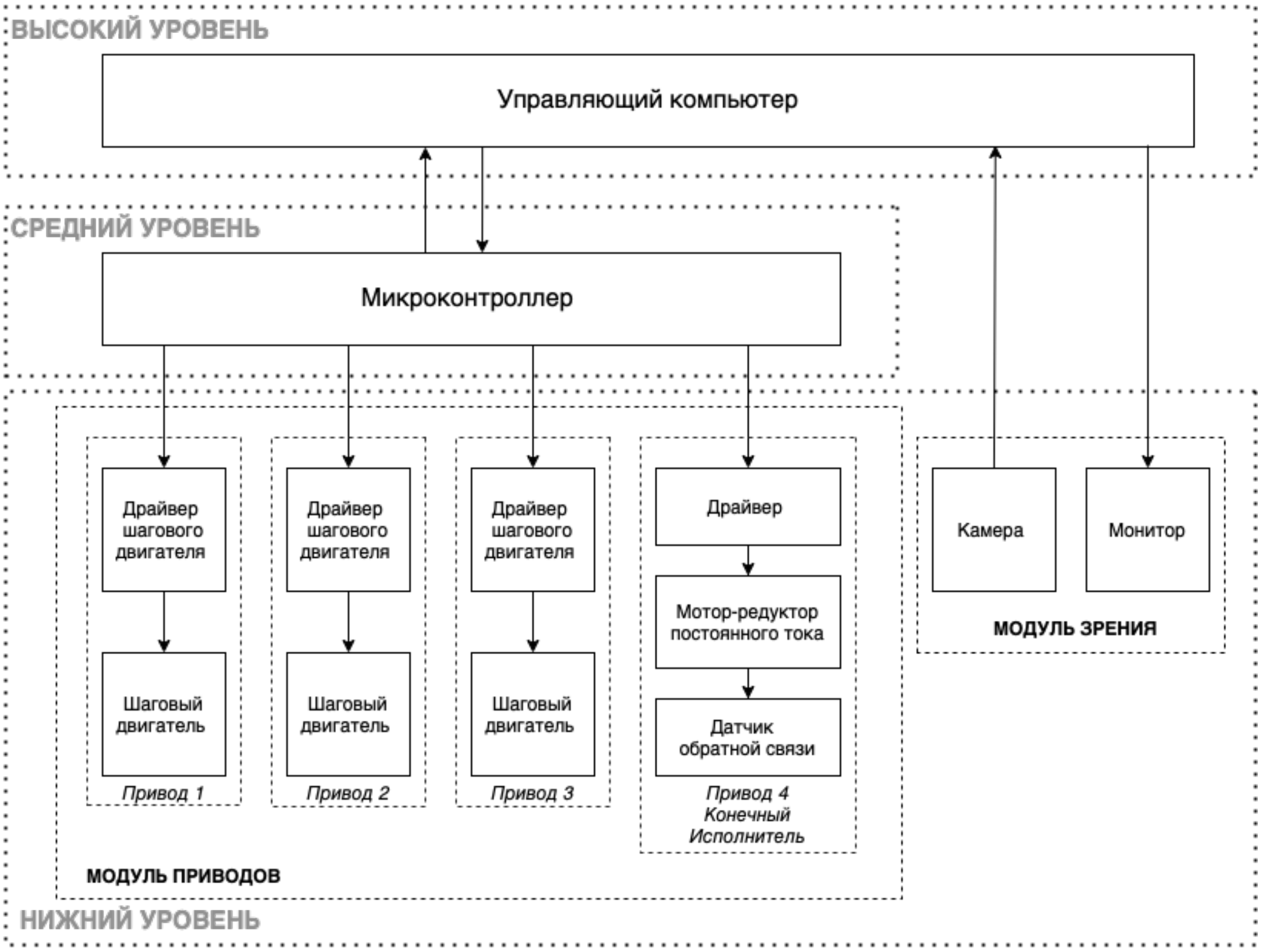

Рис. 5 - Структурная схема системы управления

## 2.4 Выбор и анализ компонентов

### 2.4.1 Анализ и выбор электромеханических устройств

Механическая структура роботизированной руки соединяет несколько жестких элементов (звеньев) с помощью суставов (шарниров). Компоненты манипулятора включают в себя руку, обеспечивающую подвижность, запястье, обеспечивающее ловкость, и конечный исполнительный механизм, необходимый для выполнения задачи.

Для звеньев и шарниров, шаговые двигатели (ШД) представляют собой электромеханические устройства, задачей которых является преобразование электрических импульсов в перемещение вала двигателя на определенный угол. Достоинствами ШД по сравнению с простыми являются:

- Возможность применения в качестве привода без редуктора.
- Возможность управлять ШД по разогнутому циклу (без датчиков обратной связи и регуляторов).
- Высокая точность позиционирования и повторяемости.

Хотя шаговые двигатели тоже именно недостатки, например, пропуска шагов под действием нагрузки и высоким ускорением, не большой крутящего момента относительно других видов двигателей, в нашем случае, когда не требуется большая нагрузка и ускорение, эти недостатки можно избежать.

Выбраны шаговые двигателя для жестких элементов и соединенных звеньями модели 42BYGH34, 42BYGH39, 42BYGH47.

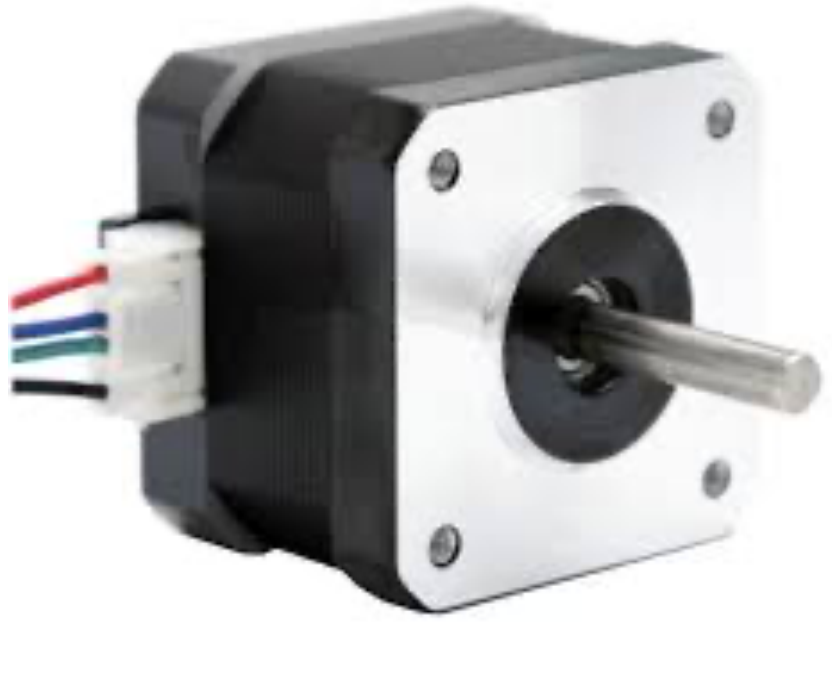

Рис. 6 - Размеры ШД 42BYGH34

Таблица 1 – Технические характеристики ШД 42BYGH34/39/47

| | |
|---|---|
| Число фаза | 2 |
| Рабочее напряжение, В | 24 |
| Крутящий момент, Н*м | 0.28/0.4/0.55 |
| Номинальный ток, А | 1.3 - 1.5 |
| Полный шаг, ° | 1.8 |
| Полных шагов на оборот | 200 |
| Диаметр вала, мм | 5 |
| Длина вала, мм | 23 |

Для управления ШД используют драйверы A4988. Драйвер A4988 позволяет контролировать ток, подаваемый на ШД. Это важно, поскольку правильная настройка тока позволяет двигателю работать эффективно и предотвращает его перегрев или повреждение. Драйвер A4988 имеет встроенную возможность настройки тока, что облегчает его использование с различными типами шаговых двигателей. A4988 поддерживает технику микрошагового управления. Микрошаговый режим позволяет шаговому двигателю выполнять более плавные и точные движения, разбивая каждый полный шаг на более мелкие инкременты. A4988 также включает в себя некоторые защитные функции, которые помогают предотвратить повреждение шагового двигателя и самого драйвера. В работе пользуется микрошаговый режим 1/16.

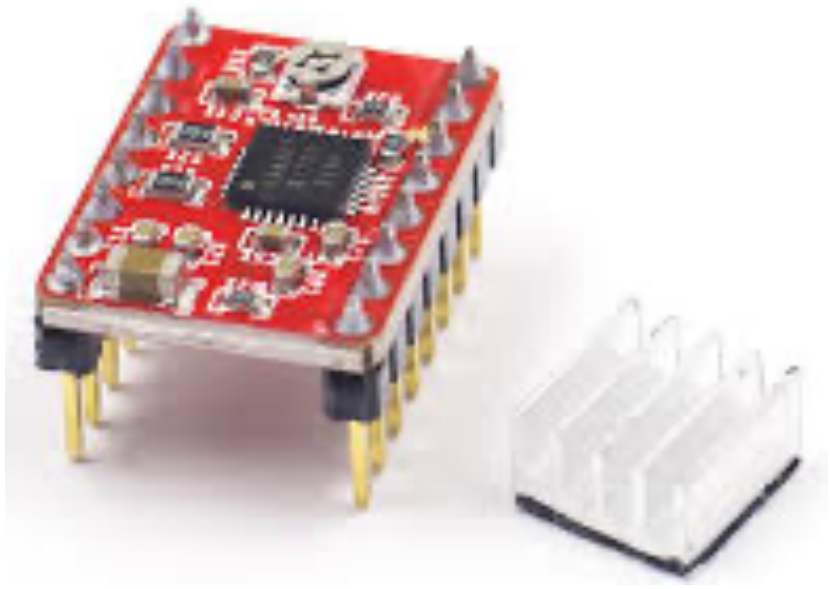

Рис. 7 - Драйвер A4988

Для подключения ШД к Arduino будет использоваться плата расширения. Он использует все разъемы плата Arduino UNO и обеспечивает удобный способ подключения к ШД и драйверами. Для данной работы выбрана CNC Shield V3.

Плата расширения CNC Shield V3 предоставляет ряд преимуществ при использовании в системах ЧПУ (числового программного управления). Она обеспечивает простое подключение и управление шаговыми двигателями, позволяет использовать драйверы A4988 или DRV8825, поддерживает микрошаговое управление для плавного движения, имеет разъемы для подключения концевых выключателей и датчиков.

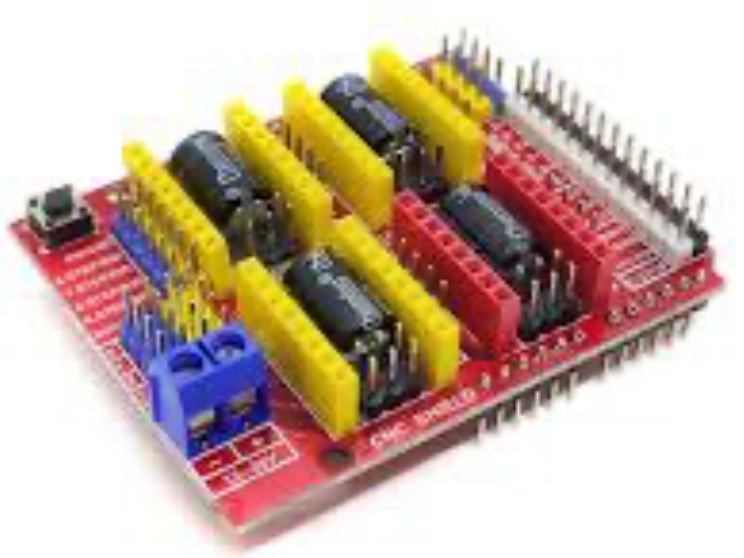

Рис. 8 - Плата расширения CNC Shield V3

Для конечного исполнительного механизма, первое требование выбора двигателя заключается в том, чтобы при относительно небольшом весе он мог обеспечивать достаточно большой крутящий момент. Второе требование связано с высокой точностью. Таким образом, двигатель постоянного тока (ДПТ) с редуктором (мотор-редуктор) обладает явными преимуществами в удовлетворении требований к небольшому весу, высокому крутящему моменту, плавному движению и низкой вибрации. В этом случае, используется замкнутой системы с датчиков обратной связи вместо разомкнутой системы. Энкодер позволяет осуществлять точное управление положением и скоростью с обратной связью, что необходимо для поддержания точной и плавной работы через алгоритмы управления, как PID-регулирование.

Выбран мотор-редуктора MG310 с энкодером HAL (рис. 9), его драйвер модели L298N (рис. 10).

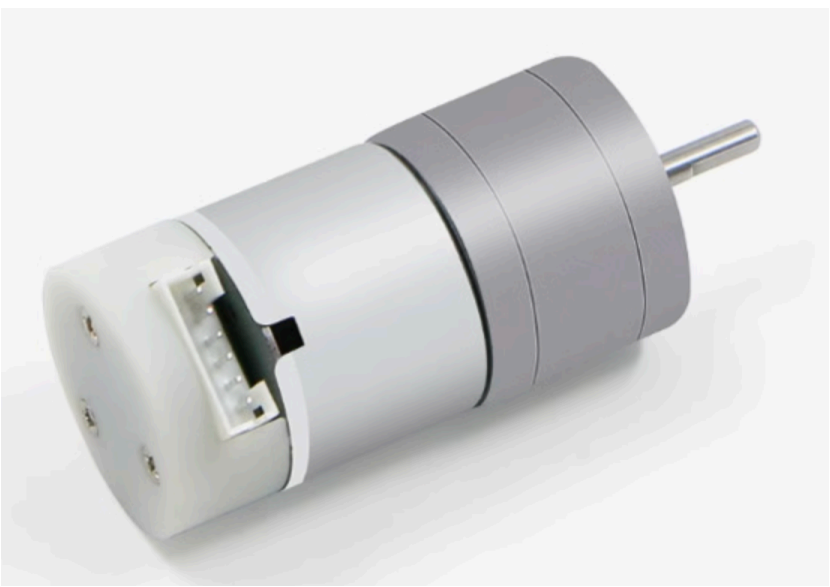

Рис. 9 - Мотор-редуктор с энкодером MG310

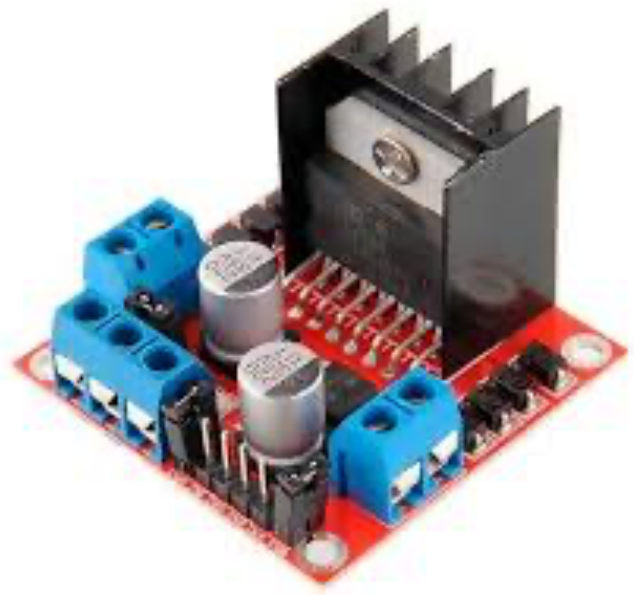

Рис. 10 - Драйвер модели L298N

Таблица 2 – Технические характеристики MG310

| | |
|---|---|
| Передаточное число | 1:20 |
| Номинальное напряжение, В | 7.4 |
| Номинальная скорость, об/мин | 500±13% |
| Номинальный ток, мА | <200 |
| Холостой ход, об/мин | 400±13% |
| Ток холостого хода, мА | <500 |
| Номинальный крутящий момент, кгс·см | 0.4 |
| Ток блокировки, мА | <500 |
| Крутящий момент блокировки, кгс·см | 1.5 |
| Масса, кг | 0.04 |

Таблица 4.3 – Технические характеристики L298N

| | |
|---|---|
| Напряжение питания логики, В | 5 |
| Потребляемый ток встроенной логики, мА | 0-36 |
| Номинальный ток, А | 2 |
| Напряжение питания драйвера, В | 5-35 |

Требование к выходному напряжению преобразователя напряжения составляет 24В, так как входное напряжение одного ШД составляет 24 В. А требование к выходному току преобразователя 6.3 А, так как номинальный ток одного ШД составляет 1,5 А и всего 3 ЩД. Мощность преобразователя должна быть 150Вт, так как:

$$P_{need} = N * I * V = 3 * 1.5 * 24 = 108 \text{ Вт}$$

В результате выбран преобразователь напряжения 220/24В-6А-150Вт.

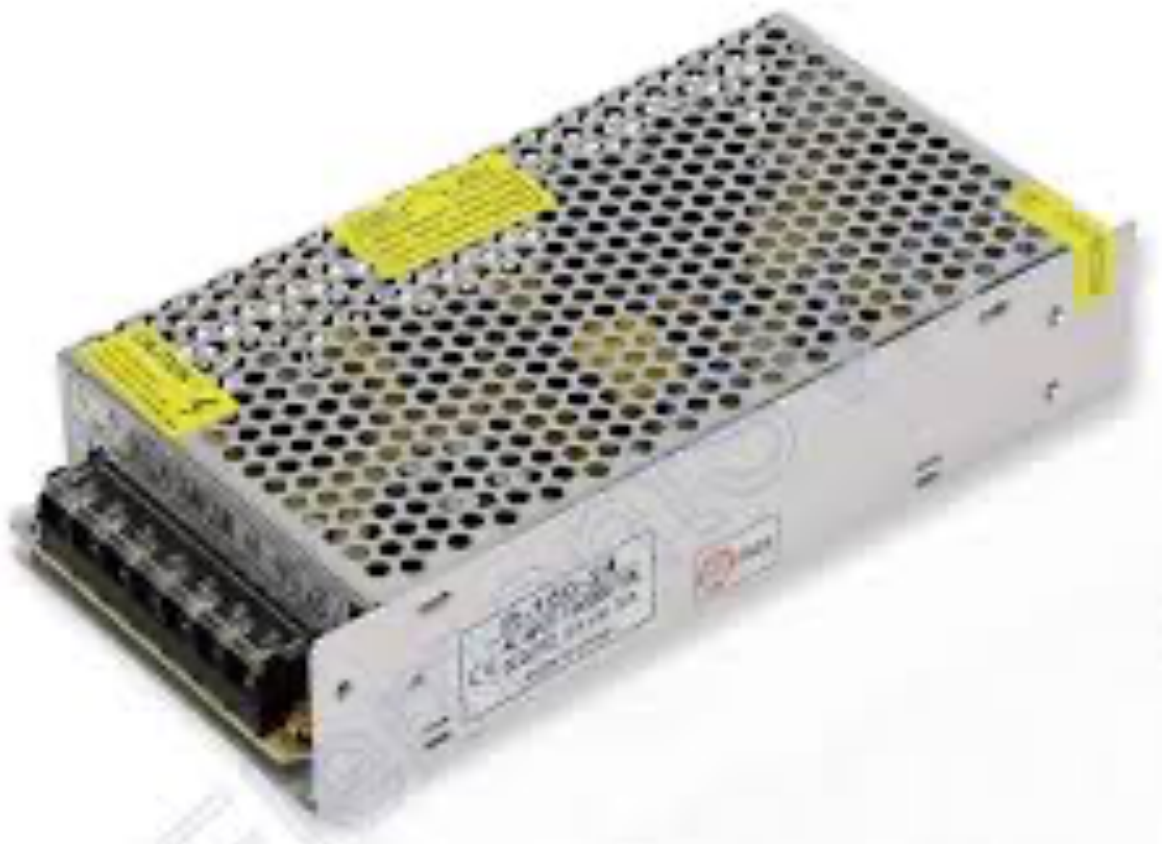

Рис. 11 - Преобразователь напряжения

### 2.4.2 Анализ и выбор устройств для системы управления

**Компонент для высокого уровня системы управления.** В данной работе в качестве управляющего компьютера был выбран компьютер с операционной системой ROS, что обусловлено его широким применением и совместимостью в области робототехники. ROS (Robot Operating System) предоставляет богатый набор инструментов и библиотек, которые значительно упрощают разработку робототехнических проектов, включая интеграцию сенсоров, системные симуляции и реальное время управления.

В качестве аппаратной платформы для контрольного устройства был выбран промышленный ПК, который способен эффективно работать с ROS. После оценки соотношения цены и качества, а также системных требований, был выбран ПК с процессором не ниже Intel i5 или эквивалентным от других производителей, и объемом оперативной памяти не менее 8 ГБ. Кроме того, промышленные ПК способны стабильно функционировать в более широком диапазоне температур, что увеличивает надежность и долговечность системы. При этом он будет получать необходимые команды как по кабелю-USB.

**Компонент для среднего уровня системы управления**. Система управления манипулятора построена на основе платы Arduino uno R3 (Рис.12). Arduino представляет собой аппаратно-программный комплекс открытой архитектуры для построения учебных систем автоматики и робототехники (рис. 3, а). В линейке устройств Arduino в основном применяются микроконтроллеры Atmel AVR ATmega XXX с частотой тактирования до 16 МГц, а также платы с процессором ARM Cortex M с частотой 48 МГц и более. Все устройства программируются через USB без использования программаторов. Микроконтроллеры на плате программируются при помощи языка Arduino (фактически C/C++) и среды разработки Arduino (основана на среде Processing).

**Характеристики Arduino uno R3**

- Микроконтроллер - ATmega328P
- Рабочее напряжение - 5V
- Рекомендуемое входное напряжение - 7-12V
- Предел входного напряжения - 6-20V
- Цифровые входно-выходные контакты - 14 (из которых 6 выходов PWM)
- Цифровые входно-выходные контакты PWM - 6
- Аналоговые входные контакты - 6
- Ток постоянного тока на контакт ввода/вывода - 20 mA
- Ток постоянного тока для контакта 3.3V - 50 mA
- Флеш-память - 32 KB
- Статическая оперативная память - 2 KB
- Clock Speed: Тактовая частота - 16 MHz
- LED_BUILTIN: Встроенный светодиод - 13

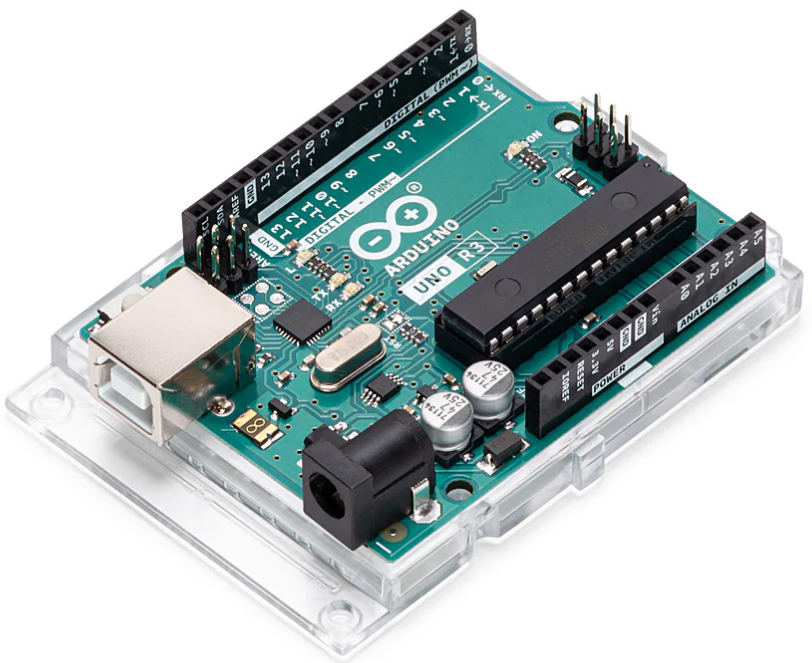

Рис. 13 - Arduino uno R3

### 2.4.3 Анализ и выбор исходного материала 3D-печатных деталей

В качестве исходного материала, из которого изготовлены детали робота был выбран PLA - пластик. Выбор данного материала обусловлен рядом его очевидных достоинств, таких как доступность и дешевизна.

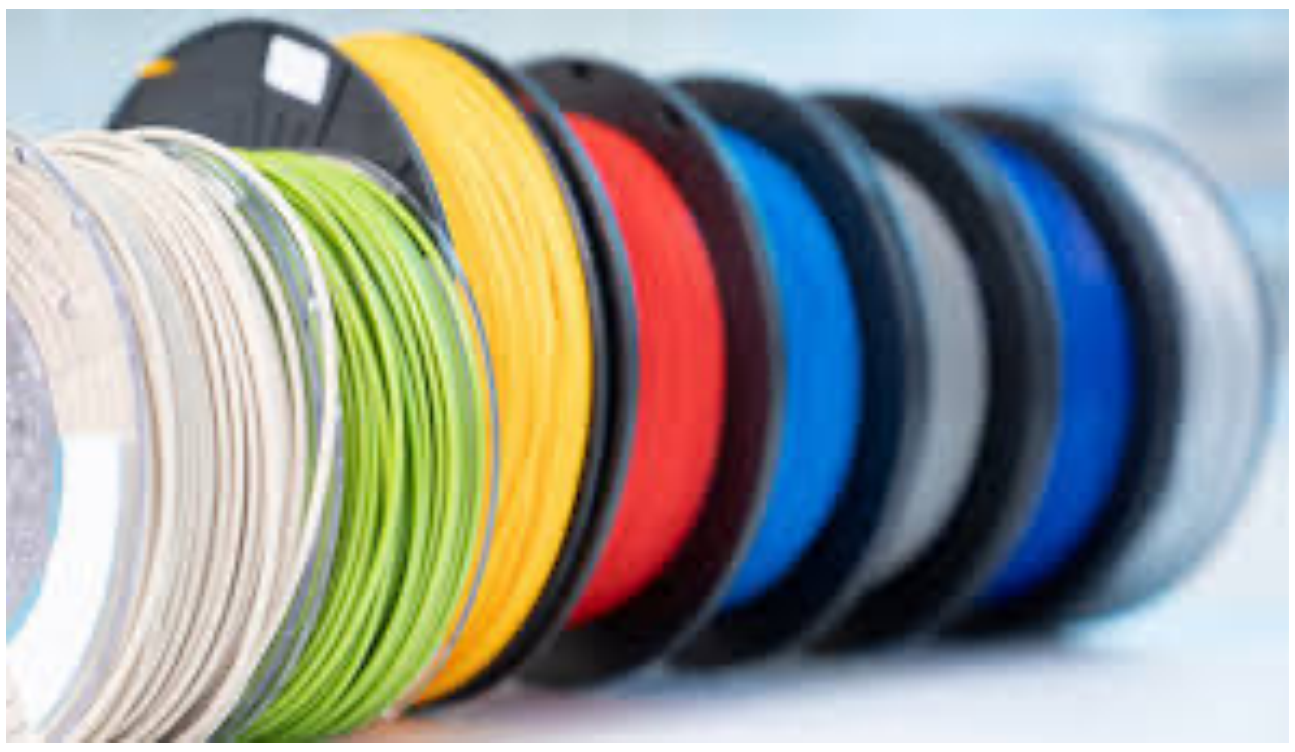

Рис. 14 - PLA

Также следует отметить:

- нетоксичность в нормальных условиях;
- долговечность в отсутствие прямых солнечных лучей и ультрафиолета;
- стойкость к щелочам и моющим средствам;
- влагостойкость;
- маслостойкость;
- кислотостойкость;
- широкий диапазон эксплуатационных температур (от −40 °C до +90 °C);
- плотность 1,02-1,06 г/см$^3$ ;

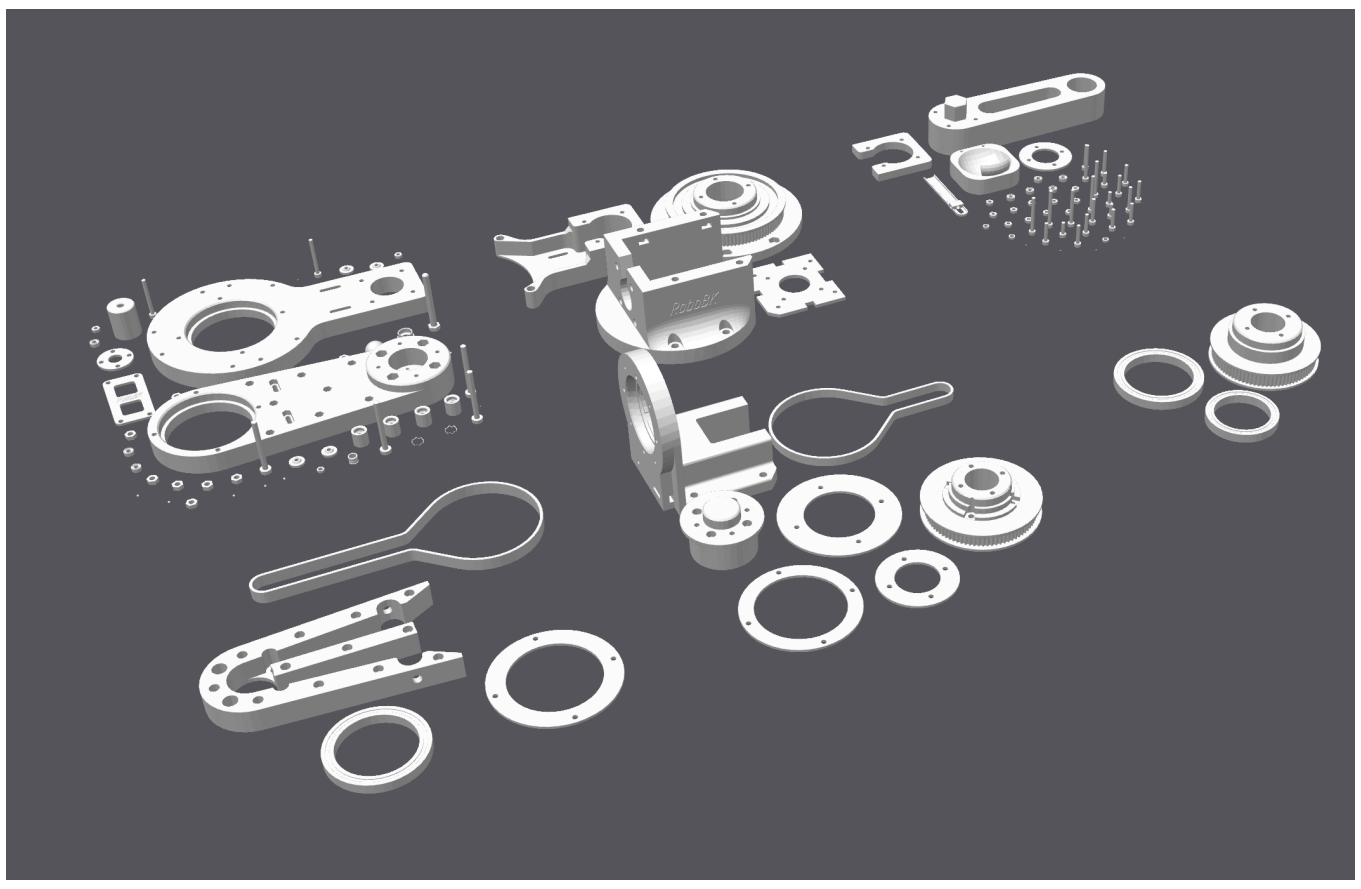

Рис. 15 - Процесс изготовления

## 2.5 Детальное проектирование

RoboBK использует шаговые двигатели для прямого привода зубчатых ремней в соединениях, что также способствует увеличению крутящего момента в соединениях. Все зубчатые ремни являются типа HTD-3M, а все шкивы и натяжные ролики изготовлены методом 3D-печати. Ниже представлена таблица, которая содержит информацию о звеньях и приводах робота.

Таблица 2 – Звена манипулятора и их приводы

| **Звено** | **Название** | **Мотор** | **Удерживающий момент мотора (Нм)** | **Снижение привода ремня** | **Идеальный крутящий момент звена (Нм)** |
|---|---|---|---|---|---|
| 0 | База | - | - | - | - |
| 1 | Колонка | 42BYGH40 | 0.40 | 6 | 2.40 |
| 2 | Плечо | 42BYGH34 | 0.28 | 5 | 1.40 |
| 3 | Локоть | 42BYGH47 | 0.55 | 5 | 2.75 |
| 4 | Конечный исполнитель | MG310 | 0.04 | - | 0.04 |

Результат детального проектирования указано в рис. 16.

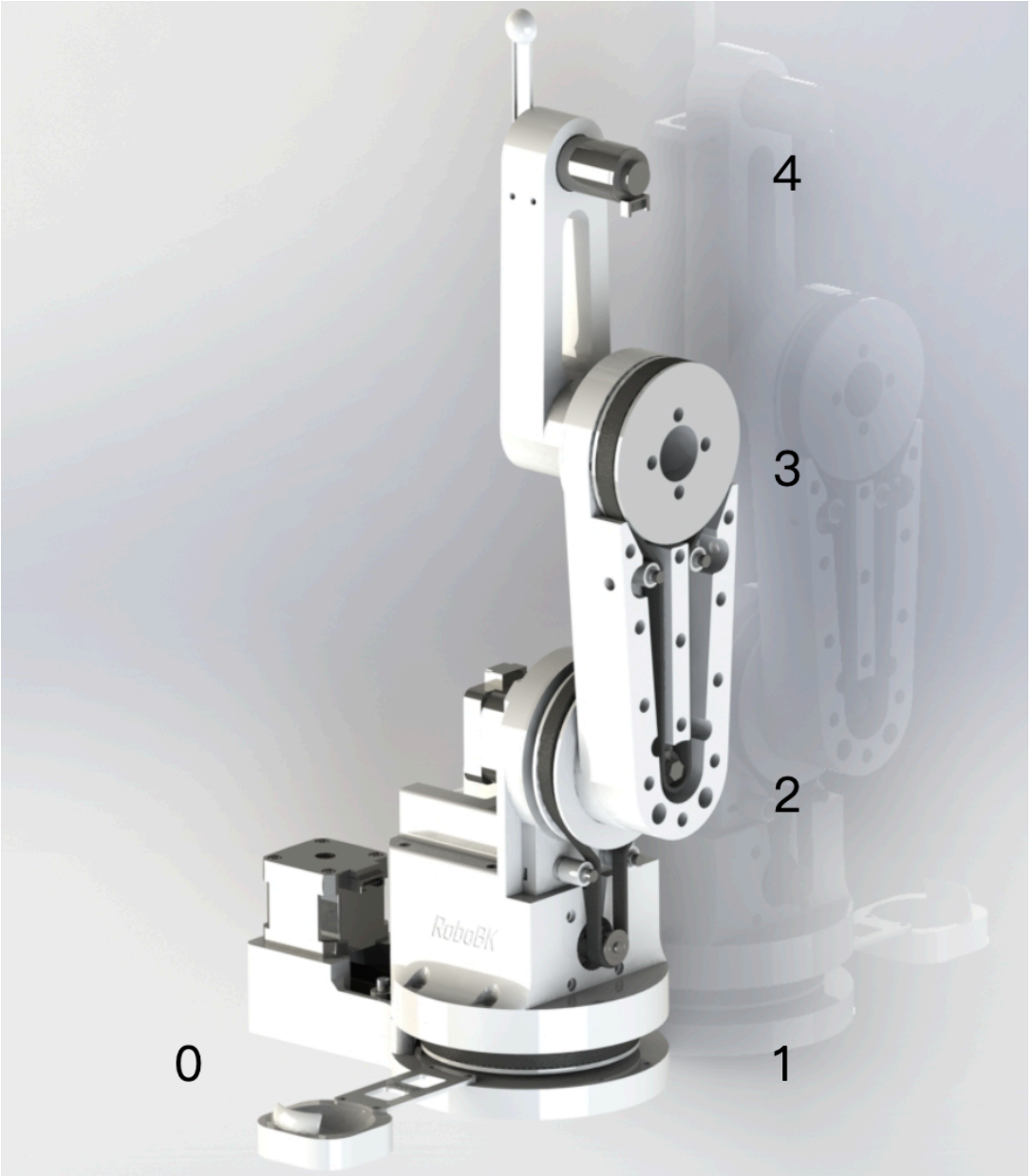

Рис. 16 - 3D-рендеринг манипулятора

В нижнем сочленении шаговый двигатель 42BYGH40 через ремень приводит в движение шкив, закрепленный в нижней части сочленения плеча

(сустав1), что обеспечивает дополнительное снижение скорости. Радиальная нагрузка на ремень, приводящий шкив, воспринимается радиальным подшипником 61808ZZ, закрепленным на нижнем сочленении. В суставах роботизированной руки зубчатые ремни обычно натягиваются с помощью 3D-печатных идлеров, вращающихся на подшипниках 683ZZ, а радиальные подшипники используются для восприятия нагрузки в местах суставов.

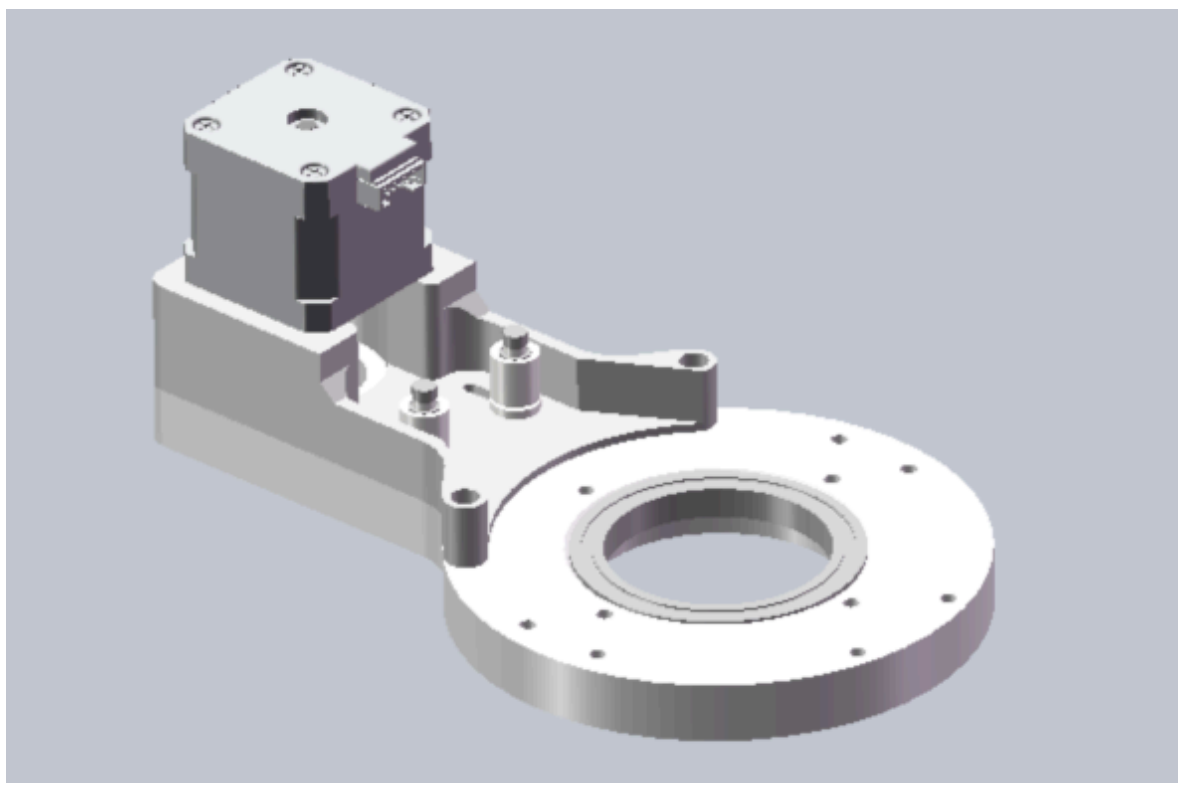

Рис. 17 - Сочленение

Плечевое соединение содержит два шаговых двигателя 42BYGH34 и 42BYGH47, каждый из которых закреплен на своей скобе, один расположен над другим. Нижний шаговый двигатель (42BYGH47) приводит в движение ремень, который в свою очередь вращает шкив, закрепленный на соединении верхней части руки (сустав 2), обеспечивая дополнительное снижение скорости. Через ось этого ведомого шкива проходит вал, изготовленный методом 3D-печати, который соединяет верхний шаговый двигатель (42BYGH34) с приводным шкивом, закрепленным на соединении верхней части руки.

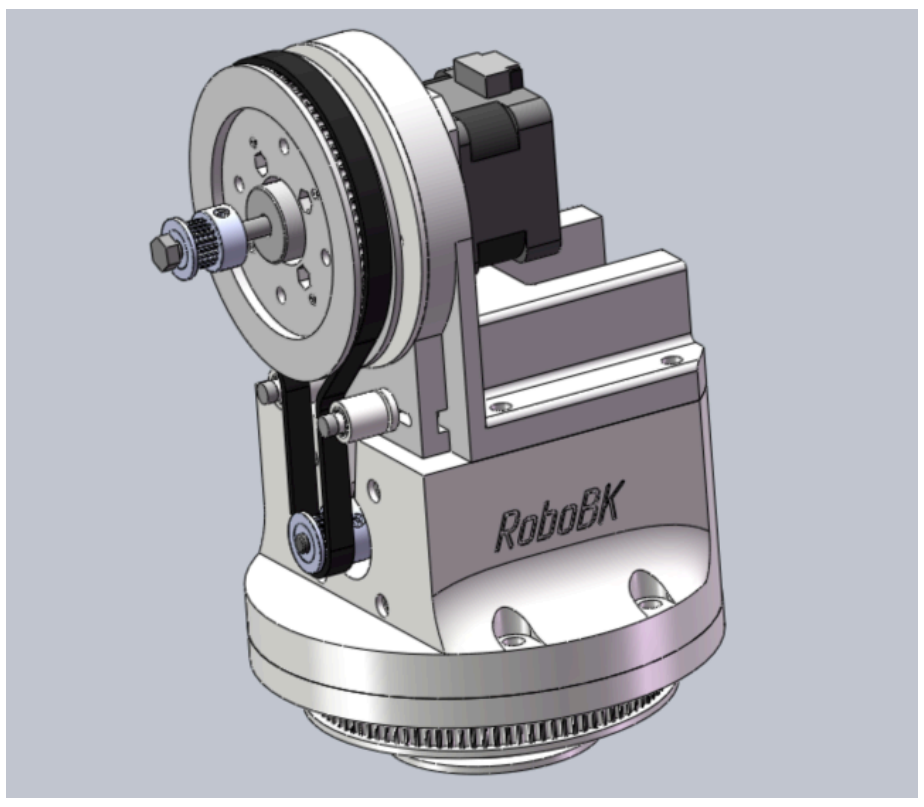

Рис. 18 - Плечо

Ремень тянется от шкива приводного шагового двигателя 42BYGH34, расположенного на соединении плеча, до ведомого шкива на соединении локтя, покрывая верхнюю часть руки). На соединении локтя установлен шаговый двигатель 42BYGH34, который приводит в движение ремень, закрепленный на шкиве соединения нижней части руки (сустав 4).

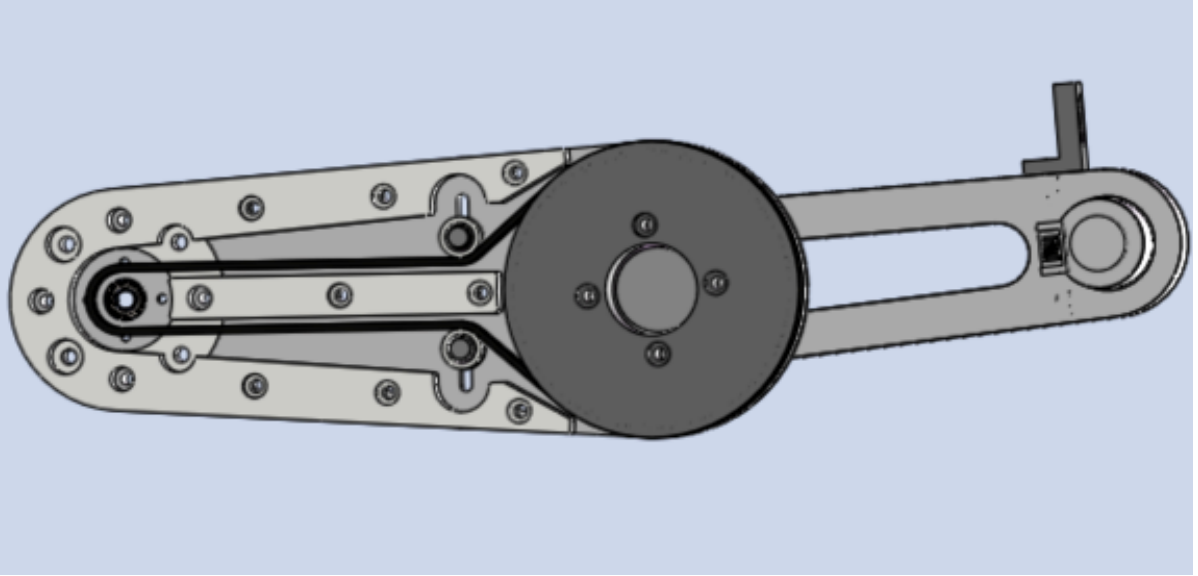

Рис. 19 - Локоть

Перечень комплектующих элементов приведен в таблице 3.

Таблица 3 – Перечень комплектующих элементов

| Категория | Тип | Штук | Цена (Рублей) |
|---|---|---|---|
| ШД | 42BYGH40 | 1 | 550 |
| ШД | 42BYGH34 | 1 | 550 |
| ШД | 42BYGH47 | 1 | 550 |
| Мотор-редуктор | MG310 | 1 | 400 |
| Драйвер | A4988 | 3 | 160 |
| Драйвер | L298N | 1 | 90 |
| Микроконтроллер | Arduino uno R3 | 1 | 360 |
| Плата расширения | CNC Shield V3 | 1 | 100 |
| Преобразователь напряжения | 220/24В-6А-150Вт | 1 | 360 |
| Камера | USB Raspberry Camera | 1 | 180 |
| Замкнутый включатель | KCD 3 | 2 | 50 |
| Муфта | Муфта 5/5 | 1 | 60 |
| Подшипник | 61808ZZ | 1 | 40 |
| Подшипник | 61810ZZ | 2 | 200 |
| Подшипник | F685 | 2 | 80 |

Продолжение таблицы 3

| Категория | Тип | Штук | Цена (Рублей) |
|---|---|---|---|
| Подшипник | 683ZZ | 12 | 150 |
| Шкив | Шкив 9мм/5мм | 3 | 600 |
| Ремень | HTD-3M-9mm-420 | 1 | 120 |
| Ремень | HTD-3M-9mm-339 | 1 | 120 |
| Ремень | HTD-3M-9mm-447 | 1 | 120 |
| 3Д печати | PLA - пластик | 1 | 4000 |

Перечень крепежных деталей приведен в таблице 4.

Таблица 4 - Детали для крепежа

| Категория | Тип | Штук | Цена (Рублей) |
|---|---|---|---|
| Винт | М3*10 | 30 | 26 |
| Винт | М3*30 | 16 | 28 |
| Винт | М3*6 | 40 | 30 |
| Винт | М3*16 | 25 | 30 |
| Винт | М3*20 | 20 | 20 |
| Винт | М4*50 | 7 | 28 |
| Винт | М4*30 | 12 | 23 |
| Винт | М4*12 | 16 | 28 |
| Винт | М4*40 | 9 | 28 |
| Винт | М5*70 | 6 | 52 |
| Гайка | М3 | 60 | 40 |
| Гайка | М4 | 25 | 16 |
| Шайба | 3мм | 25 | 16 |
| Шайба | 4мм | 25 | 18 |

Общая стоимость изготовления макета RoboBK составила 9223 рублей.

## 2.6 Энергетический расчет

Привод создает крутящий момент в суставе, чтобы преодолеть сопротивление звена движению. Сопротивление любого звена движению обусловлено воздействием силы тяжести и инерции. Воздействие силы тяжести на любое звено робота заключается в том, чтобы тянуть и ускорять его к центру земли за счет его собственного веса, таким образом создавая сопротивляющий момент. Поэтому часть создаваемого приводом крутящего момента необходима для преодоления этого сопротивляющего момента (из-за силы тяжести). Было необходимо рассчитать величину сопротивляющего момента, вызванного силой тяжести, действующей на каждое звено руки, чтобы выбрать привод с достаточной мощностью для каждого сустава.

Сопротивляющий момент, действующий на сустав из-за силы тяжести, сильно зависит от положения робота. Интуитивно понятно, что момент на плечевом суставе значительно больше, когда рука вытянута горизонтально. Поэтому для расчета необходимого крутящего момента на каждом суставе была выбрана наихудшая ситуация (полностью вытянутая рука). Рука подвергается наибольшему крутящему моменту, когда она вытянута горизонтально. На рисунке 20 показано обозначение длин звеньев манипулятора, а на рисунке 21 показана схема свободного тела манипулятора.

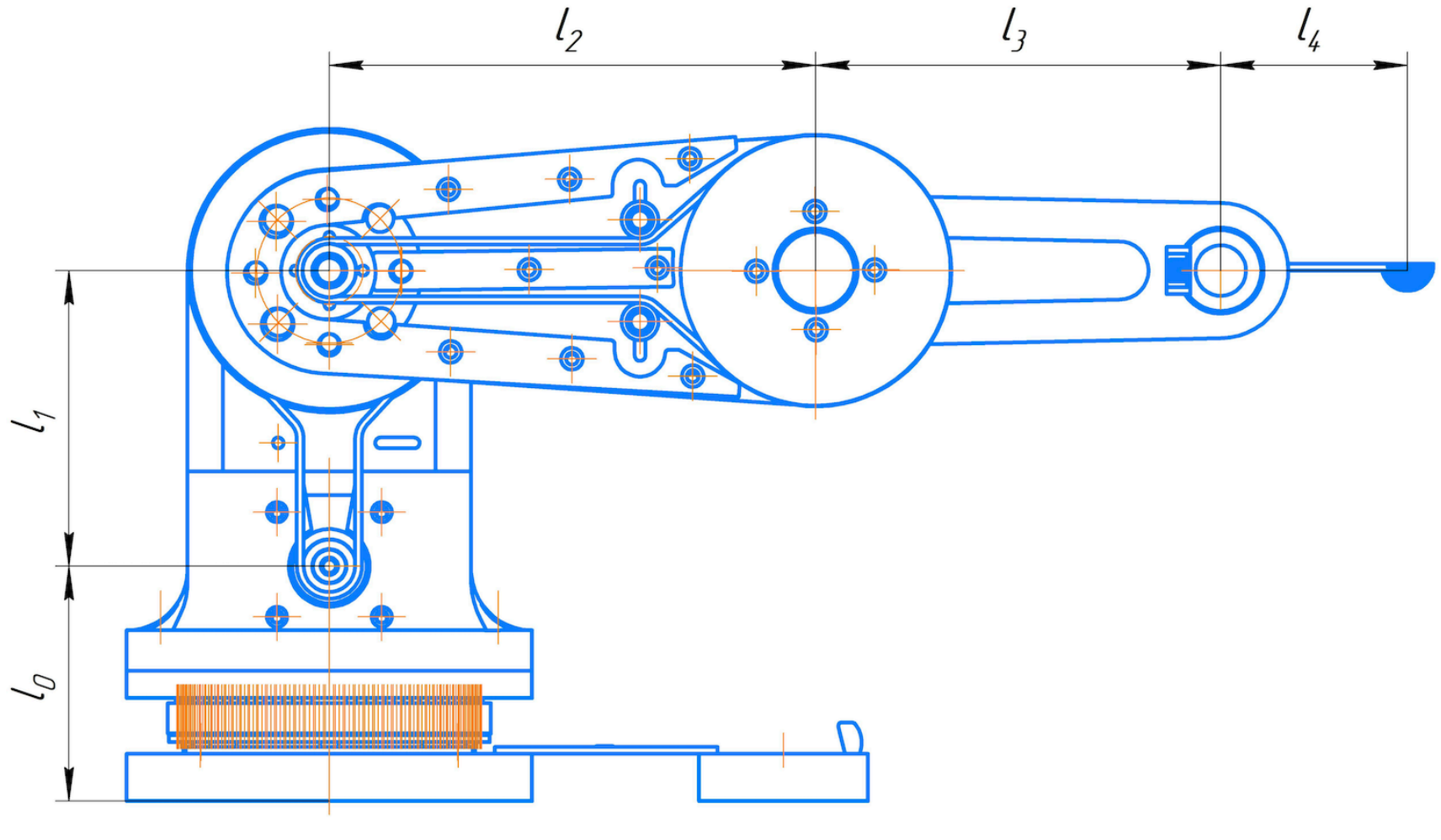

Рис. 20 - Длины звена манипулятора

Исходные данные для расчет:

$T_1 = 2.40$ Н·м - крутящий момент привода звена $L_1$

$T_2 = 1.40$ Н·м - крутящий момент привода звена $L_2$

$T_3 = 2.75$ Н·м - крутящий момент привода звена $L_3$

$T_4 = 0.04$ Н·м - крутящий момент привода звена $L_4$

$\eta_{\text{рп}} = 0.9$ - КПД ремённой передачи

$W_1$ - вес звена $L_1$

$W_2 = 2.73\ H$ - вес звена $L_2$

$W_3 = 1.25\ H$ - вес звена $L_3$

$W_4 = 0.0123\ H$ - вес звена $L_4$

$W_L$ - вес загрузки

$W_{j2}$ - вес сочленения $J_2$

$W_{j3} = 1.67\ H$ - вес сочленения $J_3$

$W_{j4} = 0.40\ H$ - вес сочленения $J_4$

$L_0 = 0.0695$ м - длина звена $L_0$

$L_1 = 0.0875$ м - длина звена $L_1$

$L_2 = 0.1440$ м - длина звена $L_2$

$L_3 = 0.1200$ м - длина звена $L_3$

$L_4 = 0.0553$ м - длина звена $L_4$

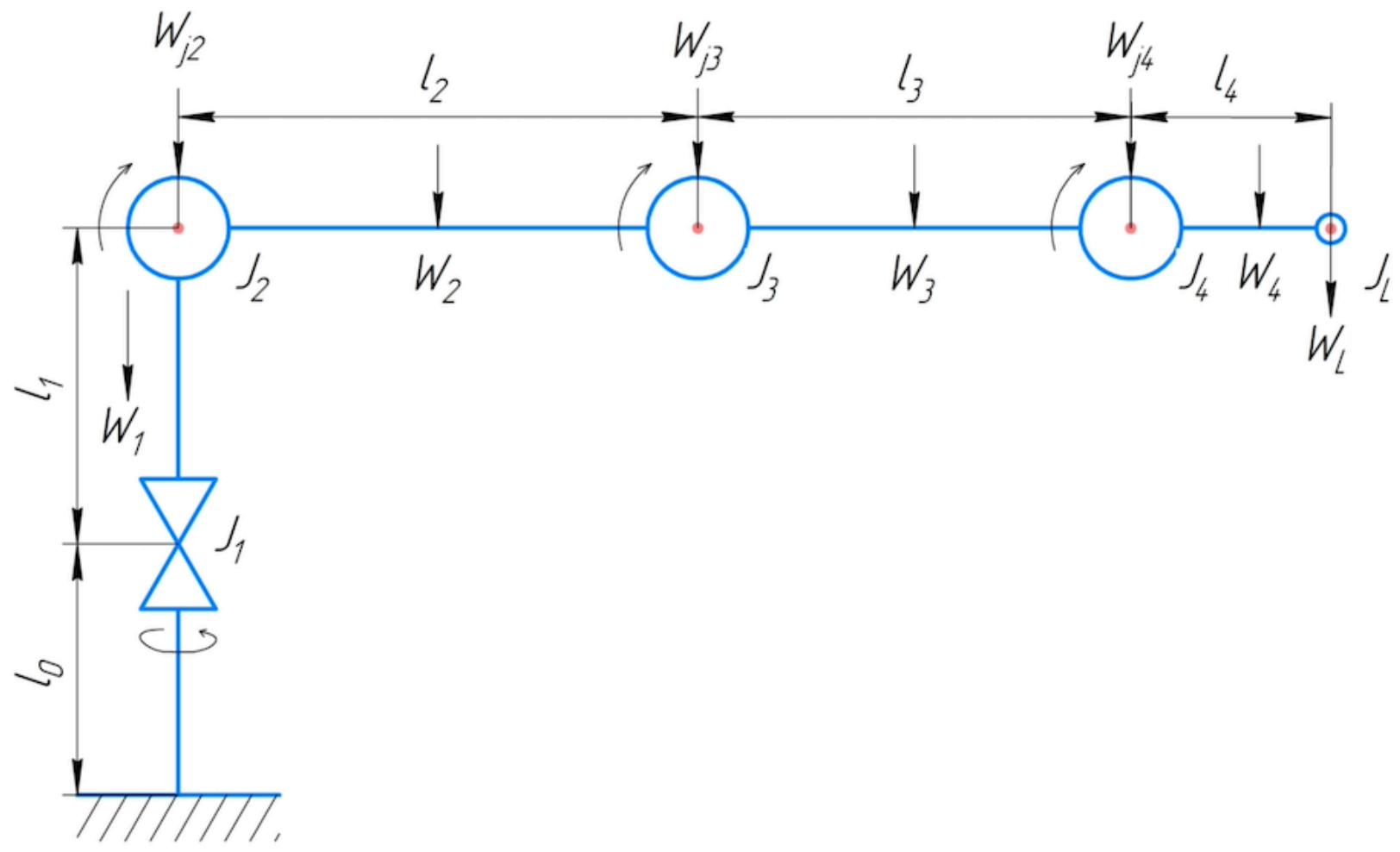

Рис. 21 - Схема свободного тела манипулятора

При условии, что центры всех шарниров находятся в их геометрических центрах, имеются следующие выражения, а именно:

Момент сопротивления в шарнире 1 под действием силы тяжести:

$$T_{1g} = 0$$

Момент сопротивления в шарнире 2 под действием силы тяжести:

$$T_{2g} = W_2\left(\frac{L_2}{2}\right) + W_{J3}L_2 + W_3\left(L_2 + \frac{L_3}{2}\right) + W_{J4}\left(L_2 + L_3\right)$$

$$+W_4(L_2 + L3 + \frac{L_4}{2}) + W_L(L_2 + L_3 + L_4)$$

Момент сопротивления в шарнире 3 под действием силы тяжести:

$$T_{3g} = W_3\left(\frac{L_3}{2}\right) + W_{J4}L_3 + W_4(L3 + \frac{L_4}{2}) + W_L(L_3 + L_4)$$

Момент сопротивления в шарнире 4 под действием силы тяжести:

$$T_{4g} = W_4\frac{L_4}{2} + W_L L_4$$

Используем максимально допустимый крутящий момент четвертого шарнира в размере 0.040 Н·м для расчета максимальной нагрузки. Проверим, чтобы крутящие моменты первых трех шарниров не превышали их максимально допустимых значений и убедимся, что крутящий момент четвертого шарнира не превосходит 0.040 Н·м. Расчет максимального значения $W_L$ с учетом шарнира 4, таким образом:

$$W_L = \frac{0.038660505}{0.0553} \approx 0.699\,\text{Н}$$

При $W_L = 0.699\,\text{N}$, поставить значение в формулы, таким образом:

$$T_{2g} = 1.0246\ \text{Н·м} < T_2 \cdot \eta_{\text{рп}} = 1.26\ \text{Н·м}$$

$$T_{3g} = 0.247\ \text{Н·м} < T_3 \cdot \eta_{\text{рп}} = 2.475\ \text{Н·м}$$

Таким образом, максимально допустимая нагрузка $W_L = 0.699\,\text{N}$ является максимальной, которую могут выдержать первые три шарнира.

Учитывая возможность замены конечного исполнительного устройства манипулятора, необходимо рассчитать максимальную нагрузку третьего шарнира без учета четвертого шарнира, таким образом:

$$T_{3g} = W_3\left(\frac{L_3}{2}\right) + W_{J4}L_3 + W_L(L_3)$$

$$T_{2g} = W_2\left(\frac{L_2}{2}\right) + W_{J3}L_2 + W_3\left(L_2 + \frac{L_3}{2}\right) + W_{J4}(L_2 + L_3) + W_L(L_2 + L_3)$$

Тогда максимальная нагрузка с учетом третьего сочленения:

$$W_L = 19.6\ \text{Н·м}$$

Тогда максимальная нагрузка с учетом третьего и четвертого сочленения:

$$W_L = 1.75\ \text{Н·м}$$

Третье сочленение могло бы выдержать нагрузку до 21.89 Н, однако возможности второго сочленения ограничивают общую нагрузку до более низкого значения.

Для учета динамики и расчета максимальной нагрузки, необходимо учитывать не только силу тяжести, но и инерционные силы, вызванные ускорением и скоростью двигателя. В частности, нам нужно рассчитать инерционные силы, возникающие при ускорении и замедлении, и их влияние на каждый шарнир.

Исходные данные для расчет:

$m_2 = 0.273$ кг - масса звена $L_2$

$m_3 = 0.125$ кг - масса звена $L_3$

$\alpha_{max} = 9.8125\ rad/s^2$- максимальное угловое ускорение приводов

Расчет момента инерции для второго и третьего звена:

$$I_2 = \frac{1}{3}m_2L_2^2 = \frac{1}{3} \times 0.273 \times (0.144)^2 \approx 0.001888\,\text{кг} \cdot \text{м}^2$$

$$I_3 = \frac{1}{3}m_3L_3^2 = \frac{1}{3} \times 0.012 \times (0.12)^2 \approx 0.0006\,\text{кг} \cdot \text{м}^2$$

Расчет инерционного момента, вызванного угловым ускорением:

$$\tau_2 = I_2 \cdot \alpha_{\text{max}} = 0.001888 \times 9.8125 \approx 0.01852\,\text{Н} \cdot \text{м}$$

$$\tau_3 = I_3 \cdot \alpha_{\text{max}} = 0.0006 \times 9.8125 \approx 0.00589\,\text{Н} \cdot \text{м}$$

Тогда суммарный крутящий момент второго сустава:

$$T_2 = W_2\left(\frac{L_2}{2}\right) + W_{J3}L_2 + W_3\left(L_2 + \frac{L_3}{2}\right) + W_{J4}(L_2 + L_3) + W_L(L_2 + L_3) + \tau_2$$
$$= 0.81616 + 0.264W_L$$

Тогда суммарный крутящий момент второго сустава:

$$T_3 = W_3\left(\frac{L_3}{2}\right) + W_{J4}L_3 + W_L(L_3) + \tau_3 = 0.12889 + 0.12W_L$$

Максимальная нагрузка для второго сочленения:

$$W_L \leq \frac{0.44384}{0.264} \approx 1.6812\,\text{Н}$$

Максимальная нагрузка для второго сочленения:

$$W_L \leq \frac{2.346}{0.12} \approx 19.55\,\text{Н}$$

Поскольку максимальная нагрузка второго сочленения меньше, мы должны выбрать более низкое значение, т.е. максимальная нагрузка $W_L = 1.6812\ H$.

# 3. Моделирование робота RoboBK

## 3.1 Кинематика

**Прямая кинематика (ПЗК)**: Это процесс определения положения и ориентации конечного эффектора (кончика роботизированной руки) на основе параметров сочленений (углы $q_1$, $q_2$, $q_3$, $q_4$ и длины звеньев $l_1$, $l_2$, $l_3$, $l_4$)

**Обратная кинематика (ОЗК)**: Это процесс определения параметров сочленений, которые позволят достичь желаемого положения и ориентации конечного эффектора.

Важной характеристикой манипулятора является число степеней его подвижности. имеет только кинематические пары пятого класса: четыре вращательные пары. Для произвольной пространственной кинематической цепи в общем случае следует использовать формулу Сомова-Малышева.

$$W_{\text{пр}} = 6 \cdot n - 5 \cdot p_5 = 6 \cdot 4 - 5 \cdot 4 = 4$$

где $W_{\text{пр}}$ – степень подвижности кинематической цепи;

$p_5$ – число КП 5-го класса;

$n$ – число подвижных звеньев кинематической цепи.

Классификация кинематических пар в нижней таблице 2.

Таблица 5 - Классификация кинематических пар

| **Обозначение КП** | **Звенья КП** | **Относительное движение** |
|---|---|---|
| O | 0,1 | Вращательная |
| A | 1,2 | Вращательная |
| B | 2,3 | Вращательная |
| C | 3,4 | Вращательная |

Перед проведением кинематического анализа роботизированной руки, сначала в Matlab создается модель руки с использованием модели Денавита-Хартенберга (Д-Х) в таблице 6 с следующими базовыми параметрами:

$\theta_i$ - это угол сочленения вокруг предыдущей оси $z$

$d_i$ - это смещение вдоль предыдущей оси $z$ к общей нормали

$a_i$ - это длина общей нормали

$\alpha_i$ - это угол вокруг общей нормали для выравнивания предыдущей оси $z$ с текущей осью $z$

Таблица 6 - Таблица Денавита-Хартенберга (Д-Х)

| $i$ | $\theta_i$ | $d_i$ **(мм)** | $a_i$ **(мм)** | $\alpha_i$ | **Limits** |
|---|---|---|---|---|---|
| 1 | $q_1$ | 155.5 | 0 | $\pi/2$ | $[-180\ 180]$ |
| 2 | $q_2$ | 74 | 144 | 0 | $[-144\ 144]$ |
| 3 | $q_3$ | -67.7 | 120 | 0 | $[-155\ 155]$ |
| 4 | $q_4$ | -6 | 63 | 0 | $[-180\ 180]$ |

Далее с помощью Д-Х параметров в среде Matlab получена модель манипулятора.

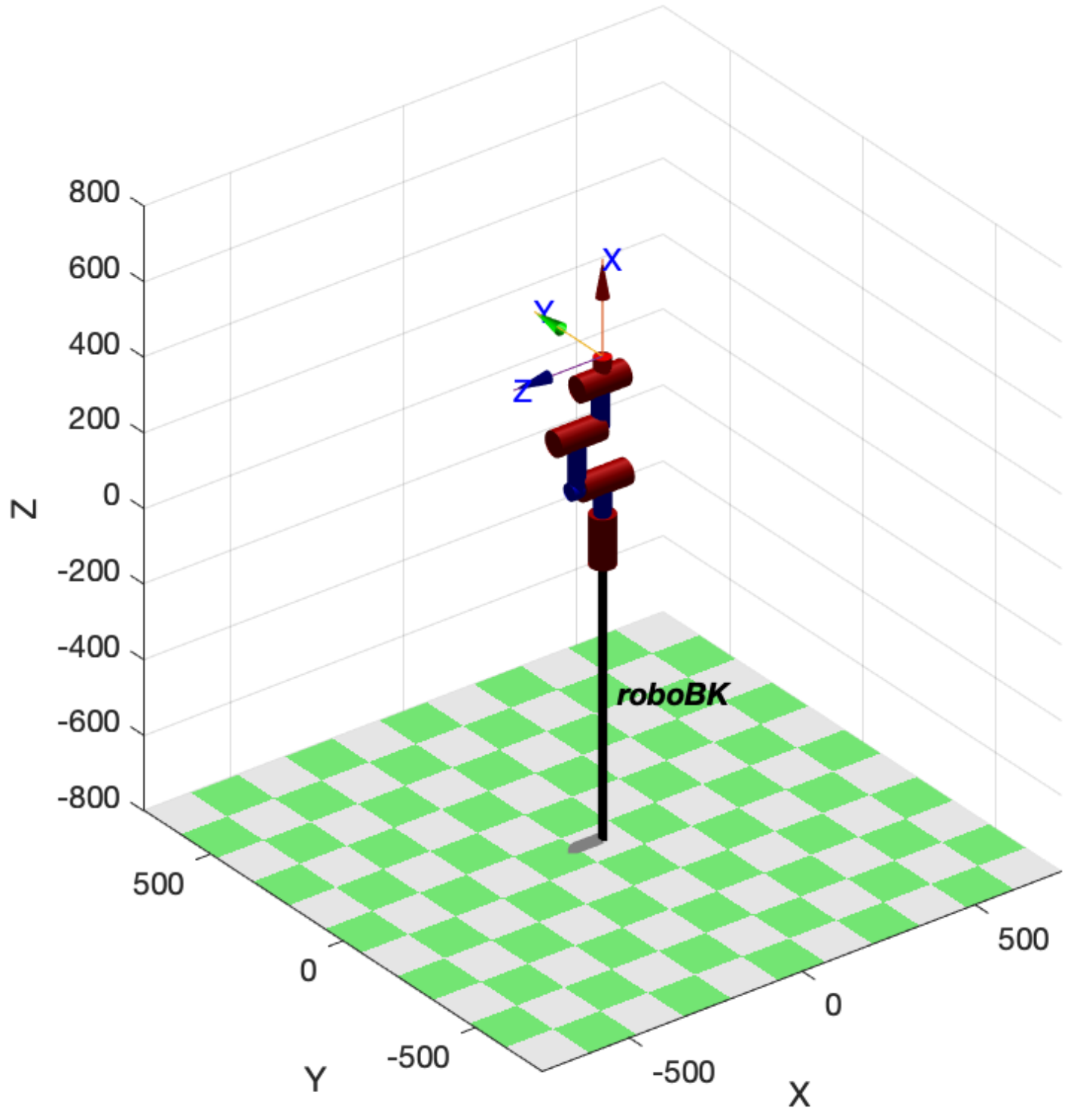

Рис. 22 - Модель манипулятора, полученная с Д-Х параметров

### 3.1.1 Прямая задача по кинематике

Для решения задачи прямой кинематики (ПЗК) мы строим матрицы преобразования для каждого звена, используя эти параметры. Матрица преобразования для каждого звена $i$ от его предыдущего звена $i-1$ задается следующим образом:

$$ {}^{i}_{i-1}T = \begin{bmatrix} c(\theta_i) & -s(\theta_i)c(\alpha_i) & s(\theta_i)s(\alpha_i) & a_i c(\theta_i) \\ s(\theta_i) & c(\theta_i)c(\alpha_i) & -c(\theta_i)s(\alpha_i) & a_i s(\theta_i) \\ 0 & s(\alpha_i) & c(\alpha_i) & d_i \\ 0 & 0 & 0 & 1 \end{bmatrix} $$

Матрицы перехода для каждого звена:

$$ {}^{0}T_1 = \begin{bmatrix} \cos\theta_1 & -\sin\theta_1 & 0 & 0 \\ 0 & 0 & 1 & 155.5 \\ -\sin\theta_1 & -\cos\theta_1 & 0 & 0 \\ 0 & 0 & 0 & 1 \end{bmatrix} $$

$$ {}^{1}T_2 = \begin{bmatrix} \cos\theta_2 & -\sin\theta_2 & 0 & 144\cos\theta_2 \\ \sin\theta_2 & \cos\theta_2 & 0 & 144\sin\theta_2 \\ 0 & 0 & 1 & 74 \\ 0 & 0 & 0 & 1 \end{bmatrix} $$

$$ {}^{2}T_3 = \begin{bmatrix} \cos\theta_3 & -\sin\theta_3 & 0 & 120\cos\theta_3 \\ \sin\theta_3 & \cos\theta_3 & 0 & 120\sin\theta_3 \\ 0 & 0 & 1 & -67.7 \\ 0 & 0 & 0 & 1 \end{bmatrix} $$

$$ {}^{3}T_4 = \begin{bmatrix} \cos\theta_4 & -\sin\theta_4 & 0 & 63\cos\theta_4 \\ \sin\theta_4 & \cos\theta_4 & 0 & 63\sin\theta_4 \\ 0 & 0 & 1 & -6 \\ 0 & 0 & 0 & 1 \end{bmatrix} $$

Объединим их в последовательность, чтобы упростить.

$$^0T_4 = {}^0T_1 \cdot {}^1T_2 \cdot {}^2T_3 \cdot {}^3T_4$$

$$^0T_4 = \begin{bmatrix} c_1c_{234} & -c_1s_{234} & 0 & c_1(d_4c_{234} + a_3c_{23} + a_2c_2) \\ s_1c_{234} & -s_1s_{234} & 0 & s_1(d_4c_{234} + a_3c_{23} + a_2c_2) \\ -s_{234} & -c_{234} & 1 & d_1 - a_2s_2 - a_3s_{23} - d_4s_{234} \\ 0 & 0 & 0 & 1 \end{bmatrix}$$

Общая матрица преобразования, связывающая координатные системы конца манипулятора с базовой координатной системой, определяется как:

$$^0T_4 = \begin{bmatrix} n_x & o_x & a_x & p_x \\ n_y & o_y & a_y & p_y \\ n_z & o_z & a_z & p_z \\ 0 & 0 & 0 & 1 \end{bmatrix}$$

Где $\{n_x, n_y, n_z\}$, $\{o_x, o_y, o_z\}$, $\{a_x, a_y, a_z\}$ представляют собой векторы ориентации, составляющие матрицу вращения, в то время как $(p_x, p_y, p_z)$ представляют собой положение конечного эффектора.

Положение конечного эффектора:

$$p_x = c_1(d_4c_{234} + a_3c_{23} + a_2c_2)$$

$$p_y = s_1(d_4c_{234} + a_3c_{23} + a_2c_2)$$

$$p_z = d_1 - a_2s_2 - a_3s_{23} - d_4s_{234}$$

Ориентация конечного эффектора:

$$n_x = c_1c_{234}$$

$$n_y = s_1c_{234}$$

$$n_z = -s_{234}$$

$$o_x = -c_1s_{234}$$

$$o_y = -s_1 s_{234}$$

$$o_z = -c_{234}$$

$$a_x = 0$$

$$a_y = 0$$

$$a_z = 1$$

где

$$c_i = \cos(\theta_i)$$

$$s_i = \sin(\theta_i)$$

$$c_{ij} = \cos(\theta_i + \theta_j)$$

$$s_{ij} = \sin(\theta_i + \theta_j)$$

$$c_{ijk} = \cos(\theta_i + \theta_j + \theta_k)$$

$$s_{ijk} = \sin(\theta_i + \theta_j + \theta_k)$$

Это выражение может быть использовано для вычисления положения рабочего органа для любого заданного набора углов сочленений $q_i$ и длин звеньев $l_i$.

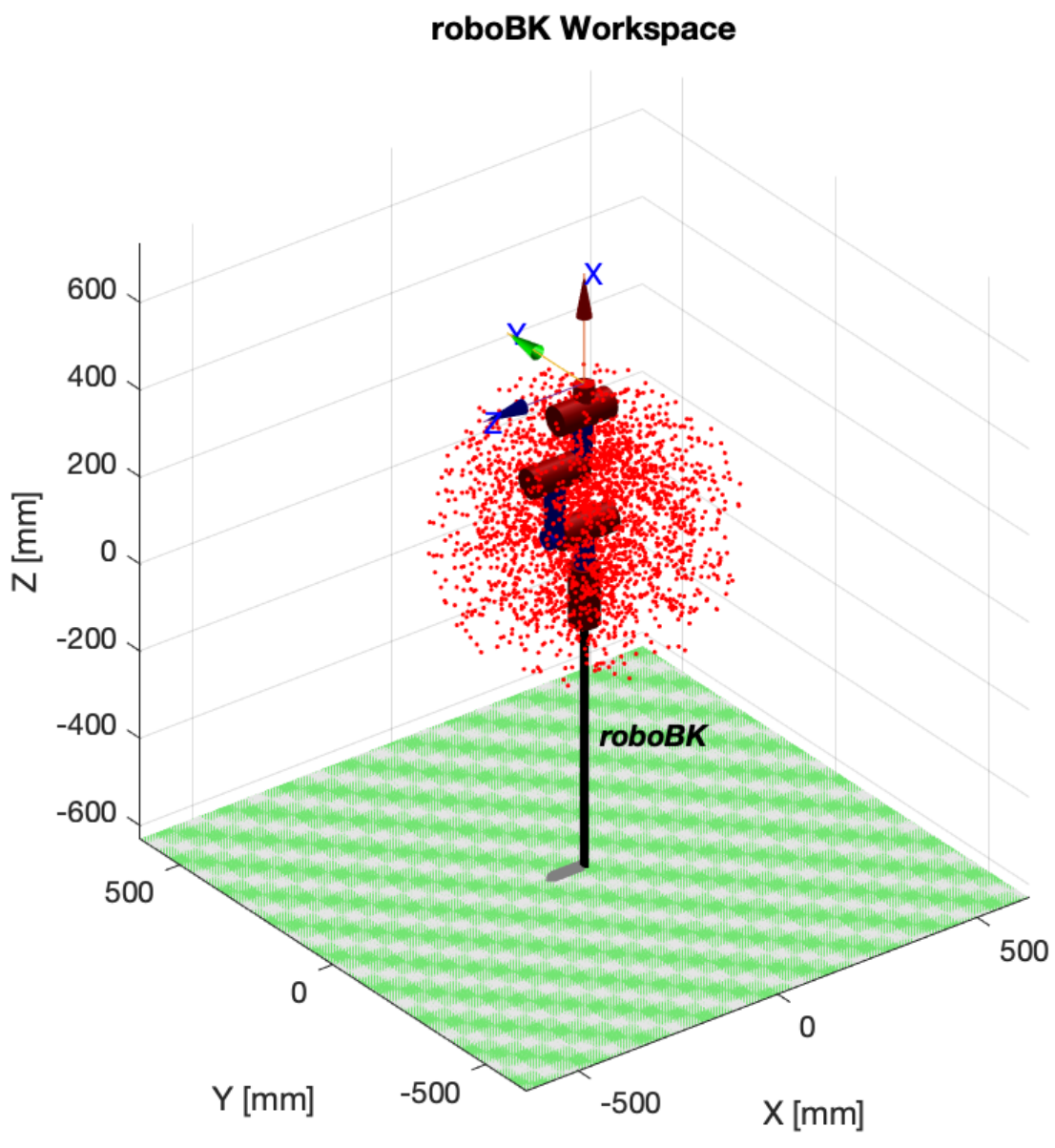

Рис. 23 - Рабочее пространство

В этом разделе метод Монте-Карло используется для решения рабочей области следующим образом: во-первых, для каждого сустава генерируются случайные переменные и создается случайный набор векторов суставного пространства для манипулятора, используя 2000 точек. Во-вторых, вычисляется положительное решение кинематики и отображается из суставного пространства в конечное рабочее пространство (декартова система координат), и, наконец, результат наносится на рисунке 24.

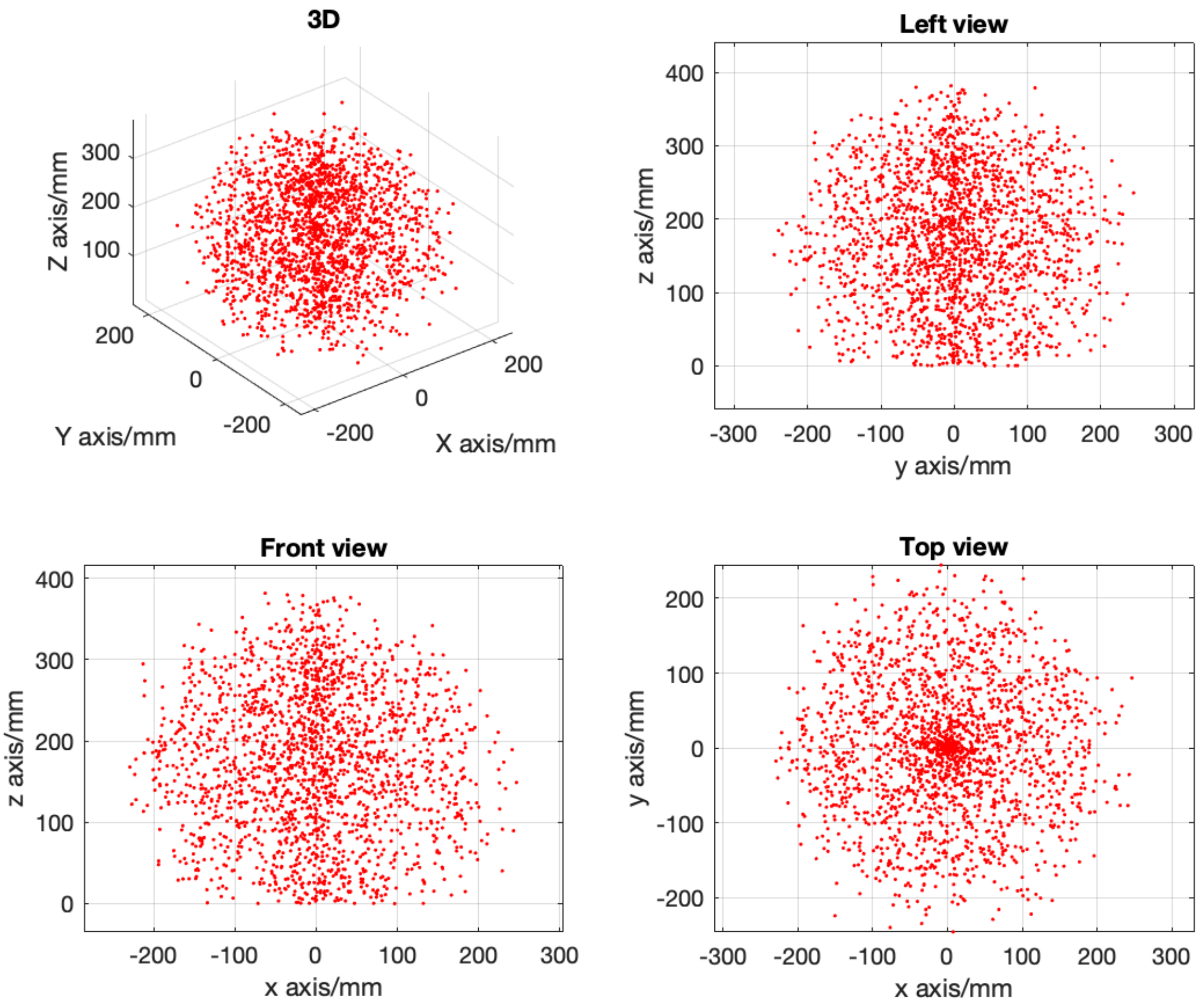

Рис. 24 - Рабочие плоскости

Чтобы проиллюстрировать прямую кинематику данного манипулятора, рассматриваются заданные углы каждого звена. Они определяется следующим параметрическим представлением:

$$\theta_1 = -1.0\sin(\omega t)$$

$$\theta_2 = 1.5\sin(\omega t)$$

$$\theta_3 = 0.8\sin(\omega t)$$

$$\theta_4 = -0.5\sin(\omega t)$$

где

$$T = 5;$$

$$\omega = 2 * pi/T;$$

$$t \in [\ 0\ ,\ 2.5\ ]$$

В Matlab используя Simscape, создана модель манипулятора (см. Приложение 1), проведена кинематическая симуляция. Заданы ориентация (рис. 25) , получено положение конечного эффекта (рис 26).

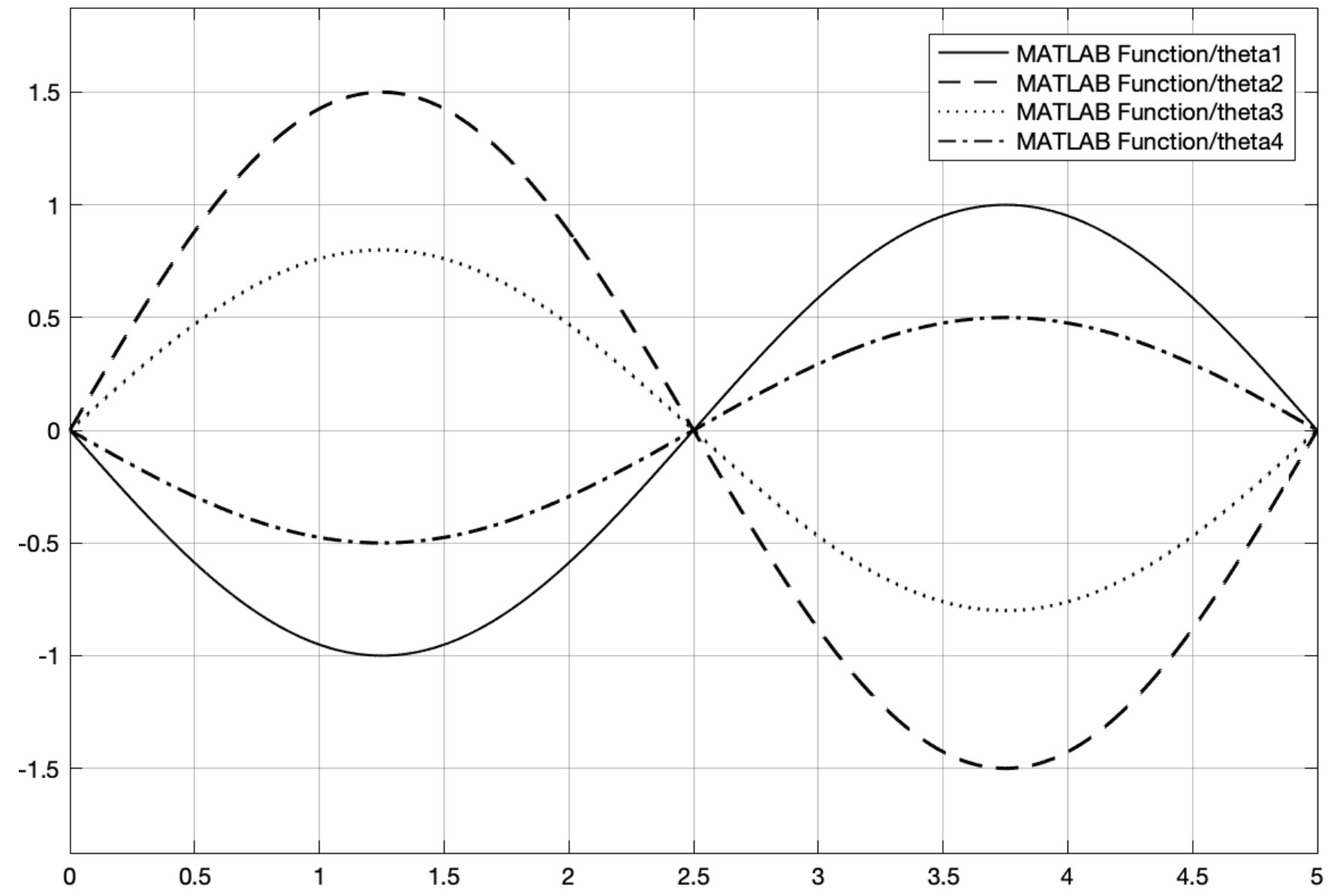

Рис. 25 - Заданные ориентации

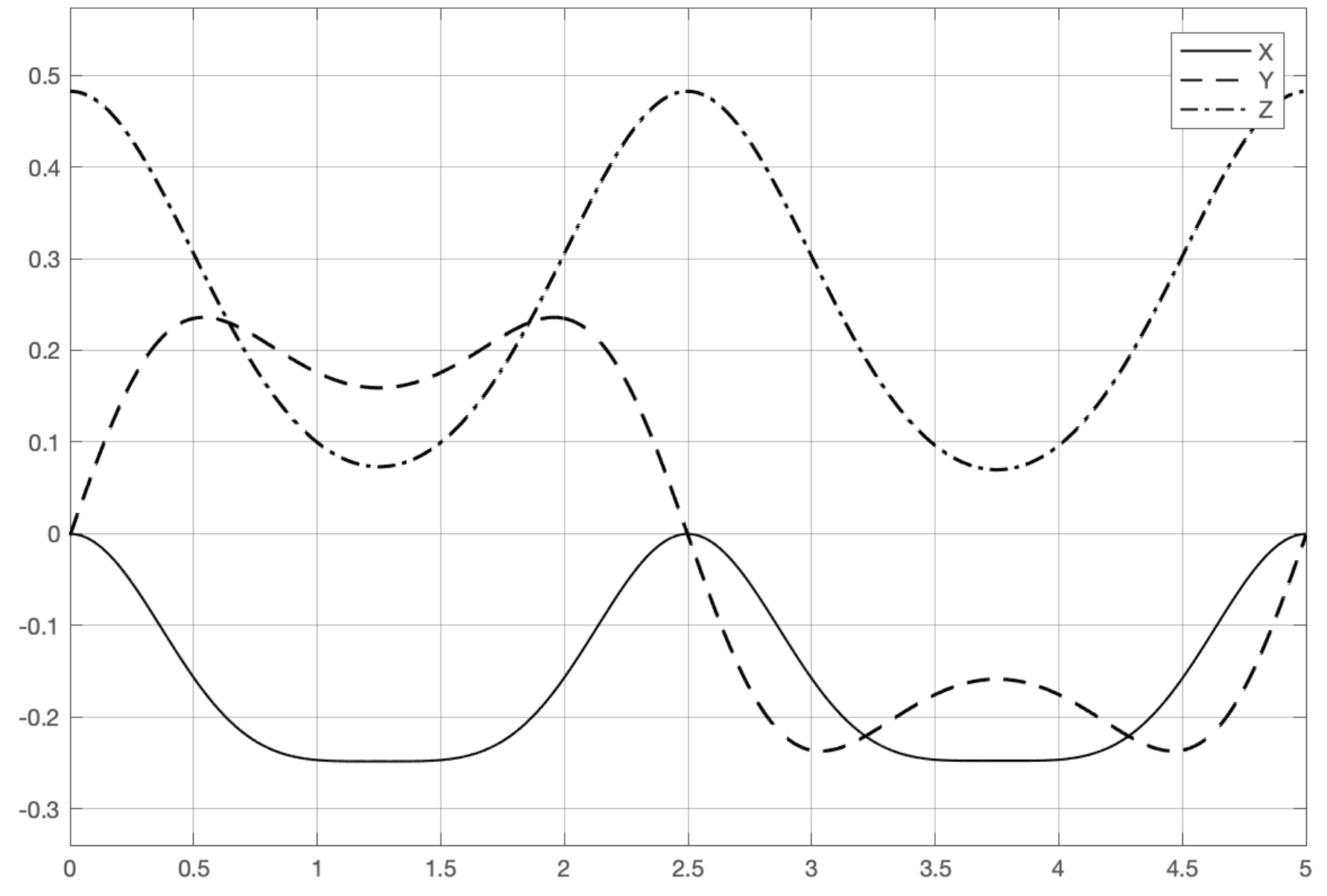

Рис. 26 - Полученное положение

### 3.1.2 Обратная задача по кинематике

Для обратной кинематики нам нужно конкретное целевое положение и, возможно, ориентация рабочего органа для решения углов сочленений.

Задача может быть решена различными методами: аналитическим методом, алгебраическим методом и так далее. Аналитический метод используется для решения задачи на следующих этапах:

Вычисление угла первого сочленения $\theta_1$:

Исходя из положения концевого эффектора, можно вывести угол первого сочленения $\theta_1$. Учитывая высоту основания робота $d_1$ и положение $(p_x, p_y)$

$$\theta_1 = \tan^{-1}\left(\frac{p_y}{p_x}\right)$$

Вычисление угла первого сочленения $\theta_3$:

Используя положение концевого эффектора $(p_x, p_y, p_z)$ и известные длины звеньев.

Определим:

$$r = \sqrt{p_x^2 + p_y^2} - a_4$$

Тогда получим:

$$\cos(\theta_3) = \frac{r^2 + (p_z - d_1)^2 - a_2^2 - a_3^2}{2a_2a_3}$$

$$\sin(\theta_3) = \sqrt{1 - \cos^2(\theta_3)}$$

Вычисление угла первого сочленения $\theta_2$:

$$\theta_2 = \tan^{-1}\left(\frac{p_z - d_1}{r}\right) - \tan^{-1}\left(\frac{a_3\sin(\theta_3)}{a_2 + a_3\cos(\theta_3)}\right)$$

Вычисление угла первого сочленения $\theta_4$:

$$\theta_4 = \theta_{234} - \theta_2 - \theta_3$$

где:

$$\theta_{234} = \tan^{-1}\left(\frac{a_z}{\sqrt{n_x^2 + o_x^2}}\right)$$

Чтобы проиллюстрировать обратную кинематику данного манипулятора, рассматривается криволинейная траектория рассматривается. Она определяется следующим параметрическим представлением:

$$p_x = 0.15 + 0.25\sin\left(\frac{2\pi}{5}t\right)$$

$$p_y = 0.15\sin\left(\frac{2\pi}{5}t + \frac{\pi}{2}\right)$$

$$p_z = 0.15 + 0.1\sin\left(\frac{2\pi}{5}t + \frac{\pi}{4}\right)$$

где $t \in [\ 0\ ,\ 2.5\ ]$

В Matlab используя Simscape, создана модель манипулятора (см. Приложение 2), проведена кинематическая симуляция. Заданное положение конечного эффекта (рис. 27) , получена ориентация (рис 28).

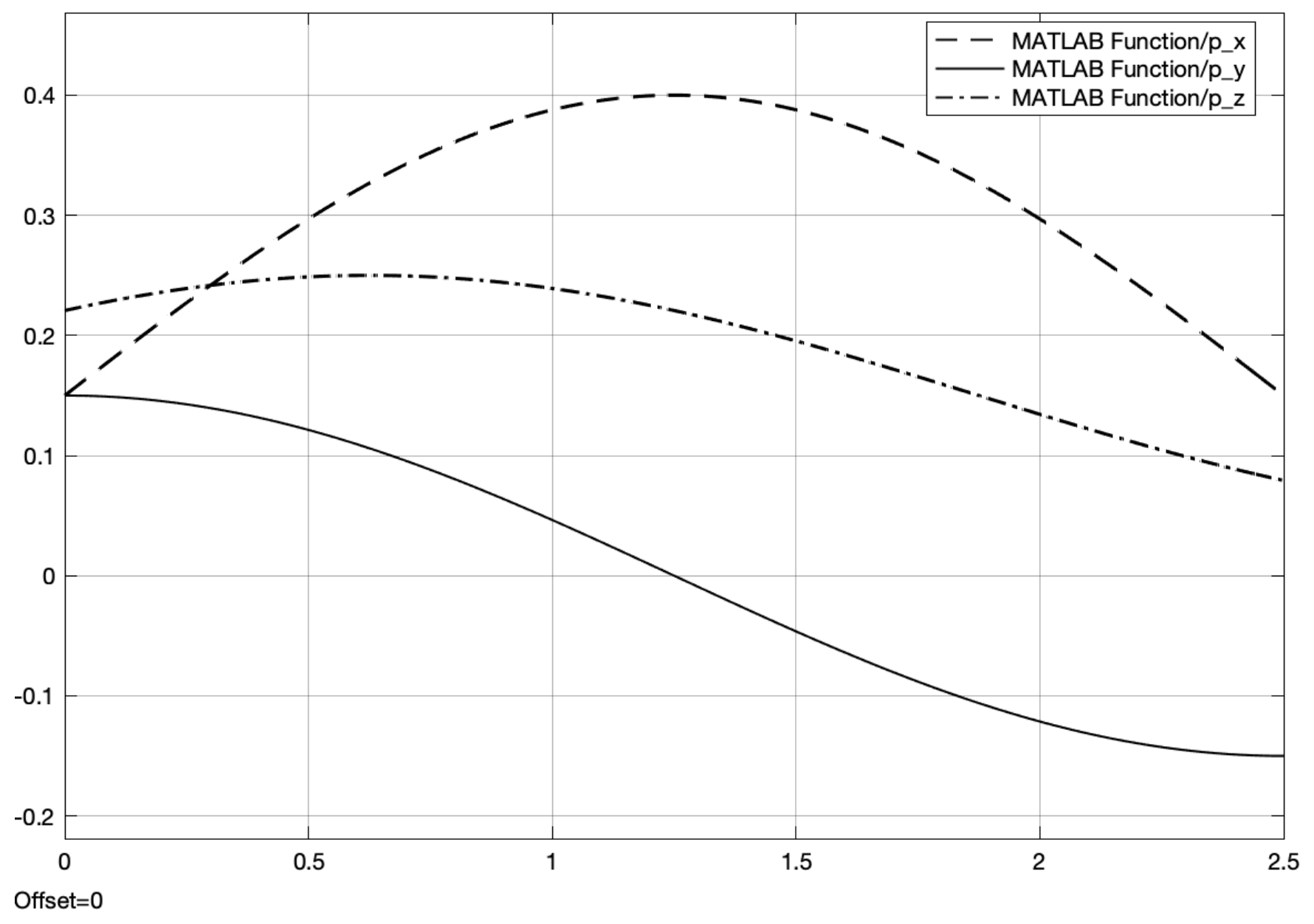

Рис. 27 - Заданное положение

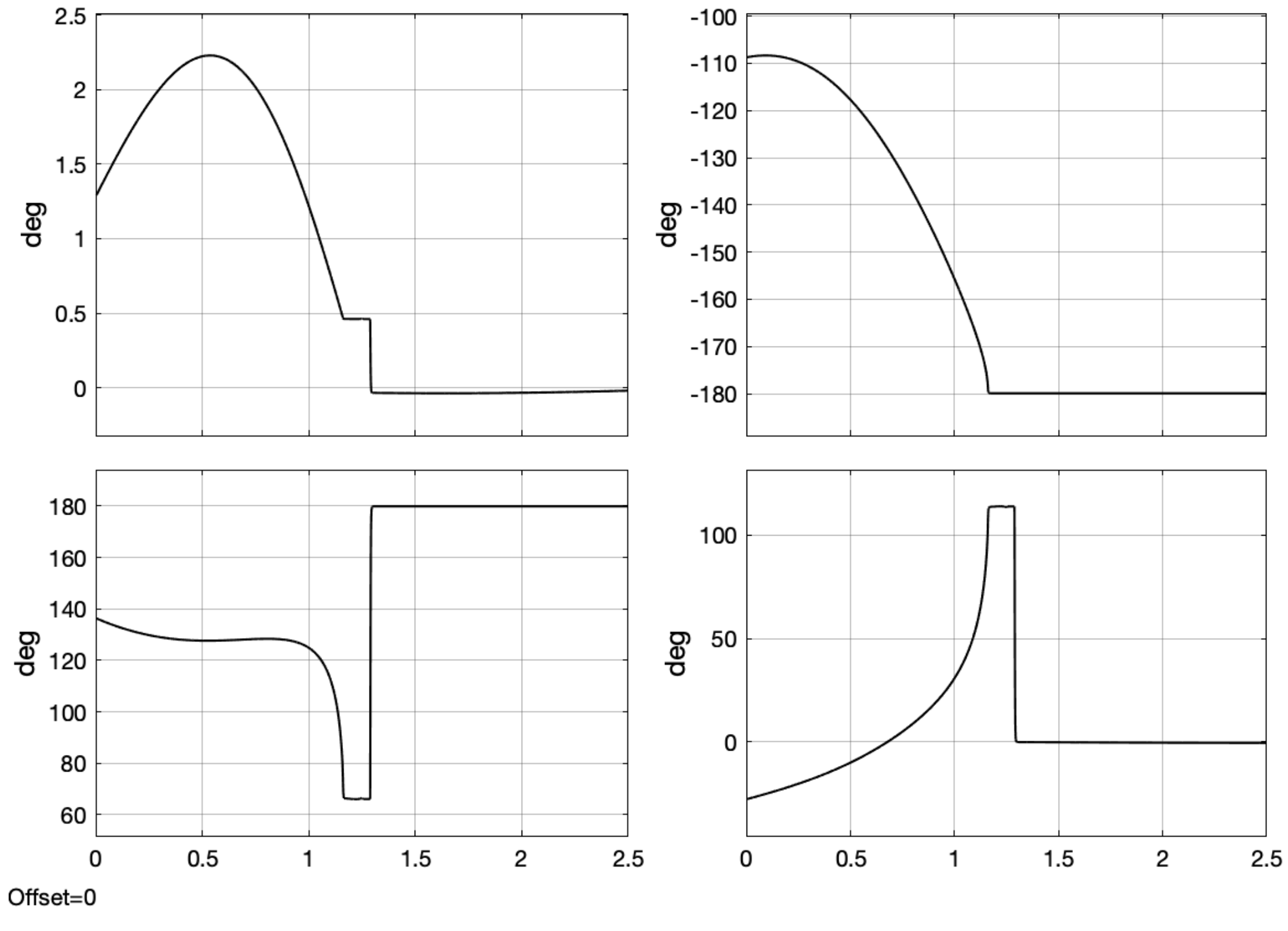

Рис. 28 - Полученные ориентации

## 3.2 Динамика

Анализ матрицы Якоби. Дифференциальное движение манипулятора также важно, когда манипулятор должен следовать траектории с разными скоростями. Отношение между скоростью концевого эффектора $\dot{X}$ и векторами скоростей суставов $\dot{\theta}$ задается матрицей Якоби $[J]$. Оно определяется следующим образом:

$$\dot{X} = [J]\,\dot{\theta}$$

где $[J]$ - матрица Якоби для RoboBK

$$[J] = \begin{bmatrix} \frac{\partial p_x}{\partial \theta_1} & \frac{\partial p_x}{\partial \theta_2} & \frac{\partial p_x}{\partial \theta_3} & \frac{\partial p_x}{\partial \theta_4} \\ \frac{\partial p_y}{\partial \theta_1} & \frac{\partial p_y}{\partial \theta_2} & \frac{\partial p_y}{\partial \theta_3} & \frac{\partial p_y}{\partial \theta_4} \\ \frac{\partial p_z}{\partial \theta_1} & \frac{\partial p_z}{\partial \theta_2} & \frac{\partial p_z}{\partial \theta_3} & \frac{\partial p_z}{\partial \theta_4} \end{bmatrix}$$

Динамический анализ также полезен для проектирования механического прототипа и вычисления различных сил и моментов, необходимых для работы манипулятора. Существует два вида задач в динамике.

Прямая динамика: процесс вычисления движения манипулятора по заданным силам и моментам привода. Обратная динамика: процесс вычисления приводных сил или моментов в суставах по заданным положениям, скоростям и ускорениям конечного эффектора. Мы используем прямую динамику как модель манипулятора, тогда как обратная динамика используется для валидации динамической модели или иногда в управлении.

Таблица 7 - Прямая задача по динамике

| **Известно** | **Найти** |
|---|---|
| Приложенные силы и моменты | Ускорения звеньев/суставов |
| Начальные положения звеньев | Скорости звеньев/суставов |
| Начальные скорости звеньев | Положения звеньев/суставов |

Таблица 8 - Прямая задача по динамике

| **Известно** | **Найти** |
|---|---|
| Положения звеньев/суставов | Приложенные силы и моменты |
| Скорости звеньев/суставов | |
| Ускорения звеньев/суставов | |

Для манипулятора RoboBK мы будем проводить динамическое моделирование и симуляцию в Simulink в среде Matlab. Однако в реальном манипуляторе из-за ограничений двигателей и датчиков, составляющих манипулятор, невозможно реализовать такое же динамическое управление, как в моделировании и симуляции.

Общий подход заключается в следующем: Используя блок *Trapezoidal Velocity Profile Trajectory* (рис.29), генерируются траектории положения (q), скорости (qd) и ускорения (qdd) для каждого сустава в зависимости от заданного времени.

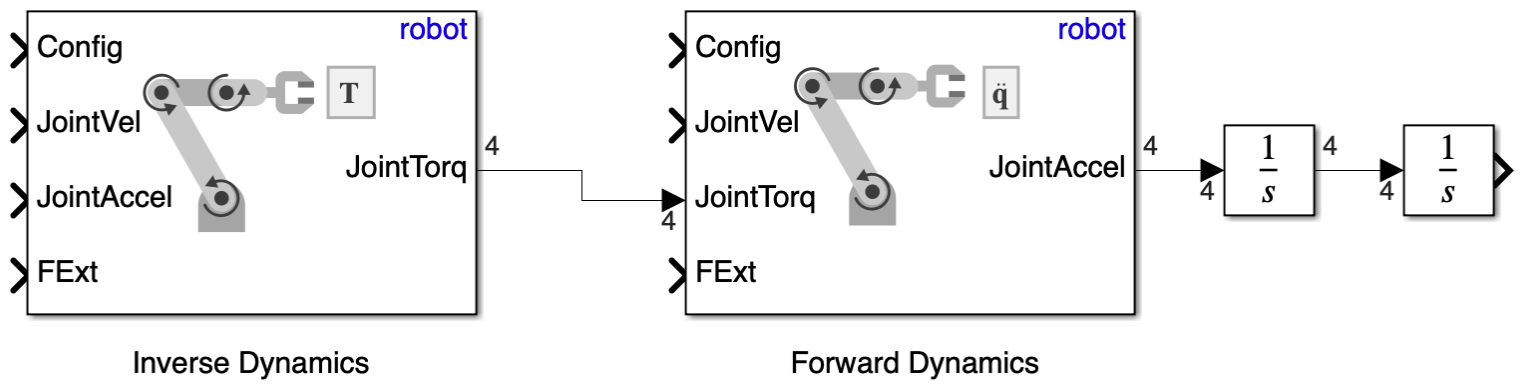

Рис. 29 - Используемые блоки для решение задачи динамики

Эти параметры суставов подаются на вход *Inverse Dynamics* для вычисления необходимых моментов суставов. Затем полученные моменты подаются на вход *Forward Dynamics*, вместе с реальными параметрами суставов, которые являются начальными условиями для механической руки. В результате получаются ускорения, которые через двойное интегрирование дают скорость и положение каждого сустава. *Inverse Dynamics* и *Forward Dynamics* могут считывать входной файл URDF, включая массу суставов, моменты инерции и другие параметры. Мы можем сравнить полученные параметры суставов с начальными параметрами, введенными в Trapezoidal Velocity Profile Trajectory, и они будут полностью совпадать. В Matlab используя Simscape, создана модель манипулятора (см. Приложение 3), проведена динамическая симуляция. Заданы моменты (рис.30), получены ускорения, скорости, перемещение (рис. 31).

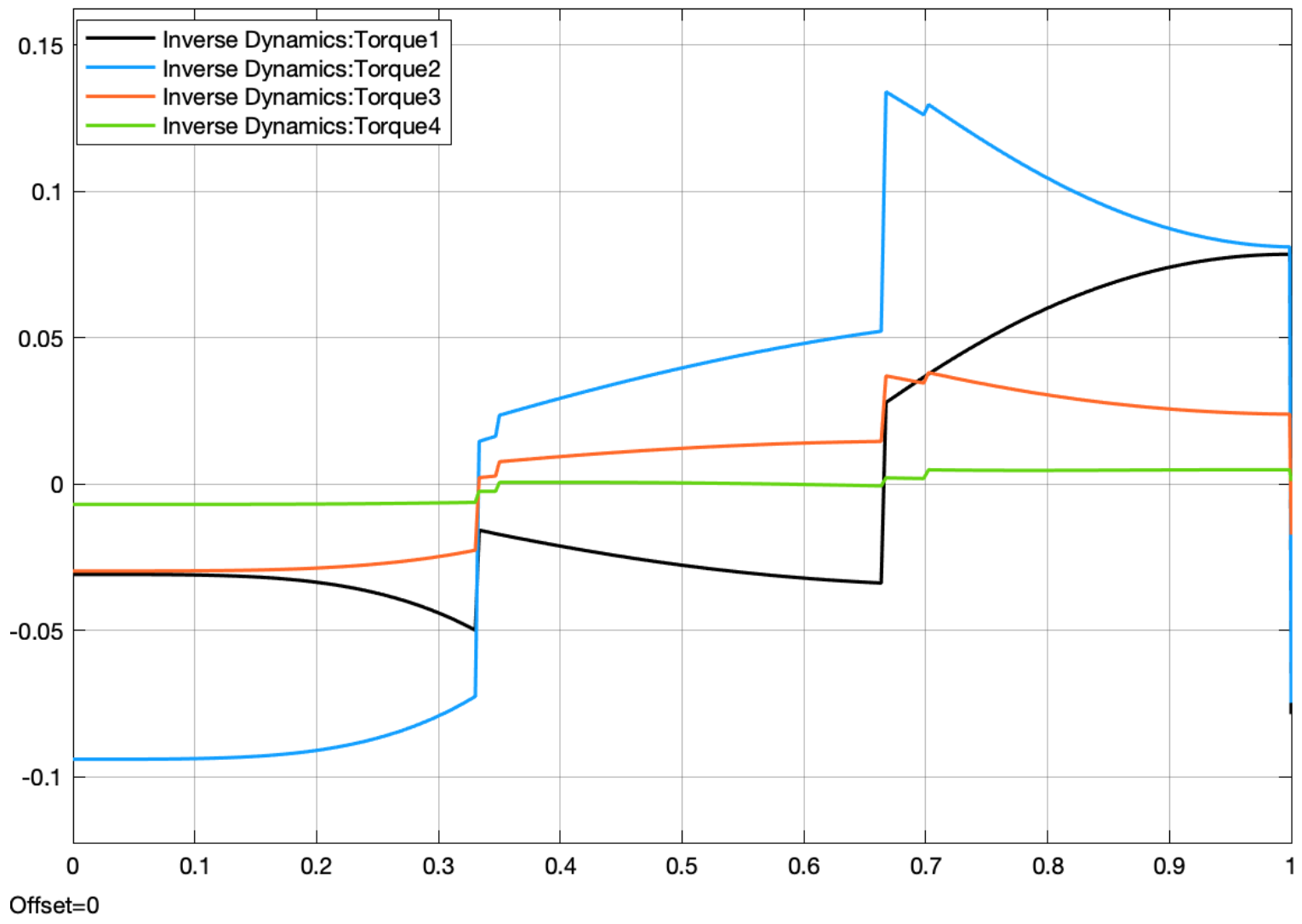

Рис. 30 - Заданные моменты

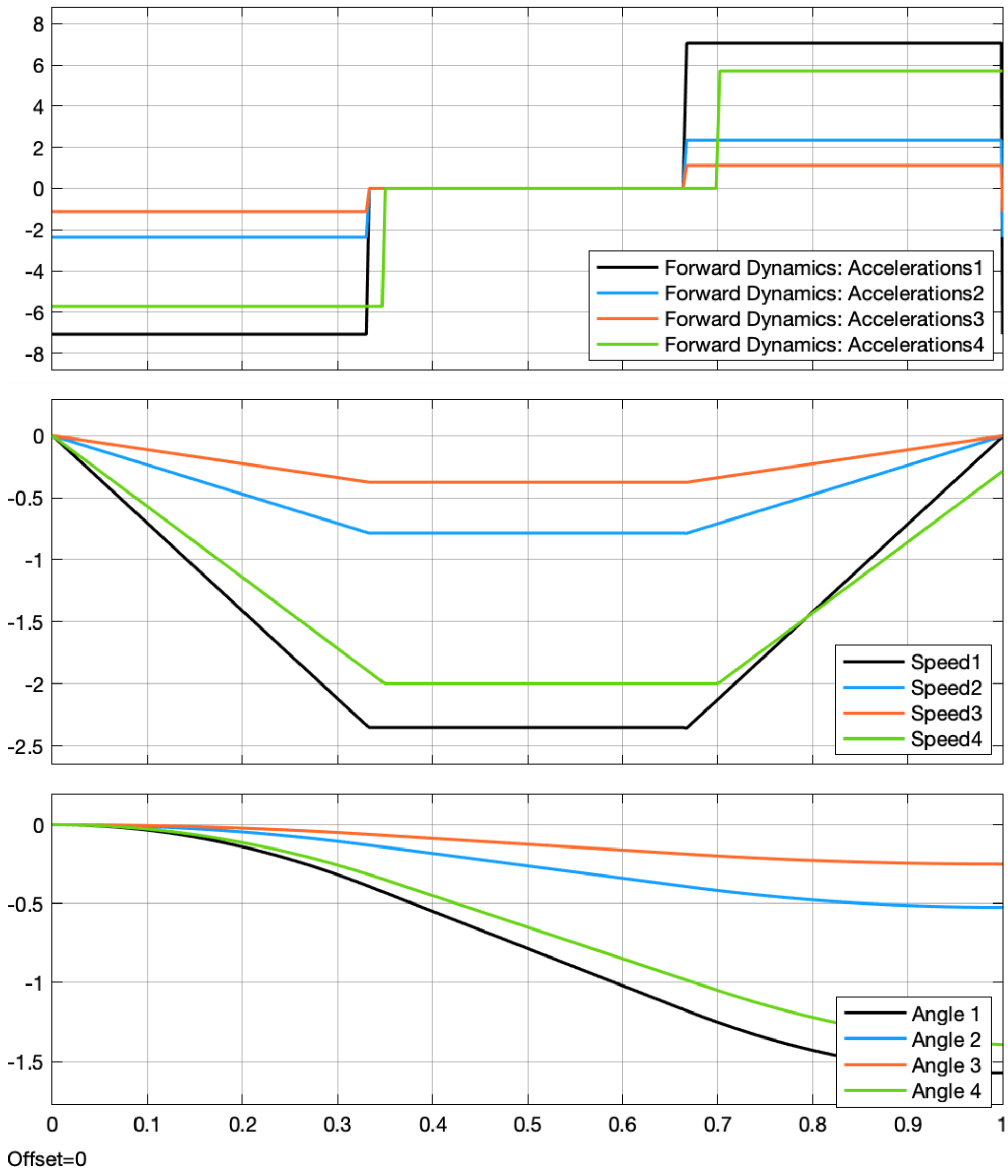

Рис. 31 - Полученные ускорения, скорости, перемещения.

## 3.3 Система управления РТС

В проекте есть два объекта:

- Управляемый
- Управляющий

Управляемый объект в нашем случае является просто объектом (пищи) , а управляющий объект регуляторами. Регулятор действует на объект не прямо, а через исполнительные механизмы (приводы) , которые могут усилить и преобразовать сигнал управления.

### 3.3.1 Система управления для ШД

Система управления шаговым двигателем (ШД) без обратной связи - это тип системы управления, который может управлять движением двигателя, не требуя получения информации о его фактическом положении или скорости. В такой системе предполагается, что команды, отправленные на двигатель, будут точно выполнены, и двигатель достигнет заданного положения без необходимости подтверждения фактического положения.

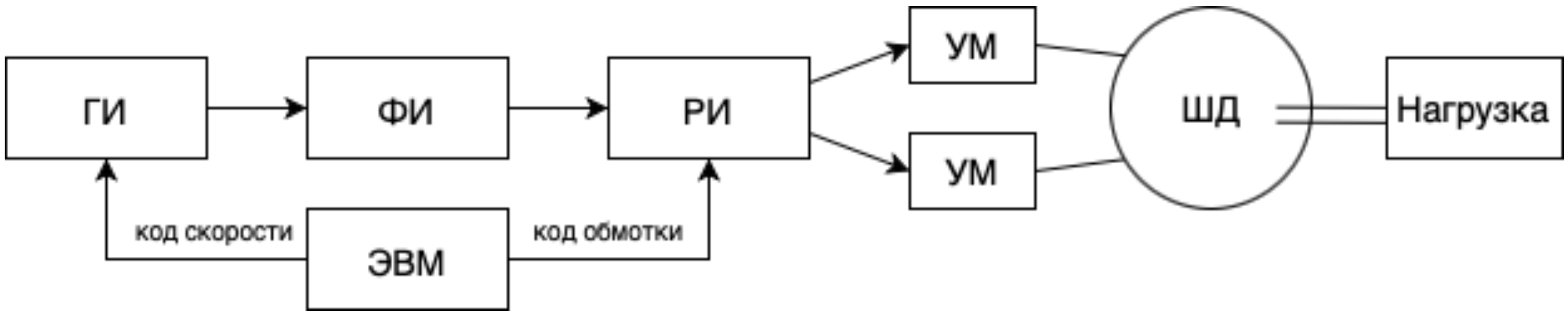

Рис.32 - Система управления ШД

Основной принцип такой системы заключается в том, что шаговый двигатель имеет фиксированное количество шагов для совершения полного оборота, и каждый импульс заставляет мотор поворачиваться на точно определенный угол. Таким образом, если известно начальное положение двигателя, то можно точно управлять его положением, отправляя определенное количество импульсов.

### 3.3.2 Система управления для мотора-редуктора

Макет нашего робота представляет собой платформу с управляющей платой Arduino UNO на контроллере Atmega328P. Приводом конечного исполнителя является MG310 со встроенными датчиками Холла (энкодерами), которые позволят нам судить о направлении вращения , его скорость и угол поворота. Экодеры подключаются к контроллеру, двигатель – к драйверу двигателя ATM8236 (рис.14).

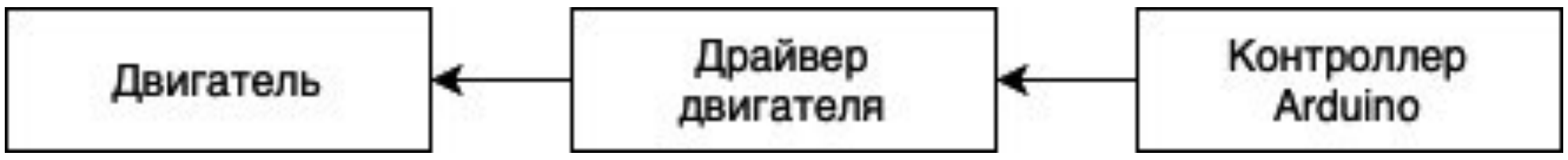

Рис.32 - Схема взаимодействия контроллера, драйвера и двигателя

Драйвер двигателей позволяет преобразовывать уровни сигналов, посылаемых контроллером (0- ~5В) в уровень напряжения питания устройства (0 ~12В)

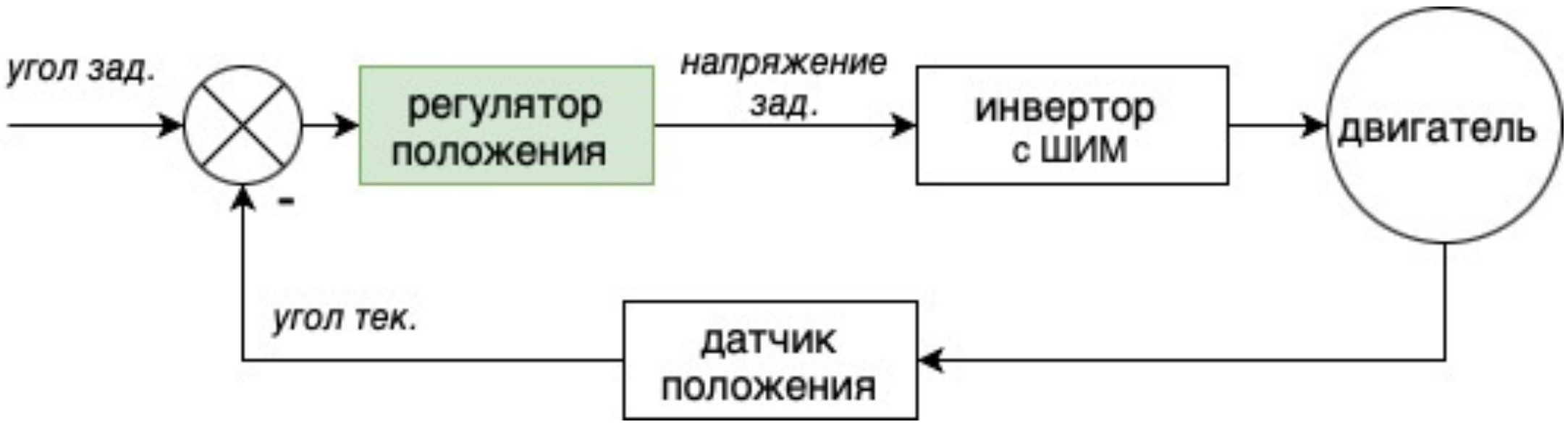

Рис.33 - Система управления двигателем с простым обратным свясью

Выбираем двухконтурную систему управления для прямого управления ДПТ предпочтительнее простого негативного обратного связи (рис.15) из-за нескольких причин. Основная причина заключается в том, что двухконтурная система обеспечивает более высокую точность и динамику управления, поддерживая заданное положение вала ротора даже при возмущающих воздействиях (рис.16). Каждый контур (скорости и положения) "подчиняется" вышестоящему и обеспечивает оптимальное управление определенной характеристикой двигателя. Это позволяет системе эффективно реагировать на изменения и поддерживать стабильное управление двигателем. Мы используем холловский энкодер как датчик обратной связи. Холловский энкодер — это тип датчика, который преобразует механическое перемещение вала в импульсный или цифровой сигнал с помощью магнитного преобразования. Холловские энкодеры состоят из магнита и сенсорного элемента Холла. Магнит закреплен на валу и вращается вместе с ним. Сенсорный элемент Холла обнаруживает изменения магнитного поля, что позволяет определить положение и направление вращения вала. Как правило, холловские энкодеры имеют два выхода с фазовым сдвигом, что позволяет точно определить направление и скорость вращения.

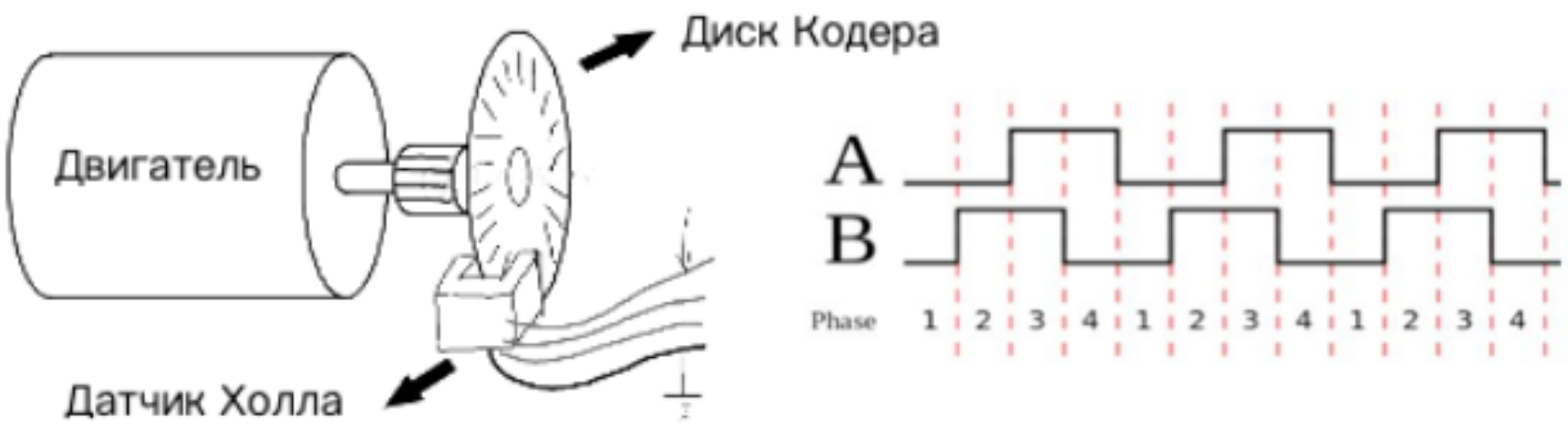

Рис.34 - Принцип работы датчика Холла

Алгоритм измерения скорости: При вращении магнита датчик Холла генерирует электрические сигналы. Датчик выдает два квадратных сигнала (фазы A и B) с фазовым сдвигом 90 градусов. Тогда скорость V равна:

$$V = \frac{N}{T}$$

Где N - количество импульсов

T - временной интервал

Алгоритм измерения положения: Направление вращения вала определяется анализом фазового сдвига между сигналами фаз A и B. Положение рассчитывается путем суммирования количества импульсов, что позволяет определить угловое перемещение вала. Тогда текущее положение $P$ равно:

$$P = \frac{N \times 360^{\circ}}{PPR}$$

Где N - количество импульсов

PPR - количество импульсов на один оборот

PID-параметры для контура скорости и контура положения можно получить у поставщика, продающего MG310.

PID-параметры контура положения:

- $K_{p1} = 120$
- $K_{i1} = 10$
- $K_{d1} = 500$

PID-параметры контура скорости:

- $K_{p2} = 20$
- $K_{i2} = 30$

Чтобы определить передаточную функцию двигателя на основе PID-параметров, нужно понять связь между PID-контроллером и передаточной функцией двигателя. PID-параметры отражают динамику системы управления, что позволяет оценить частотные характеристики и вывести передаточную функцию двигателя.

Передаточная функция для контура скорости может быть приближена к системе первого порядка, так как используется PI-регулятор:

$$G_v(s) = \frac{K}{Js + b}$$

Где $J$ - момент инерции

$b$ - коэффициент трения

$K$ - постоянная двигателя

На основе параметров PI-контроллера для контура скорости можно получить частотные характеристики системы. Предположим, что система в замкнутом контуре скорости имеет следующие характеристики:

Вычисление собственной частоты $\omega_n$:

$$\omega_n \approx \sqrt{K_{p2}} \approx \sqrt{20} \approx 4.47$$

Вычисление коэффициента демпфирования $\zeta$:

$$\zeta \approx \frac{K_{i2}}{2\sqrt{K_{p2}}} \approx \frac{30}{2\sqrt{20}} \approx 1.67$$

Определение параметров передаточной функции:

$$J \approx \frac{1}{\omega_n^2} = \frac{1}{4.47^2} \approx 0.05$$

$$b \approx 2\zeta\omega_n J = 2 \times 1.67 \times 4.47 \times 0.05 \approx 0.75$$

Предположим, что постоянная двигателя приблизительно равна 1 тогда передаточная функция будет:

$$G_v(s) = \frac{K}{Js + b} = \frac{1}{0.05s + 0.75}$$

Структурная схема контура положения с внутренними контурами скорости указана на рисунке 35. В рис. 36 и 37 указаны результаты моделирования. Фактически все нормально.

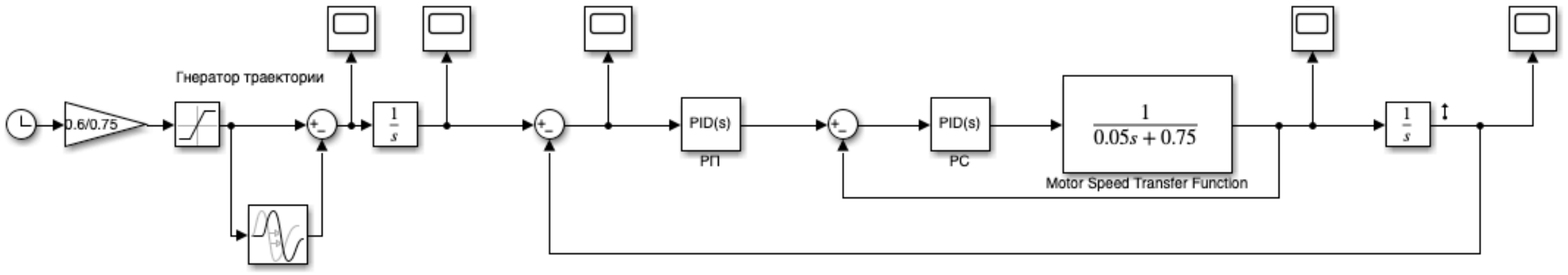

Рис.35 - Структурная схема контура положения

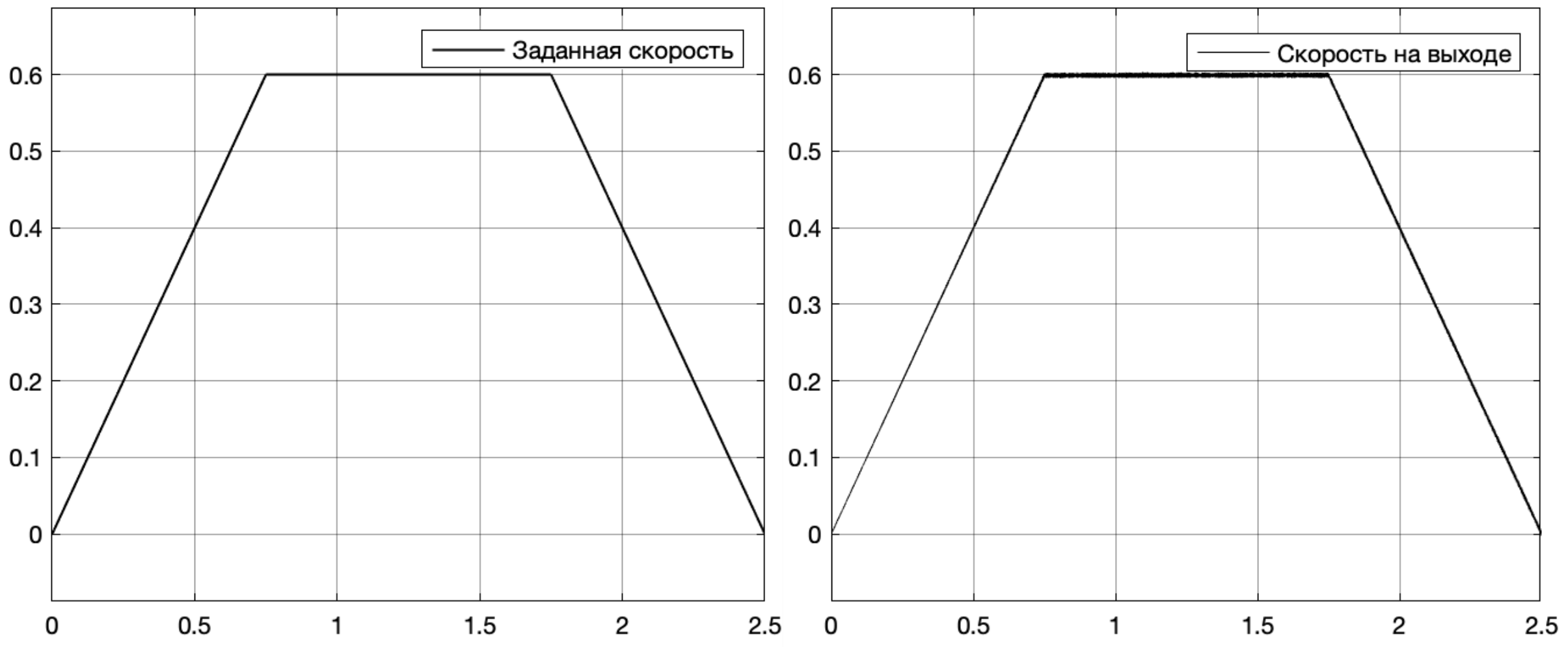

Рис.36 - Графики переходного процесса контура скорости при сигнале с заданным законом изменения скорости

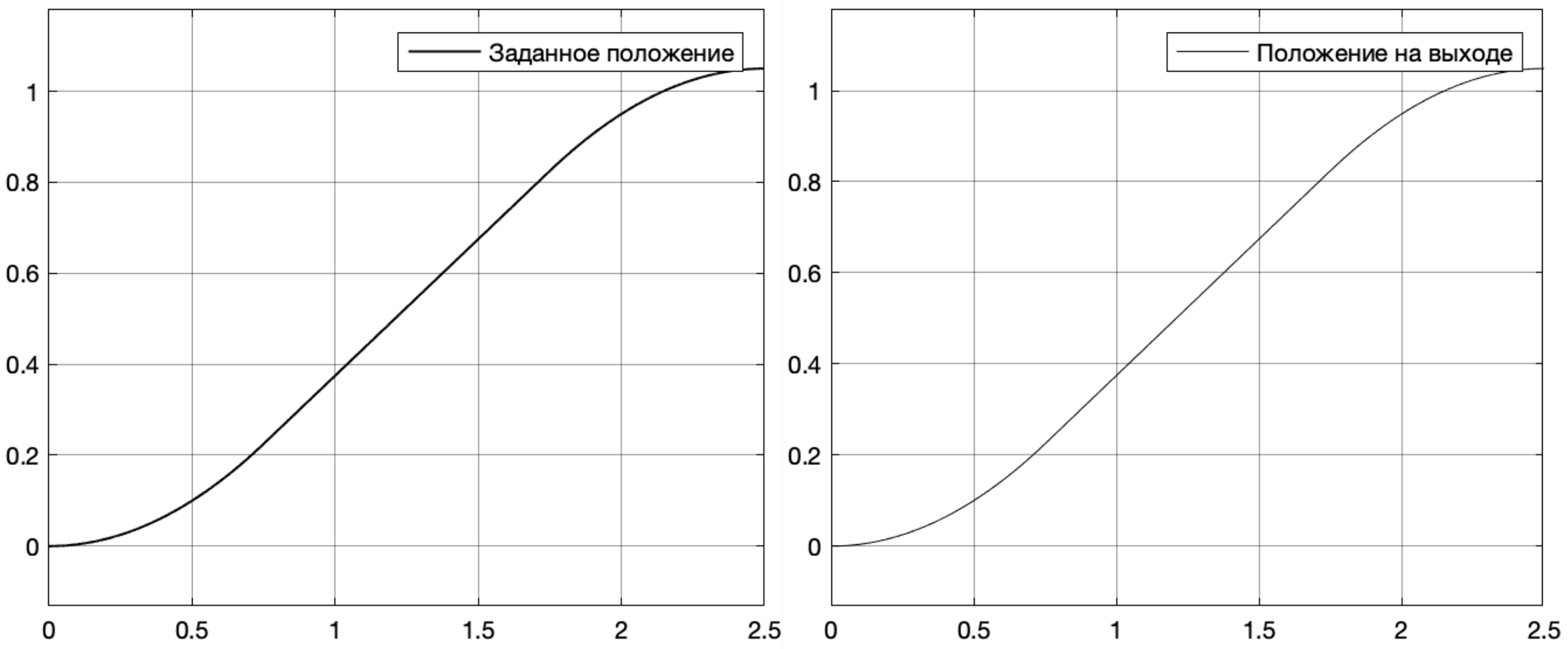

Рис.37 - Графики переходного процесса контура положения по при сигнале с заданным законом изменения положения

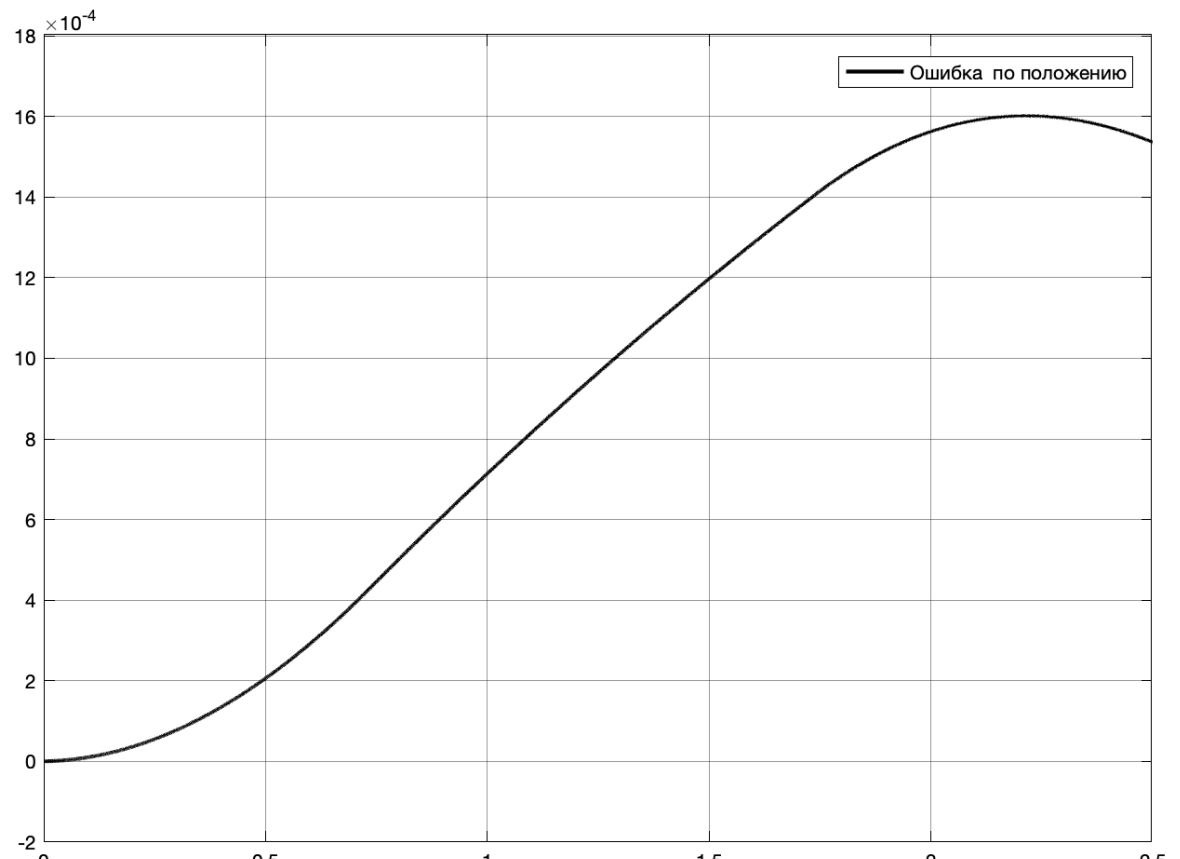

Рис.38 - График ошибки контура положения при заданном сигнале

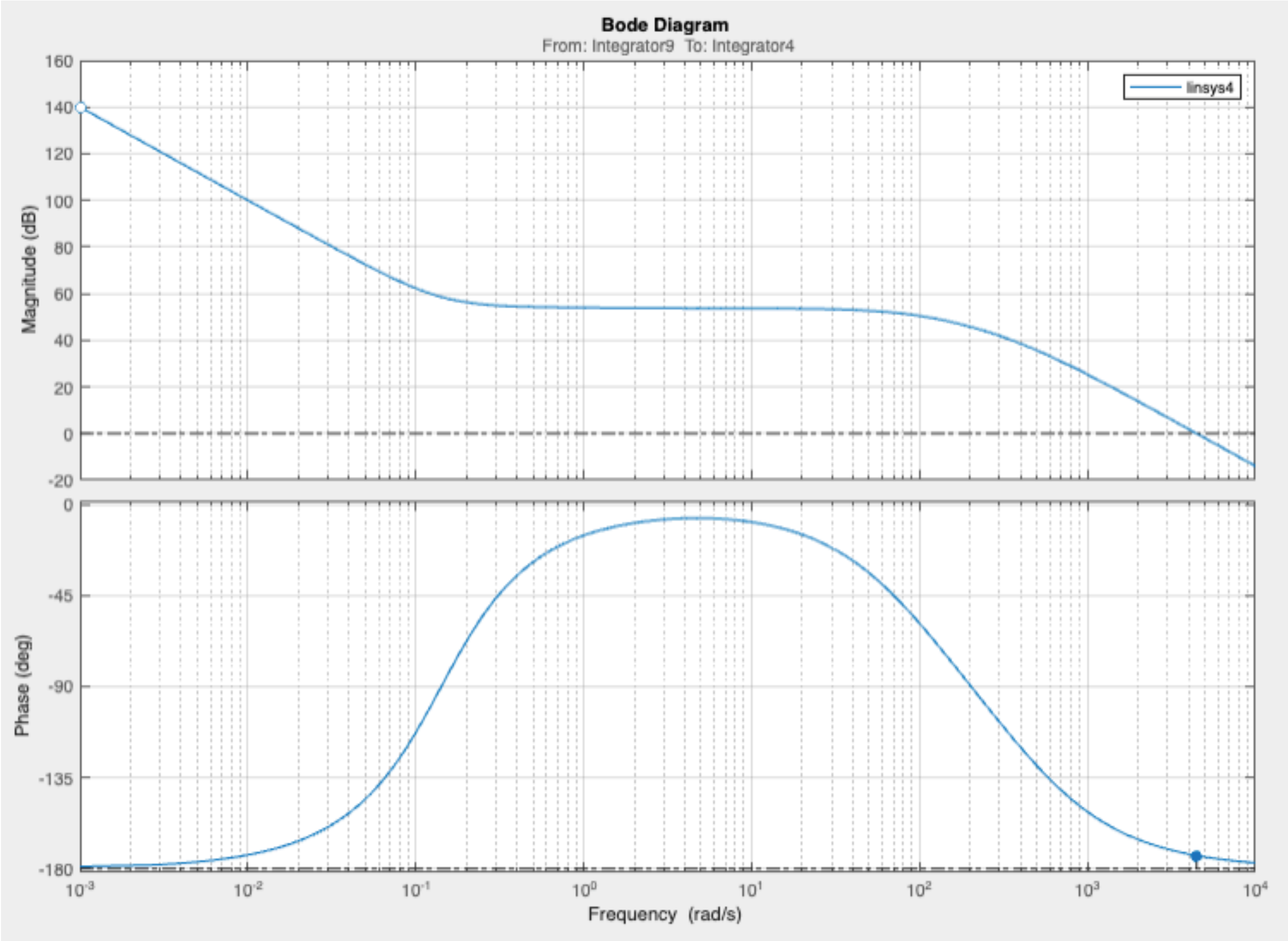

Рис.39 - ЛФЧХ

На практике было обнаружено, что ошибка положения двигателя довольно велика и не может быть достигнута такая же точность, как в моделировании. Анализ показал следующие причины:

**Трение и статическое трение:**

Внутреннее трение двигателя и трение в передающей системе (особенно статическое трение) могут привести к тому, что двигатель будет трудно запускаться или реагировать на малые углы поворота. Для преодоления статического трения требуется достаточно большая сила, и если входной сигнал недостаточен для этого, двигатель может не двигаться.

**Ограничение датчиков:**

Разрешение датчика может быть недостаточным для обнаружения очень малых угловых изменений. Если разрешение датчика низкое, фактическое изменение положения двигателя может не быть точно обнаружено и контролируемо. Два датчика Холла могут генерировать только два сигнала с фазовым сдвигом обычно в 90 градусов, что позволяет обнаруживать только четыре положения за один полный цикл ротора двигателя. Это может быть недостаточно для точного управления.

**Механические зазоры и люфт:**

Зазоры и люфт в механической системе могут вызывать значительные ошибки при изменении направления. Для малых угловых корректировок механические зазоры могут привести к замедленной и неточной реакции двигателя в реальной эксплуатации.

Таким образом, в данной ситуации мы заменим двигатель постоянного тока MG310 на шаговый двигатель 28BYJ048, чтобы повысить точность и упростить управление.

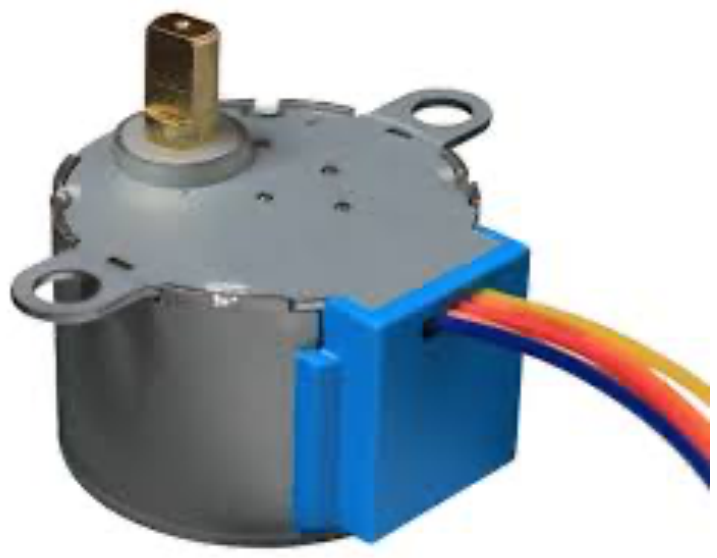

Рис.40 - 28BYJ048

# 4. Управление реальным роботом RoboBK

Управление реальным роботом RoboBK можно разделить на две категории:

- Система программного обеспечения (ROS)
- Система технического зрения (СТЗ)

На рис.41 указана логика управления реальным роботом.

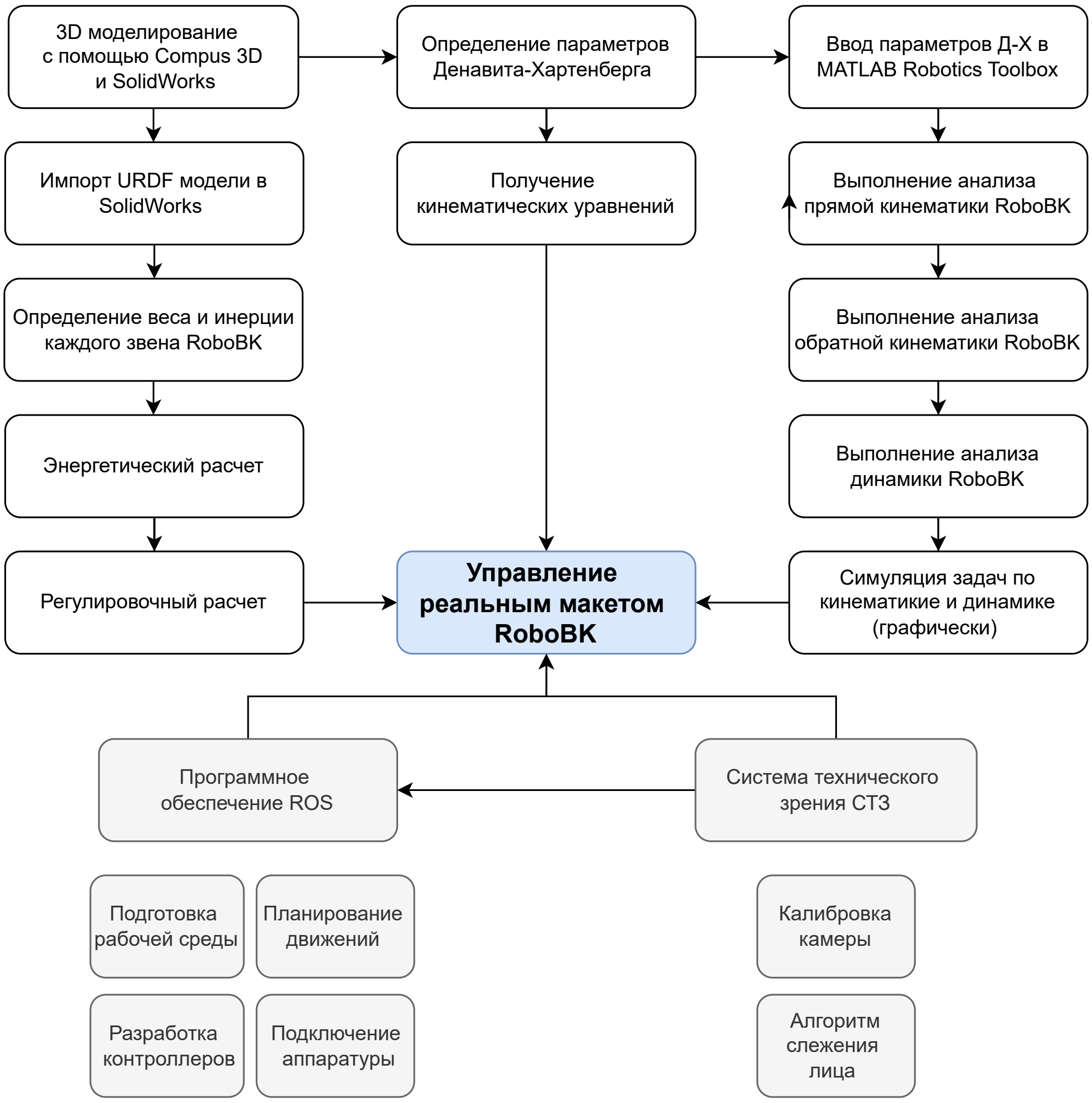

Рис.41 - Логика выполнения проекта

## 4.1 Система программного обеспечения (ROS)

### 4.1.1 Настройка системы и конфигурация

Robot Operating System (ROS) - это гибкая платформа для разработки программного обеспечения роботов. Это набор разнообразных инструментов, библиотек и определенных правил, целью которых является упрощение задач разработки ПО роботов. В нашем проекте RoboBK используется система ROS1.

Используемся файл URDF, созданный в SolidWorks, для генерации пакета функций MoveIt с помощью *MoveIt Setup Assistant*, включая создание матрицы самоколлизии в пакете.

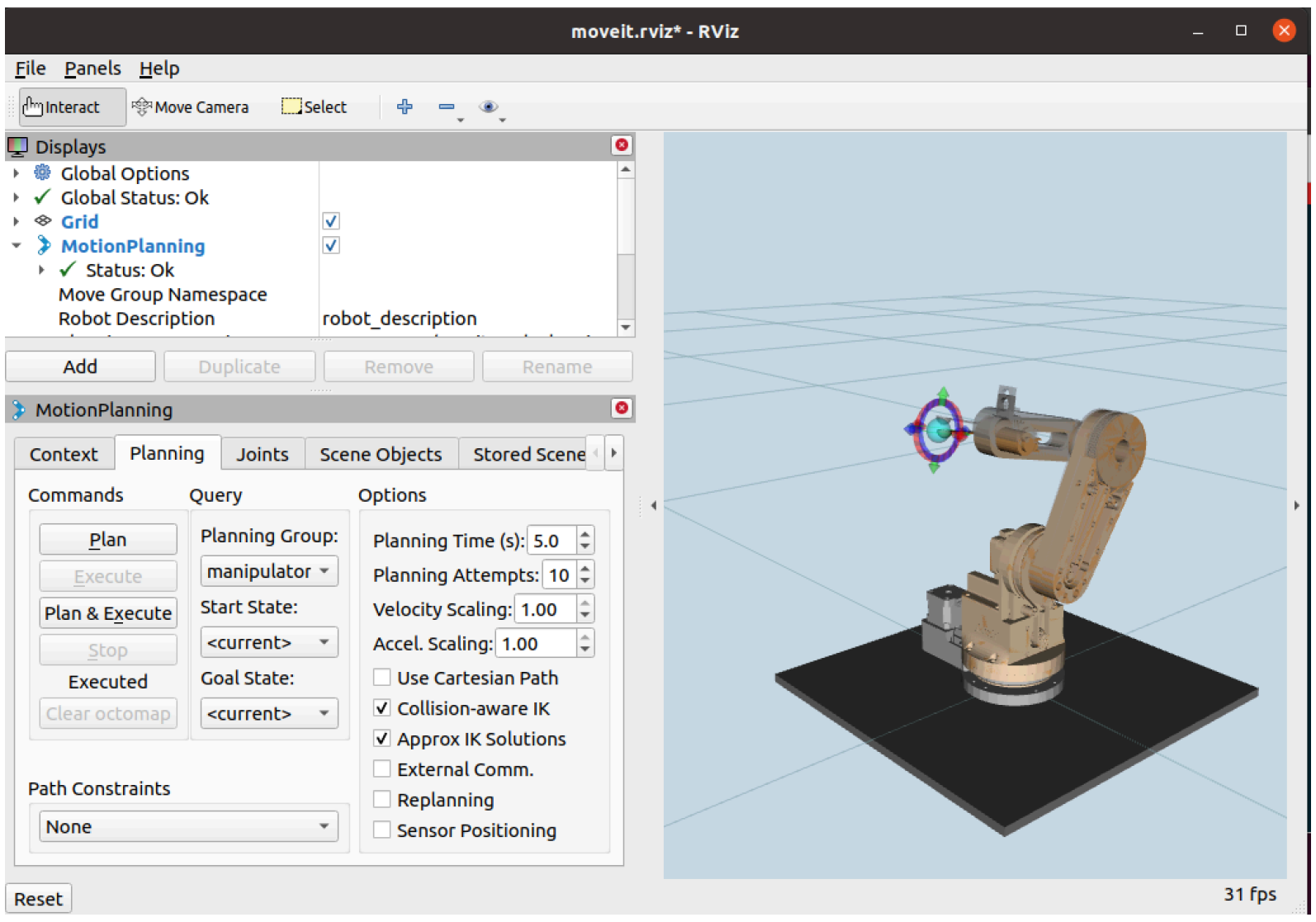

Рис.42 - UI Moveit

В управлении роботом термин "Состояние сочленений" (Joint State) обычно относится к текущему состоянию робота. Это состояние публикуется контроллером робота с целью информирования фреймворка MoveIt о текущем положении робота. Кроме того, публикатор состояния робота (Robot State Publisher) использует эту информацию для публикации фреймов tf, что позволяет rviz отображать текущую позицию робота в реальном времени.

Команда "Следовать по траектории сочленений" (Follow Joint Trajectory) служит для указания роботу, куда ему двигаться. Эта команда публикуется контроллером фиктивного робота в MoveIt.

### 4.1.2 Контроллеры и коммуникация

Существует два основных способа создания контроллера робота:

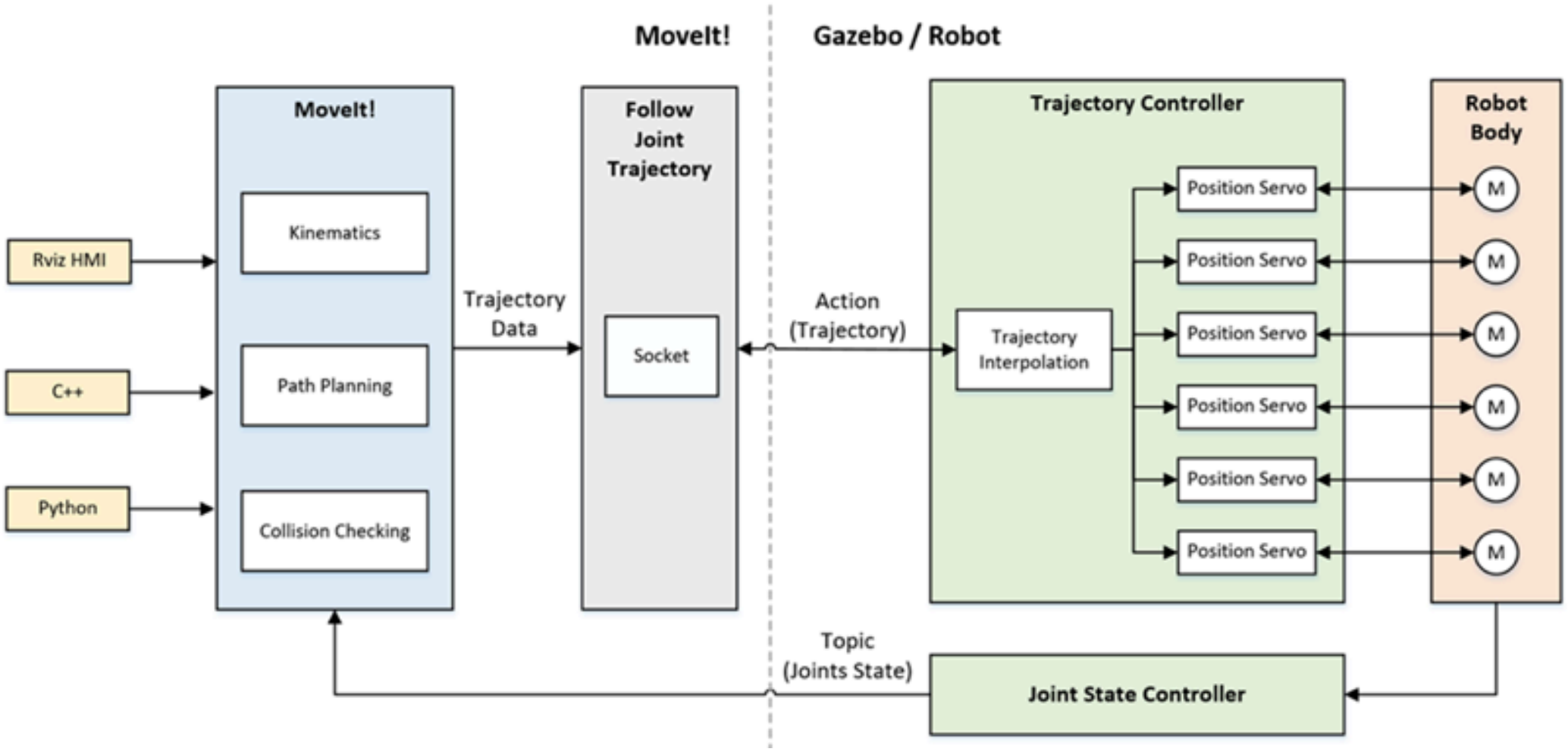

Рис.43 - Связи между MoveIt! и роботом

**Первый метод** включает написание контроллера робота, который получает сообщения FollowJointTrajectory и перемещает робота в указанное место. В то же время контроллер должен публиковать сообщения о состоянии сочленений, чтобы информировать систему о текущем положении робота. В этом случае MoveIt будет использовать фиктивный контроллер робота для публикации сообщений FollowJointTrajectory, а написанный контроллер будет подписываться на эти сообщения.

```
    ros::Subscriber joint_state_sub = nh.subscribe("/
move_group/fake_controller_joint_states", 50,
jointStateCallback);
    ros::Publisher joint_step_pub = nh.advertise
<robobk_moveit::ArmJointState>("/joint_steps", 50);
```

Листинг 1 - Subscriber и Publisher moveit_convert.cpp

Мы можем подписаться на тему в /move_group/joint_states_controller, чтобы получать информацию о текущем положении робота в MoveIt и публиковать сообщения на /joint_steps. Arduino подписывается на эти сообщения, тем самым управляя движением моторов. Преимущество этого метода заключается в том, что мы можем синхронизировать движение с планированием в MoveIt, но недостаток заключается в том, что при отсутствии энкодеров для обратной связи на шаговых двигателях, невозможно достичь непрерывного движения. Это связано с тем, что решение планирования движения состоит из множества точек (как показано на рисунке ниже), и, хотя можно уменьшить паузы с помощью интерполяционных вычислений, по сути, это всё равно множественные старты и остановки двигателя.

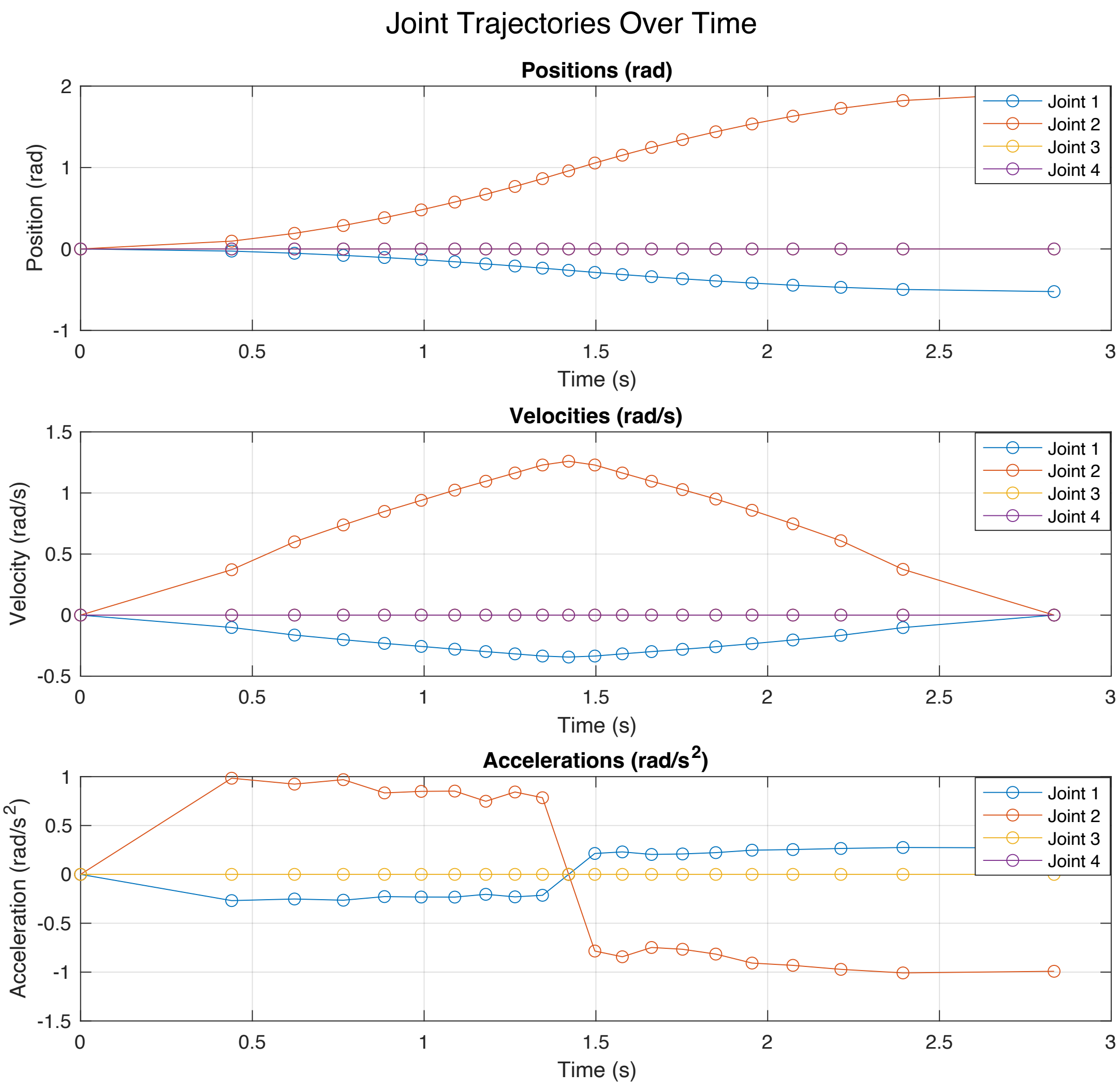

Рис.44 - Результат планирования траектории

Для решения проблемы заедания механического руки мы можем подписаться на display_planned_path в move_group, чтобы одномоментно получить начальные и конечные точки траектории. Функция обратного вызова displayTrajectoryCallback вычисляет необходимое количество шагов для каждого сустава для каждого движения и публикует их на /joint_steps. Хотя это решает проблему заедания, действия в MoveIt не синхронизируются в реальном времени.

```
    ros::Subscriber traj_sub=nh.subscribe("move_group/
display_planned_path",50, displayTrajectoryCallback);
    ros::Publisher joint_step_pub = nh.advertise
<robobk_moveit::ArmJointState>("/joint_steps", 50);
```

Листинг 2 - Subscriber и Publisher plan_transfer.cpp

**Второй метод** заключается в замене фиктивного контроллера робота в MoveIt на собственный плагин для управления роботом. Этот метод позволяет разработчикам взаимодействовать с контроллером посредством механизмов, отличных от сообщений ROS, или использовать другие сообщения ROS.

В нашем проекте используется первый метод.

**rqt_graph** в ROS предоставляет графическое представление вычислительной графики ROS. Этот график показывает узлы и темы управления RoboBK, которые они публикуют и на которые они подписываются, а также службы, предоставляемые каждым узлом (рис. 45).

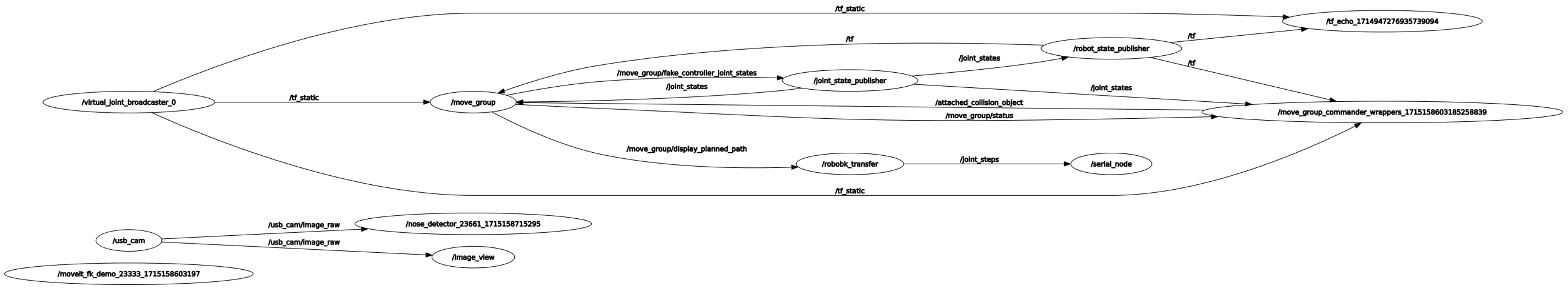

Рис.45 - rqt_graph

Узел */move_group* играет центральную роль в вычислительном графе, взаимодействуя со многими другими узлами.Взаимодействие включает подписку на несколько тем, таких как */move_group/fake_controller_joint_states, / joint_states* и */move_group/display_planned_path*, а также публикацию в темы, такие как */move_group/status*.

Узлы */robot_state_publisher* и */joint_state_publisher* участвуют в публикации состояния робота в тему */tf*, что важно для оценки и визуализации состояния робота.

Узел */serial_node* взаимодействует с узлом */robobk_transfer* через тему */ joint_steps.*

Узел /usb_cam обрабатывает ввод с камеры и публикует данные изображения для обработки другими узлами.

## 4.2 Система технического зрения (СТЗ)

Система технического зрения (СТЗ) является главным инструментом, наделяющим робота механизмом восприятия. Иногда камера устанавливается в фиксированном положении, например, лазерный дальномер, или камера расположена на концевом эффекторе.

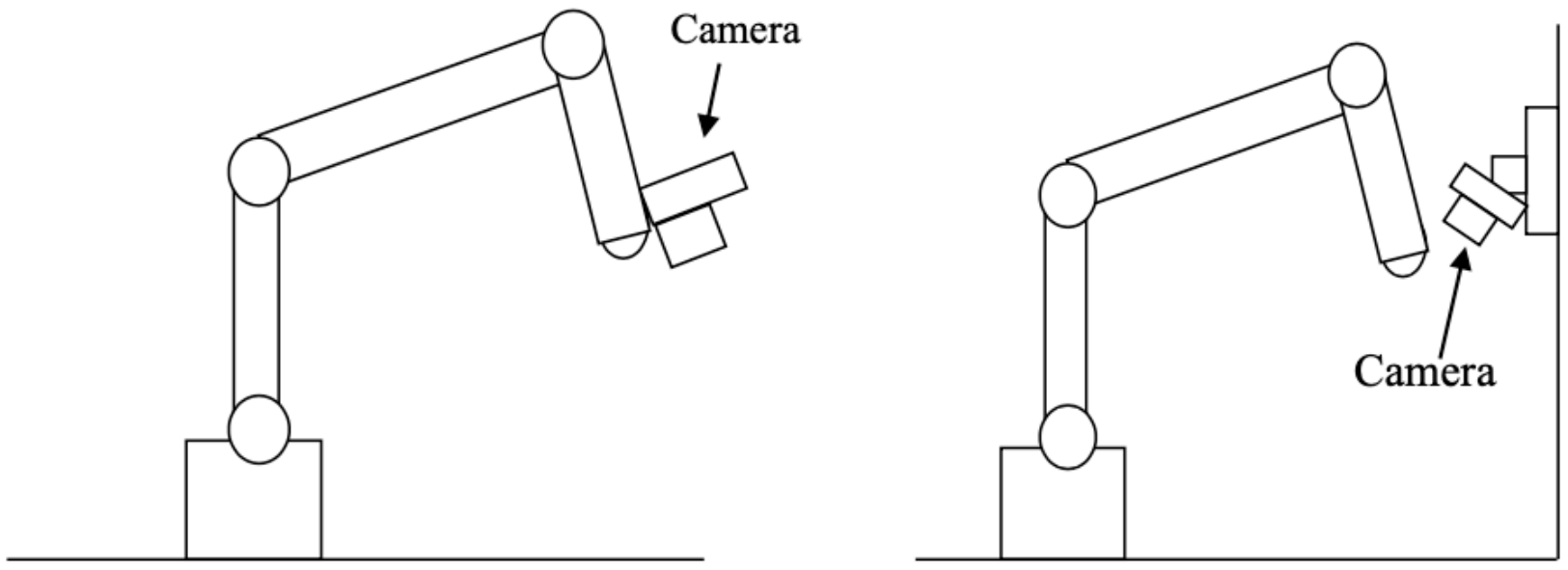

Рис.46 - Виды проставления камеры

Основной целью СТЗ является использование информации, полученной от датчиков зрения, для обратного контроля положения или движения робота. Визуальные датчики, такие как камеры, обеспечивают визуальный вход. Эти данные используются в качестве обратной связи, и различные особенности видео/изображения используются для оценки положения и движения манипулятора.

### 4.2.1 Калибровка камеры

В roboBK мы установили камеру на конце манипулятора. Для получения изображения используется цифровая камера (USB Camera for Raspberry Pi). Технические характеристики камеры приведены в таблице.

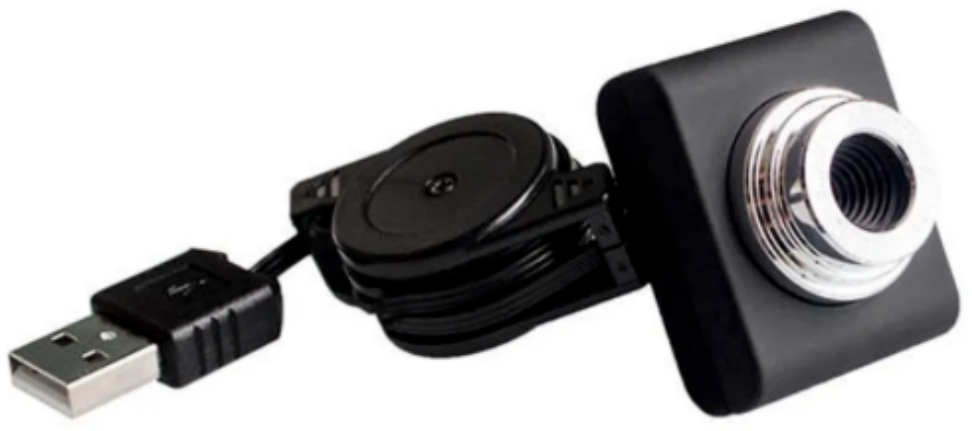

Рис.47 - USB Camera for Raspberry Pi

Таблица 9 - Технические характеристики камеры

| | |
|---|---|
| Фокусное расстояние объектива | F6.0MM |
| Диапазон фокусировки | от 20MM до бесконечности |
| Разрешение видео | 640*480 |

Калибровка камеры – это процесс определения внутренних (фокусное расстояние, центр проекции) и внешних (положение и ориентация) параметров камеры. Эти параметры необходимы для того, чтобы точно определить, как точки 3D пространства отображаются на 2D изображении.

**Калибровка для внутренних параметров:**

На рисунке представлены два основных типа радиальной дисторсии ( $k > 0$ и $k < 0$). Радиальное искажение проявляется в виде эффекта «бочка» или «рыбий глаз».

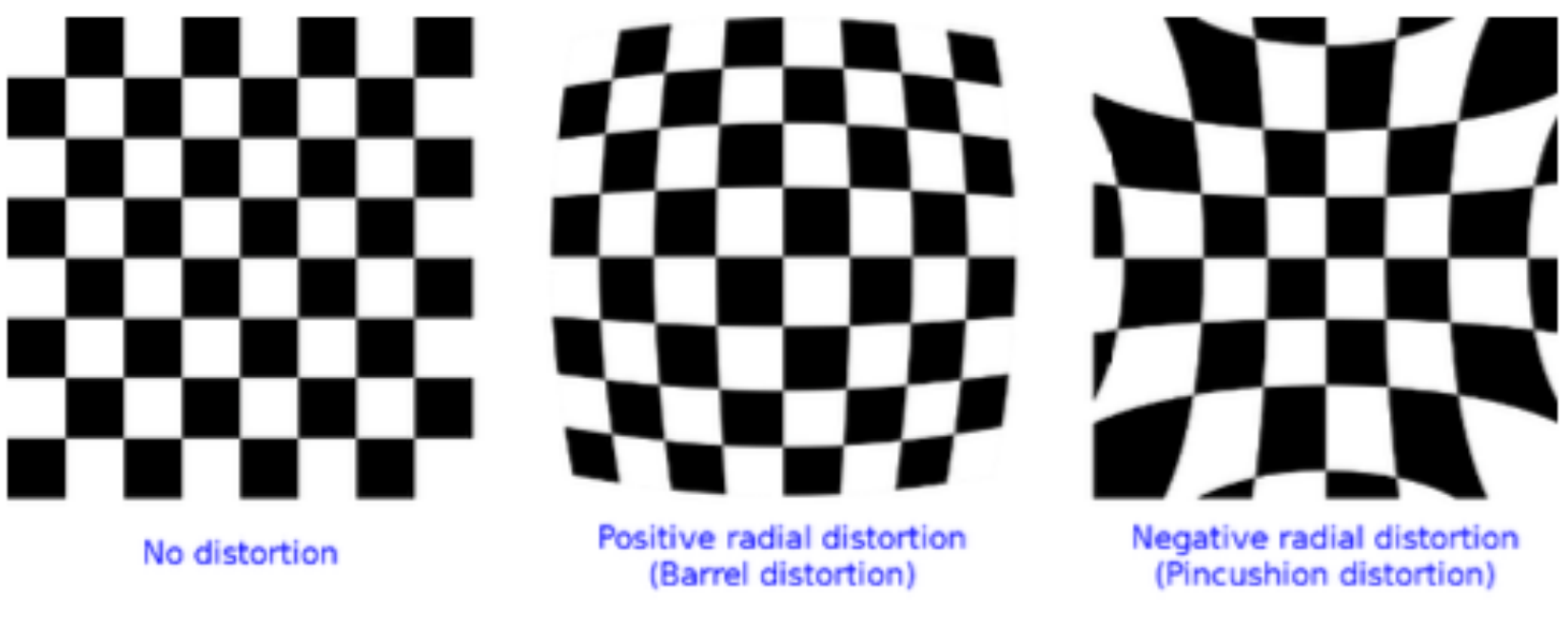

Рис.48 - Радиальная дисторсия

Для калибровки камеры необходимо иметь набор 3D точек и набор соответствующих 2D точек. Двухмерные точки легко находятся из изображения. Для нахождения трехмерных точек необходимо условие, что шахматная доска неподвижна в плоскости XY, координаты Z всех точек равны нулю, камера подвижна. Предположим, что длина стороны одной клетки равна единице, тогда трехмерные точки принимают значения (0, 0, 0), (0, 1, 0), (0, 2, 0), ... , (5, 7, 0), (5, 8, 0).

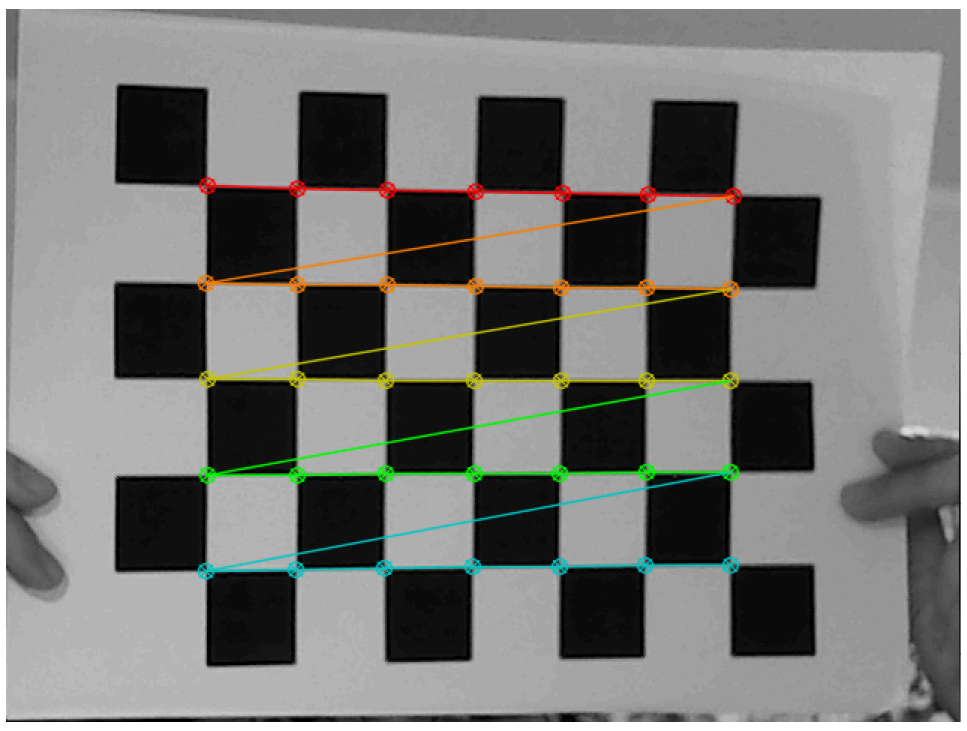

Рис.49 - Калибровка внутренних параметров

Далее будет производиться калибровка камеры с помощью встроенных функций в библиотеку OpenCV по отснятым шаблонам шахматной доски.

Результат калибровки внутренних параметров:

```
image_width: 640
image_height: 480
camera_name: narrow_stereo
camera_matrix:
  data: [1410.98768,    0.     ,  153.16333,
            0.     , 1411.54333,  312.17826,
            0.     ,    0.     ,    1.     ]
distortion_model: plumb_bob
distortion_coefficients:
  data: [-0.091805, 0.008574, 0.002489, -0.030940, 0.000000]
rectification_matrix:

  data: [1., 0., 0.,
         0., 1., 0.,
         0., 0., 1.]
projection_matrix:
  data: [1364.97693,    0.     ,  145.62117,    0.     ,
            0.     , 1410.63196,  312.19188,    0.     ,
            0.     ,    0.     ,    1.     ,    0.     ]
```

Рис.50 - Результат калибровки для внутренних параметров

**Калибровка для внешних параметров:**

В нашему случае, камера находится на манипуляторе. Цель калиброки внешних параметров - это есть установление соответствия между координатной системой камер и координатной системой манипулятора.

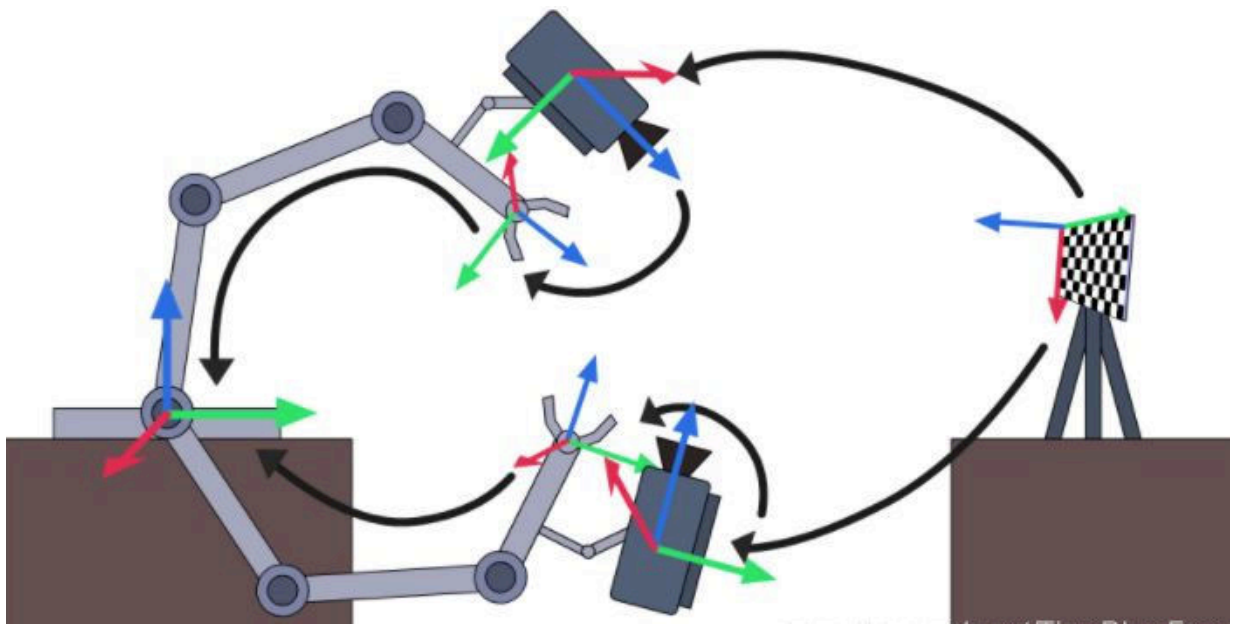

Рис.51 - Калибровка для внешних параметров

$$M_{\text{рука}}^{\text{камера}} = M_{\text{калибр}}^{\text{камера}} * M_{\text{база}}^{\text{калибр}} * M_{\text{рука}}^{\text{база}}$$

где

| | |
|---|---|
| $M_{\text{калибр}}^{\text{камера}}$ | Матрица, полученная во время фотометрической калибровки камеры, использует внутренние параметры и коэффициенты искажения. |
| $M_{\text{база}}^{\text{калибр}}$ | Матрица, которая описывает положение калибровочной пластины, которая закреплена и не перемещается во время калибровки, обеспечивая постоянство её параметров для всех изображений. |
| $M_{\text{рука}}^{\text{база}}$ | Матрица из параметров конечного положения манипулятора. |

Калибровка  внешних параметров выполнено с помощью ROS пакета - easy_handeye. Результат проставлен в нижнем картинке

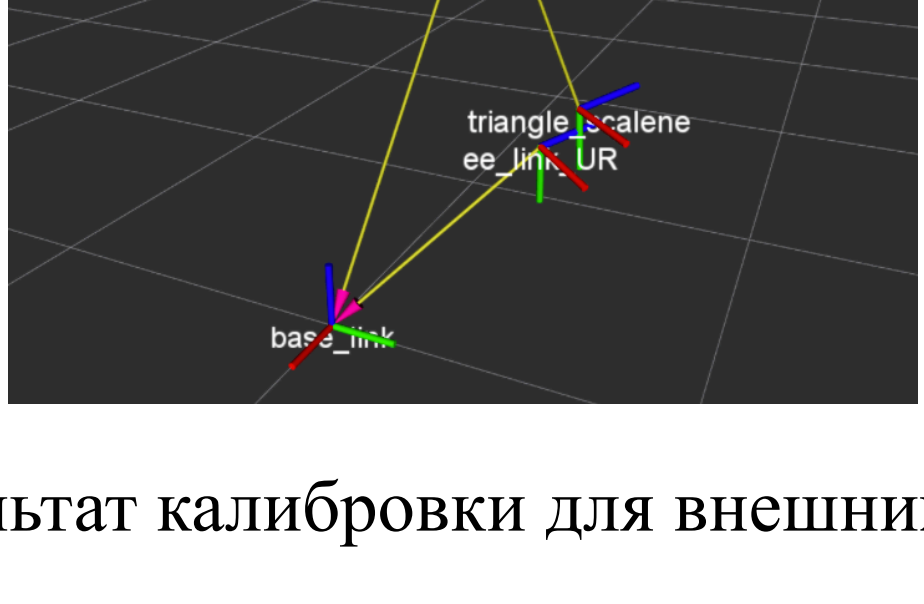

Рис.52 - Результат калибровки для внешних параметров

### 4.2.2 Алгоритм обнаружения лица

Используемся каскадные классификаторы Haar из библиотеки OpenCV для обнаружения носа на лице. Каскад Haar - это эффективный метод обнаружения объектов, который широко используется в обработке изображений в реальном времени. Эта технология основана на машинном обучении, где классификатор должен быть обучен с использованием большого количества положительных и отрицательных образцов изображений. Основной причиной выбора детекции носа является то, что, как показали практические тесты, детекция носа происходит быстрее и с большей успешностью, и из-за узкого поля зрения используемой камеры сложно захватить целое лицо в кадре.

На практике мы используем предварительно обученную модель каскада Хаара haarcascade_mcs_nose.xml для распознавания носа на изображении. Эта модель быстро определяет положение носа, анализируя характеристики яркости в изображении для определения потенциальных целевых областей. С помощью скользящего окна шаблон характеристик сопоставляется по всему изображению, и каскадные деревья решений подтверждают целевую область.

В практическом использовании ROS подписываемся на получение изображений Image_raw от USB-камеры, а также создаем сервис в ROS для публикации координат лица в реальном времени.

```
    self.image_sub = rospy.Subscriber("/usb_cam/
image_raw", Image, self.image_callback)
    self.service = rospy.Service('get_face_position',
FacePosition, self.handle_face_position)
    self.bridge = CvBridge()
    self.nose_cascade = cv2.CascadeClassifier('/home/
bingkun/robobk_ws/src/face_tracking/haarcascades/
haarcascade_mcs_nose.xml')
```

Листинг 3 - Алгоритм распознания face_detector.py

Для повышения точности и эффективности обнаружения мы настроили параметры scaleFactor и minNeighbors. Параметр scaleFactor определяет масштаб уменьшения размера изображения на каждом уровне масштаба, а параметр minNeighbors - количество соседних окон обнаружения, которые должны быть сохранены. Эти параметры в совокупности помогают уменьшить количество ошибочных обнаружений и повышают стабильность и надежность системы.

```
    gray = cv2.cvtColor(cv_image, cv2.COLOR_BGR2GRAY)
    noses = self.nose_cascade.detectMultiScale(gray,
scaleFactor=1.3, minNeighbors=8)
```

Листинг 4 - Параметры распознания face_detector.py

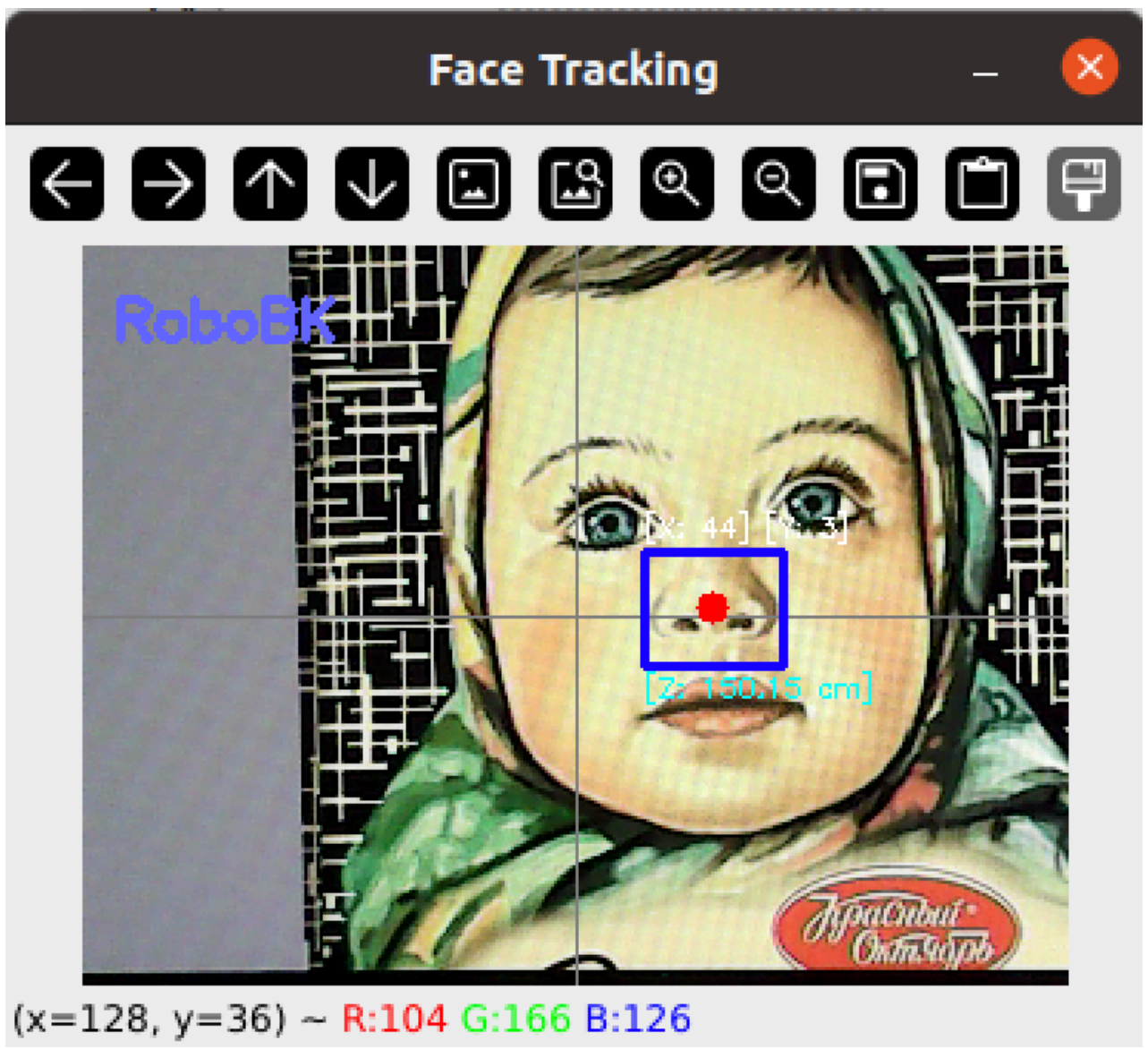

Рис.53 - UI face_detector.py

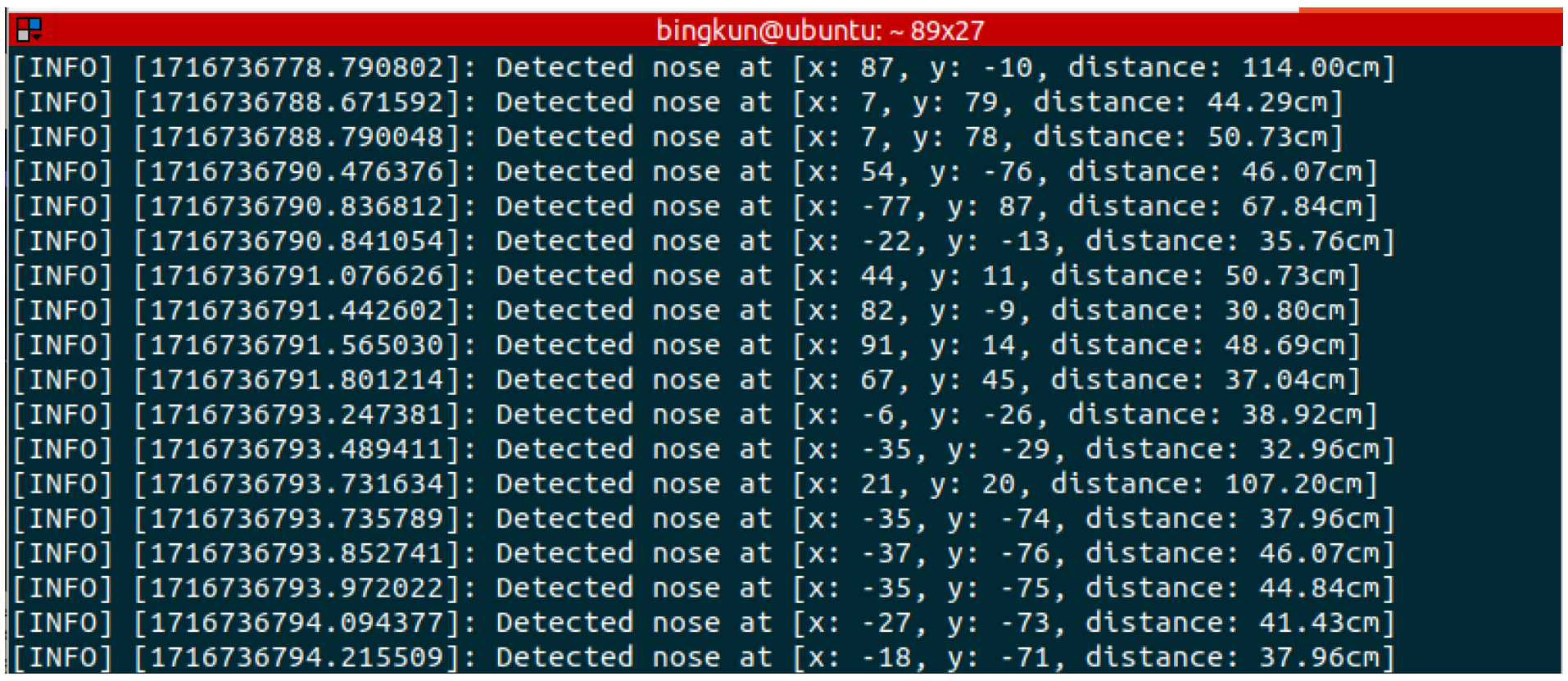
bingkun@ubuntu: ~ 89x27
[INFO] [1716736778.790802]: Detected nose at [x: 87, y: -10, distance: 114.00cm]
[INFO] [1716736788.671592]: Detected nose at [x: 7, y: 79, distance: 44.29cm]
[INFO] [1716736788.790048]: Detected nose at [x: 7, y: 78, distance: 50.73cm]
[INFO] [1716736790.476376]: Detected nose at [x: 54, y: -76, distance: 46.07cm]
[INFO] [1716736790.836812]: Detected nose at [x: -77, y: 87, distance: 67.84cm]
[INFO] [1716736790.841054]: Detected nose at [x: -22, y: -13, distance: 35.76cm]
[INFO] [1716736791.076626]: Detected nose at [x: 44, y: 11, distance: 50.73cm]
[INFO] [1716736791.442602]: Detected nose at [x: 82, y: -9, distance: 30.80cm]
[INFO] [1716736791.565030]: Detected nose at [x: 91, y: 14, distance: 48.69cm]
[INFO] [1716736791.801214]: Detected nose at [x: 67, y: 45, distance: 37.04cm]
[INFO] [1716736793.247381]: Detected nose at [x: -6, y: -26, distance: 38.92cm]
[INFO] [1716736793.489411]: Detected nose at [x: -35, y: -29, distance: 32.96cm]
[INFO] [1716736793.731634]: Detected nose at [x: 21, y: 20, distance: 107.20cm]
[INFO] [1716736793.735789]: Detected nose at [x: -35, y: -74, distance: 37.96cm]
[INFO] [1716736793.852741]: Detected nose at [x: -37, y: -76, distance: 46.07cm]
[INFO] [1716736793.972022]: Detected nose at [x: -35, y: -75, distance: 44.84cm]
[INFO] [1716736794.094377]: Detected nose at [x: -27, y: -73, distance: 41.43cm]
[INFO] [1716736794.215509]: Detected nose at [x: -18, y: -71, distance: 37.96cm]

Рис.54 - Координаты центра лица

## 4.2.3 Кинематическое управление с помощью СТЗ

В нашем проекте RoboBK существуют 2 типа управления с помощью СТЗ, которые можно классифицировать следующим образом: СТЗ на основе изображения и СТЗ на основе положения.

**СТЗ на основе изображения:** Он состоит из двух контуров управления(рис. 55). Один из них — это управление углами сочленений робота, а другой — контур управления на основе зрения. В контуре управления углами сочленений в качестве устройства обратной связи используются датчики углов сочленений, тогда как в контуре управления на основе зрения в качестве сигнала обратной связи используются особенности изображения. В этом методе ошибка вычисляется непосредственно на изображении извлечённых особенностей на плоскости 2D изображения, без прохождения через 3D реконструкцию. Робот должен двигаться так, чтобы привести текущие особенности изображения к их желаемым значениям. Особенностями этой схемы являются вычислительные преимущества. В этой схеме точность позиционирования системы менее чувствительна к ошибкам калибровки камеры и шумам измерений на изображениях.

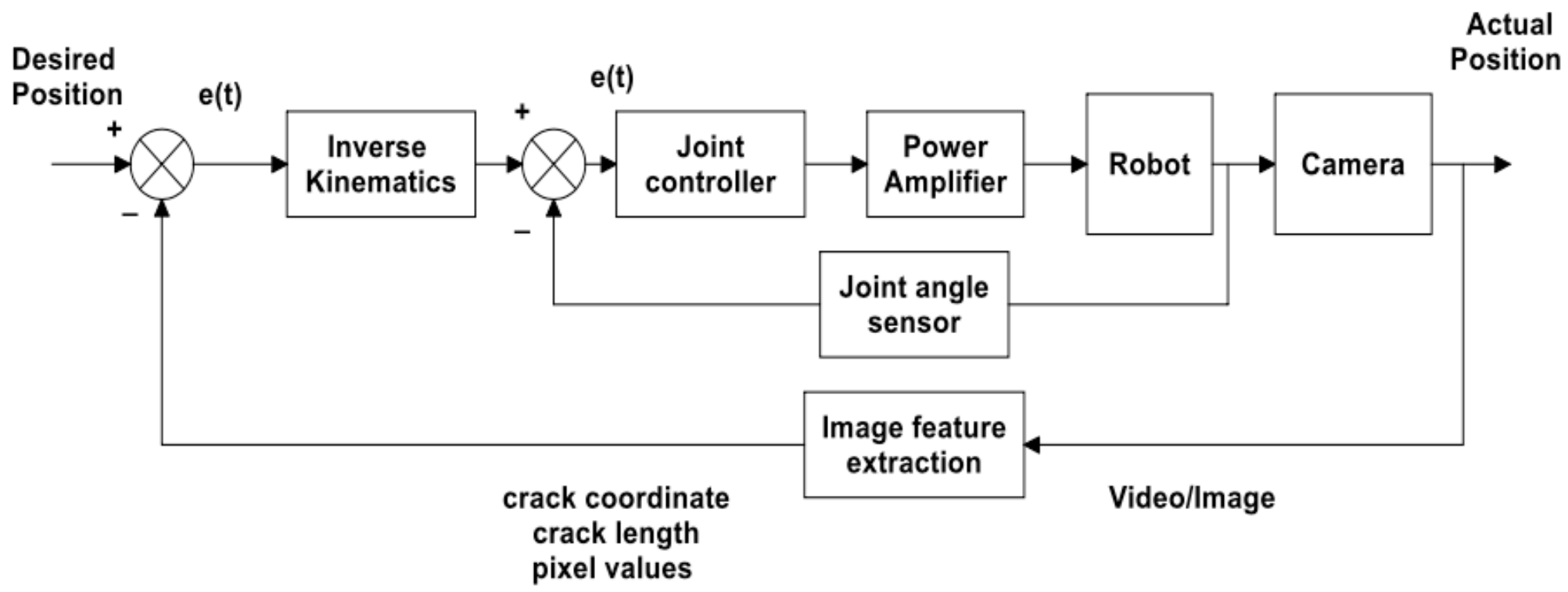

Рис.55 - СТЗ на основе изображения

Цель визуального сервоконтроля на основе изображений заключается в минимизации набора ошибок признаков $e(t)$, определенных в данных изображения как

$$e(t) = s(t) - s*$$

где $s(t)$ - это вектор визуальных признаков, меняющийся со временем,

$s^*$ - представляет желаемые значения признаков, которые можно считать постоянными, когда цель фиксирована. Изменения $s(t)$зависят как от движения камеры, так и от движения цели.

В нашем случае:

$$s(t) = \sqrt{x_{\text{offset}}^2 + y_{\text{offset}}^2}$$

$$s^* = 0$$

Если в течение 3 секунд, ошибка $e(t)$ не превышает 20 ( то есть центр носа находится внутри окружности радиуса 20 ), то цель считается успешно найденной (листинг 5).

```
x_offset = position.point.x
y_offset = -position.point.y
distance = math.sqrt(x_offset**2 + y_offset**2)
if distance <= self.radius_threshold:
   if self.stable_time_start is None:
      self.stable_time_start = time.time()
     elif  time.time()  -  self.stable_time_start  >=
self.stable_duration:
            self.move_to_target_position(position.point)
                return
        else:
             …
```

Листинг 5 - Проверка на наождения лица

**СТЗ на основе положения:** на основе положения управляет ошибкой между желаемыми и фактическими позами и работает непосредственно в 3D рабочем пространстве. Основное преимущество этого подхода заключается в том, что можно непосредственно контролировать положение концевого эффектора относительно цели. Основной недостаток заключается в том, что оценка положения и движения зависит от ошибок калибровки камеры и точности модели цели.

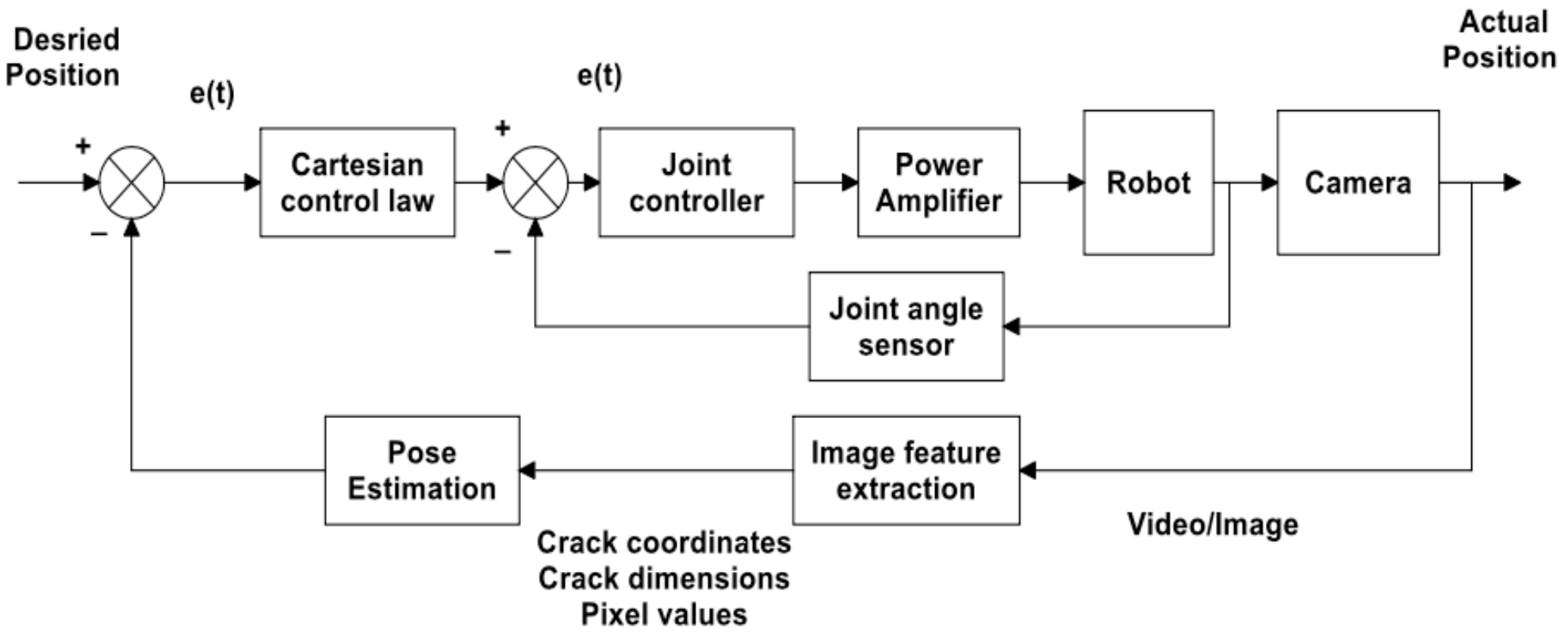

Рис.56 - СТЗ на основе положения

## 4.3 Алгоритм кормления

Сеть Петри, которая показывает общий алгоритм реализации.

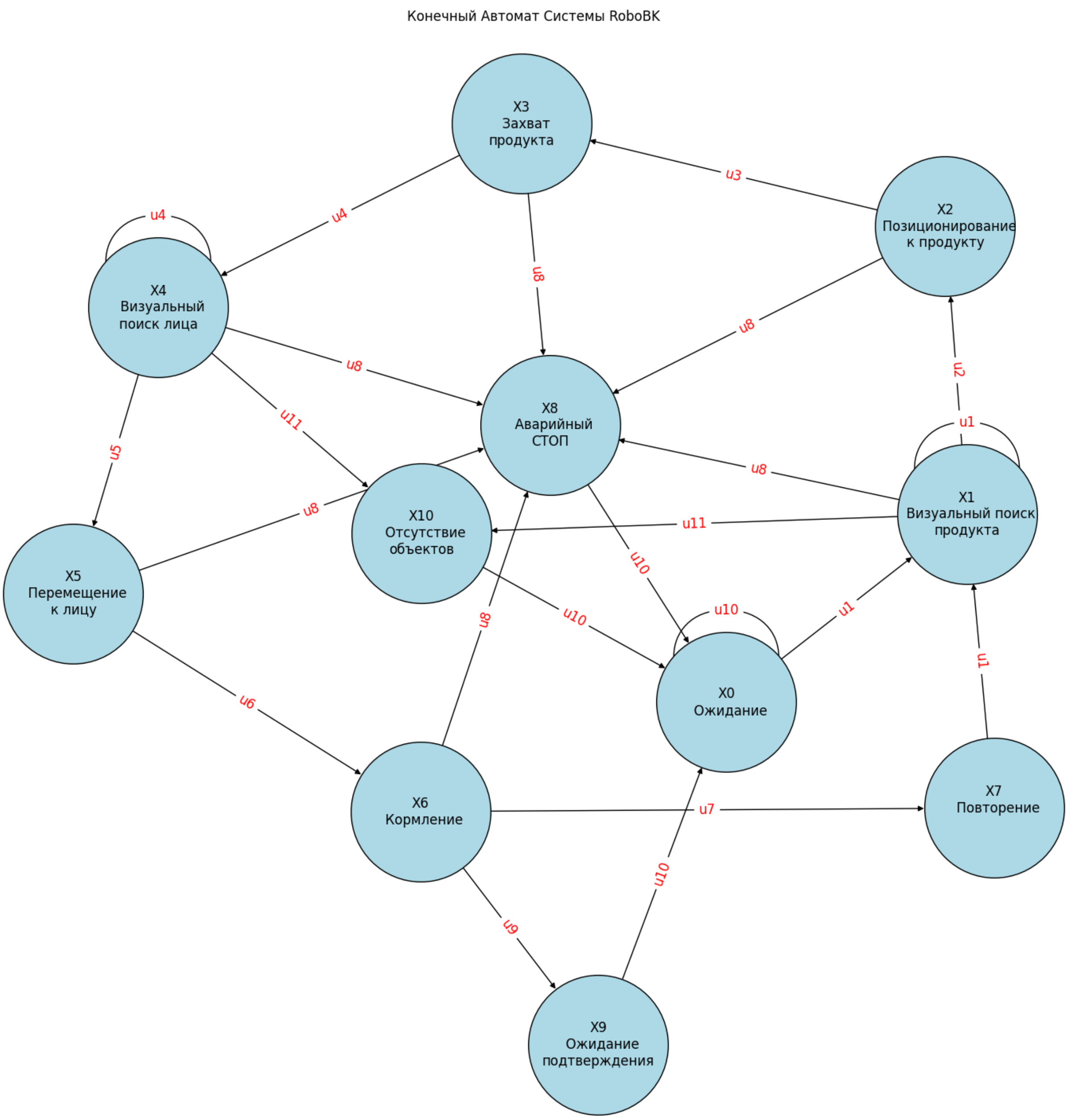

### Состояния

- X0 - Ожидание
- X1 - Визуальный поиск продукта
- X2 - Позиционирование к продукту
- X3 - Захват продукта
- X4 - Визуальный поиск лица

- X5 - Перемещение к лицу
- X6 - Кормление
- X7 - Повторение
- X8 - Аварийный СТОП
- X9 - Ожидание подтверждения
- X10 - Отсутствие объектов

**Входные сигналы:**

- u1 - Команда начала работы
- u2 - Сигнал завершения позиционирования
- u3 - Сигнал подтверждения захвата
- u4 - Визуальное позиционирование лица
- u5 - Подтверждение достижения места кормления
- u6 - Сигнал завершения кормления
- u7 - Сигнал повторного кормления
- u8 - Сигнал аварийной остановки
- u9 - Сигнал ожидания подтверждения
- u10 - Команда на паузу/остановку
- u11 - Сигнал неудачи поиска цели

Выходные сигналы для каждого из состояний соответствуют самому состоянию.

# ЗАКЛЮЧЕНИЕ

В ходе проделанной работы были получены следующие результаты:

- Проектирование и конструирование манипулятора
- Определение структуры системы управления
- Энергетический расчет манипулятора
- Кинематический расчет манипулятора
- Моделирование задач по кинематике и динамике
- Исследование модели привода конечного исполнителя
- Разработка системы технического зрения
- Разработка программного обеспечения в ROS
- Изготовление макета робота

В результате выполнения данной работы все поставленные задачи были решены. В данной работе разработан робот и система управления им, имеющая научное и практическое значение для людей с ограниченными возможностями.

# СПИСОК ИСПОЛЬЗОВАННЫХ ИСТОЧНИКОВ

# ПРИЛОЖЕНИЕ А

## Модель для решения прямой задачи по кинематике в среде Simscape

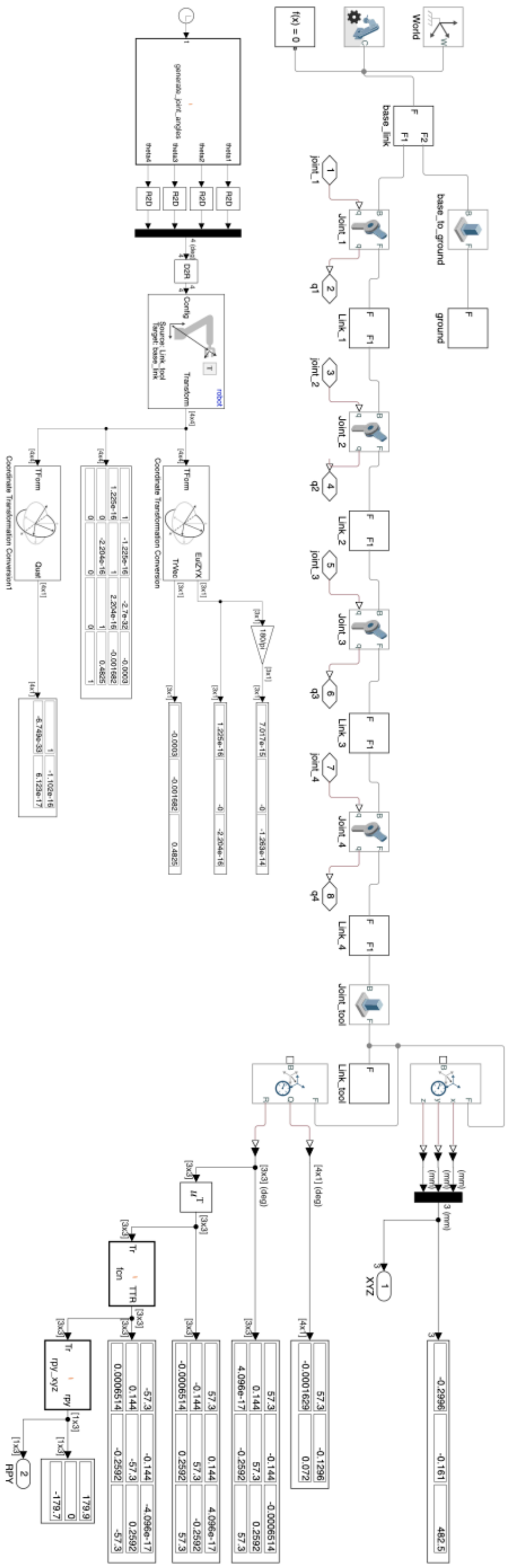

# ПРИЛОЖЕНИЕ Б

## Модель для решения обратной задачи по кинематике в среде Simscape

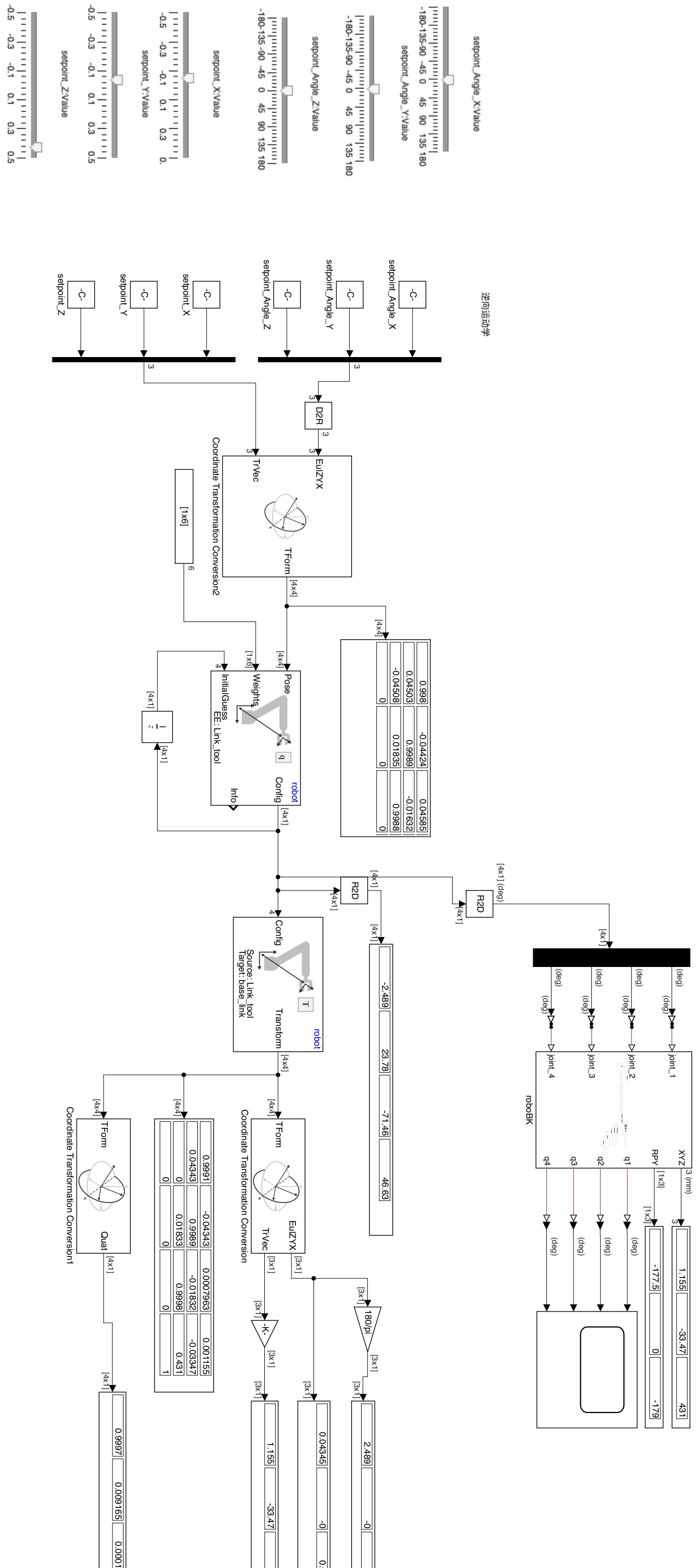

# ПРИЛОЖЕНИЕ В

## Модель для решения задач по динамике в среде Simscape

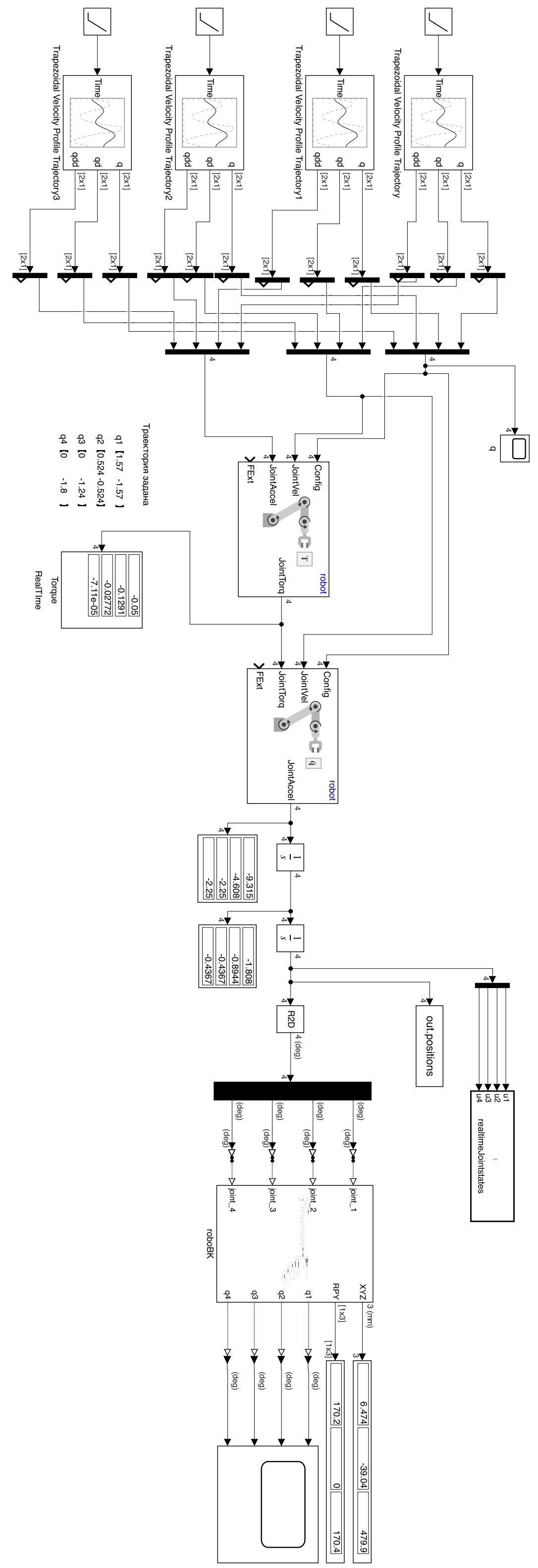

# ПРИЛОЖЕНИЕ Г

## Сборочный чертеж RoboBK

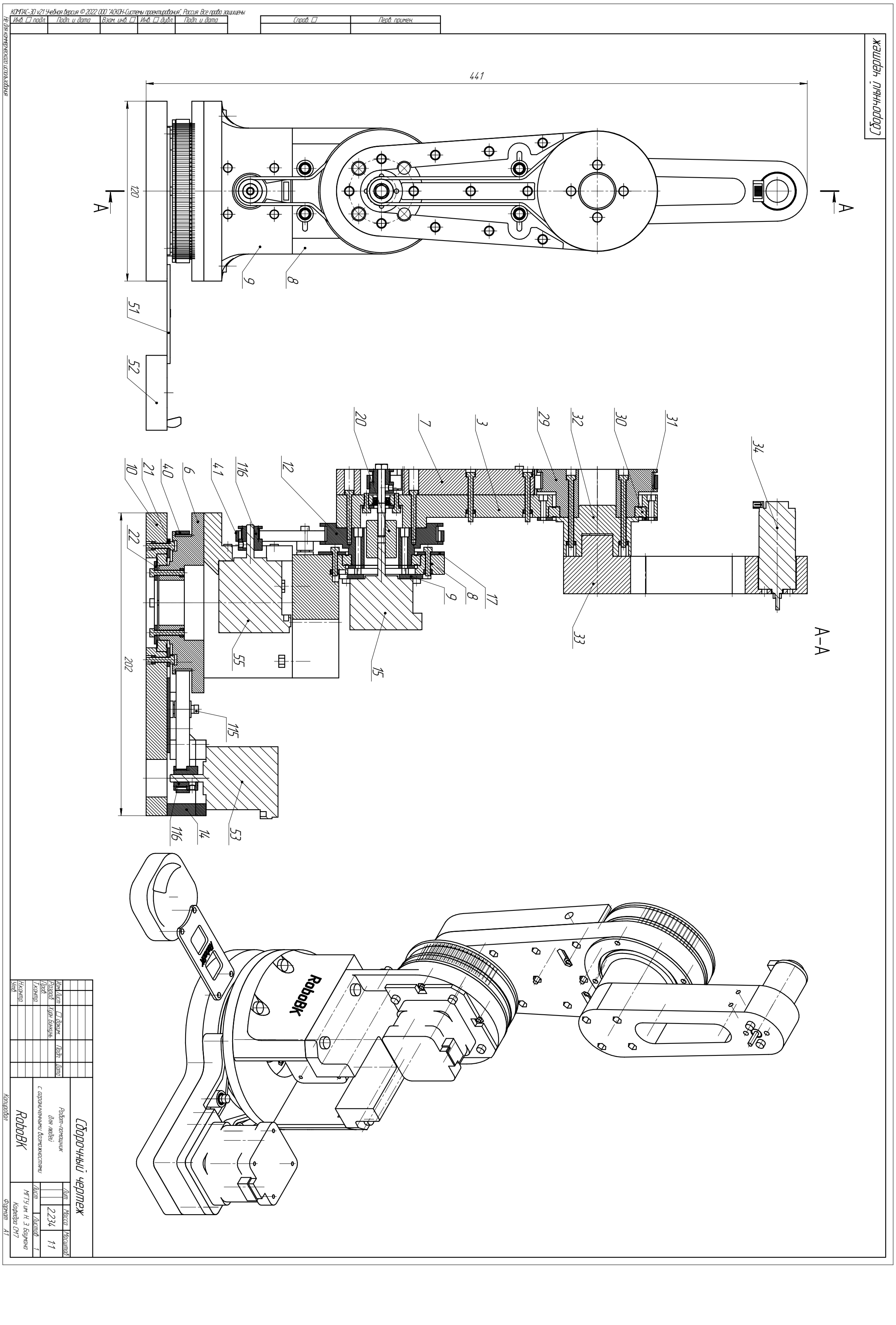

# ПРИЛОЖЕНИЕ Д

## Спецификация

| Формат | Зона | Поз. | Обозначение | Наименование | Кол. | Приме-чание |
|---|---|---|---|---|---|---|
| | | 14 | Base_Link_001 | Стойка мотора | | |
| | | 21 | Base_Link_002 | Верхняя крышка подшипника | | |
| | | 22 | Base_Link_003 | Нижняя крышка подшипника | | |
| | | 10 | Base_Link-004 | База | | |
| | | 2 | Base_Link_005 | Малая стойка | | |
| | | 6 | Link_1_001 | Шкив | | |
| | | 8 | Link_1_002 | Стойка | | |
| | | 25 | Link_1_003 | Внутреняя крышка подшипника | | |
| | | 17 | Link_1_004 | Внешняя крышка подшипника | | |
| | | 9 | Link_1_005 | | | |
| | | 12 | Link_2_001 | Шкив | | |
| | | 3 | Link_2_002 | Плечо | | |
| | | 7 | Link_2_003 | Крышка плеча | | |
| | | 29 | Link_3_001 | Шкив | | |
| | | 31 | Link_3_002 | Крышка подшипника | | |
| | | 32 | Link_3_003 | Шея | | |
| | | 33 | Link_3_004 | Плечо | | |
| | | 28 | Link_4_004 | Крышка подшипника | | |
| | | 116 | Ведущий шкив | Ведущий шкив | | |
| | | 40 | HTD3M-BELT_001 | Ремень-9mm-420mm | | |
| | | 41 | HTD3M-BELT_002 | Ремень-9mm-339mm | | |
| | | 42 | HTD3M-BELT_003 | Ремень-9mm-447mm | | |
| | | 23 | 61808-ZZ | 61808-ZZ | | |
| | | 30 | 61810-ZZ | 61810-ZZ | | |
| | | 37 | Винт M3*12 | Винт M3*12 | | |
| | | 112 | Винт M3*16 | Винт M3*16 | | |
| | | 39 | Винт M3*20 | Винт M3*20 | | |
| | | 113 | Винт M3*30 | Винт M3*30 | | |
| | | 112 | Винт M4*40 | Винт M4*40 | | |

Перв. примен. | Справ. № | Подп. и дата | Инв. № дубл. | Взам. инв. № | Подп. и дата | Инв. № подл.

RoboBK

Изм. | Лист | № докум. | Подп. | Дата

Разраб. Хуан Бинкунь
Пров.
Н.контр.
Утв.

Манипулятор-помощник для людей с ограниченными возможностями

Лит. | Лист 1 | Листов 2

МГТУ им. Н. Э. Баумана
Кафедра СМ7

Копировал    Формат A4

Рисунок Д.1 Спецификация RoboBK (лист 1)

| Формат | Зона | Поз. | Обозначение | Наименование | Кол. | Приме-чание |
|---|---|---|---|---|---|---|
| | | 38 | Гайка М3 | Гайка М3 | | |
| | | 114 | Гайка М4 | Гайка М4 | | |
| | | 115 | Затяжное устройство шкива | Затяжное устройство шкива | | |
| | | 1 | | Болт М5-6gx45 ГОСТ 7805-70 | | |
| | | 49 | | 5g spoon v1 | | |
| | | 53 | | Мотор1 | | |
| | | 52 | | Миска | | |
| | | 26 | | F685ZZ | | |
| | | 15 | | Мотор3 | | |
| | | 20 | | Муфта | | |
| | | 51 | | Стойка миски | | |
| | | 34 | | MG310 | | |
| | | 55 | | Мотор2 | | |

Инв. № подл. | Подп. и дата | Взам. инв. № | Инв. № дубл. | Подп. и дата

| Изм. | Лист | № докум. | Подп. | Дата | RoboBK | Лист 2 |
|---|---|---|---|---|---|---|

 Копировал Формат А4

Рисунок Д.2 Спецификация RoboBK (лист 2)

# ПРИЛОЖЕНИЕ Е

## robobk.urdf

```
<?xml version="1.0" encoding="utf-8" ?>
<!-- This URDF was automatically created by SolidWorks to URDF
Exporter! Originally created by Stephen Brawner
(brawner@gmail.com)
     Commit Version: 1.6.0-1-g15f4949  Build Version:
1.6.7594.29634
     For more information, please see http://wiki.ros.org/
sw_urdf_exporter -->
<robot name="robobk">
    <link name="ground">
        <visual>
            <geometry>
                <box size="0.4 0.4 0.01" />
            </geometry>
            <material name="">
                <color rgba="0.5 0.5 0.5 1" />
            </material>
        </visual>
        <collision>
            <geometry>
                <box size="10 10 0.01" />
            </geometry>
        </collision>
        <inertial>
            <mass value="0.1" />
            <origin xyz="0 0 0" rpy="0 0 0" />
            <inertia ixx="0.03" iyy="0.03" izz="0.03" ixy="0.0"
ixz="0.0" iyz="0.0" />
        </inertial>
    </link>
    <joint name="base_to_ground" type="fixed">
        <parent link="base_link" />
```

```
        <child link="ground" />
        <origin xyz="0 0 -0.005" />
    </joint>
    <link name="base_link">
        <inertial>
            <origin xyz="0.0468000592995429 -9.78453298031929E-07
0.0108831624523131" rpy="0 0 0" />
            <mass value="0.235394988200003" />
            <inertia ixx="0.00017792297819014"
ixy="2.57069719971704E-10" ixz="-4.4831076156236E-06"
iyy="0.000574466470315655" iyz="8.05097386419217E-10"
izz="0.000740732568412719" />
        </inertial>
        <visual>
            <origin xyz="0 0 0" rpy="0 0 0" />
            <geometry>
                <mesh filename="package://robobk/meshes/
base_link.STL" />
            </geometry>
            <material name="">
                <color rgba="0.752941176470588 0.752941176470588
0.752941176470588 1" />
            </material>
        </visual>
        <collision>
            <origin xyz="0 0 0" rpy="0 0 0" />
            <geometry>
                <mesh filename="package://robobk/meshes/
base_link.STL" />
            </geometry>
        </collision>
    </link>
    <link name="Link_1">
        <inertial>
```

```
            <origin xyz="-0.00448652892293219 6.12942803739949E-06
0.0437134740389649" rpy="0 0 0" />
            <mass value="0.680889778304849" />
            <inertia ixx="0.000694993880297736"
ixy="1.93545050745036E-07" ixz="3.12516172213851E-05"
iyy="0.000591458301361505" iyz="9.43125126491587E-08"
izz="0.000859356675763497" />
        </inertial>
        <visual>
            <origin xyz="0 0 0" rpy="0 0 0" />
            <geometry>
                <mesh filename="package://robobk/meshes/
Link_1.STL" />
            </geometry>
            <material name="">
                <color rgba="0.96078431372549 0.96078431372549
0.949019607843137 1" />
            </material>
        </visual>
        <collision>
            <origin xyz="0 0 0" rpy="0 0 0" />
            <geometry>
                <mesh filename="package://robobk/meshes/
Link_1.STL" />
            </geometry>
        </collision>
    </link>
    <joint name="Joint_1" type="revolute">
        <origin xyz="0 0 0.037" rpy="0 0 0" />
        <parent link="base_link" />
        <child link="Link_1" />
        <axis xyz="0 0 -1" />
        <limit lower="-3.14" upper="3.14" effort="100"
velocity="1" />
    </joint>
```

```
    <link name="Link_2">
        <inertial>
            <origin xyz="-0.0302698200579945 3.10319955295635E-06
0.0486298800043187" rpy="0 0 0" />
            <mass value="0.340007908318219" />
            <inertia ixx="0.000811506769730464"
ixy="-1.32608457566965E-09" ixz="9.69085594591264E-06"
iyy="0.000671912472769019" iyz="-2.18735494241919E-08"
izz="0.000158789300088672" />
        </inertial>
        <visual>
            <origin xyz="0 0 0" rpy="0 0 0" />
            <geometry>
                <mesh filename="package://robobk/meshes/
Link_2.STL" />
            </geometry>
            <material name="">
                <color rgba="0.847058823529412 0.847058823529412
0.847058823529412 1" />
            </material>
        </visual>
        <collision>
            <origin xyz="0 0 0" rpy="0 0 0" />
            <geometry>
                <mesh filename="package://robobk/meshes/
Link_2.STL" />
            </geometry>
        </collision>
    </link>
    <joint name="Joint_2" type="revolute">
        <origin xyz="-0.034 0 0.1185" rpy="0 0 0" />
        <parent link="Link_1" />
        <child link="Link_2" />
        <axis xyz="-1 0 0" />
```

```
        <limit lower="-3.14" upper="3.14" effort="100"
velocity="1" />
    </joint>
    <link name="Link_3">
        <inertial>
            <origin xyz="0.0381719053763429 -3.19914930681121E-06
0.0381144655868358" rpy="0 0 0" />
            <mass value="0.394126112012252" />
            <inertia ixx="0.000638101627459446"
ixy="7.71890824312977E-10" ixz="-9.99944821613126E-06"
iyy="0.000589789219214866" iyz="-1.19382836891704E-08"
izz="0.000111974819748886" />
        </inertial>
        <visual>
            <origin xyz="0 0 0" rpy="0 0 0" />
            <geometry>
                <mesh filename="package://robobk/meshes/
Link_3.STL" />
            </geometry>
            <material name="">
                <color rgba="0.705882352941176 0.705882352941176
0.705882352941176 1" />
            </material>
        </visual>
        <collision>
            <origin xyz="0 0 0" rpy="0 0 0" />
            <geometry>
                <mesh filename="package://robobk/meshes/
Link_3.STL" />
            </geometry>
        </collision>
    </link>
    <joint name="Joint_3" type="revolute">
        <origin xyz="-0.04 0 0.144" rpy="0 0 0" />
        <parent link="Link_2" />
```

```
        <child link="Link_3" />
        <axis xyz="-1 0 0" />
        <limit lower="-3.14" upper="3.14" effort="100"
velocity="1" />
    </joint>
    <link name="Link_4">
        <inertial>
            <origin xyz="0.00599999871668854 -0.000224549997759475
0.0273469778514014" rpy="0 0 0" />
            <mass value="0.00118182281041904" />
            <inertia ixx="2.75102561623165E-07"
ixy="1.44823186496416E-14" ixz="-2.96008751252037E-14"
iyy="2.8742558912867E-07" iyz="4.10830209999222E-09"
izz="1.69404305672287E-08" />
        </inertial>
        <visual>
            <origin xyz="0 0 0" rpy="0 0 0" />
            <geometry>
                <mesh filename="package://robobk/meshes/
Link_4.STL" />
            </geometry>
            <material name="">
                <color rgba="0.564705882352941 0.564705882352941
0.564705882352941 1" />
            </material>
        </visual>
        <collision>
            <origin xyz="0 0 0" rpy="0 0 0" />
            <geometry>
                <mesh filename="package://robobk/meshes/
Link_4.STL" />
            </geometry>
        </collision>
    </link>
    <joint name="Joint_4" type="revolute">
```

```
        <origin xyz="0.0677 0 0.12" rpy="0 0 0" />
        <parent link="Link_3" />
        <child link="Link_4" />
        <axis xyz="-1 0 0" />
        <limit lower="-3.14" upper="3.14" effort="100"
velocity="1" />
    </joint>
    <link name="Link_tool">
        <inertial>
            <origin xyz="3.636E-05 0.00011904 -1.149E-05" rpy="0 0
0" />
            <mass value="8.1812E-09" />
            <inertia ixx="5.1133E-17" ixy="1.1693E-48"
ixz="-7.4184E-65" iyy="5.1133E-17" iyz="2.059E-80"
izz="5.1133E-17" />
        </inertial>
        <visual>
            <origin xyz="0 0 0" rpy="0 0 0" />
            <geometry>
                <mesh filename="package://robobk/meshes/
Link_tool.STL" />
            </geometry>
            <material name="">
                <color rgba="0.56471 0.56471 0.56471 1" />
            </material>
        </visual>
        <collision>
            <origin xyz="0 0 0" rpy="0 0 0" />
            <geometry>
                <mesh filename="package://robobk/meshes/
Link_tool.STL" />
            </geometry>
        </collision>
    </link>
    <joint name="Joint_tool" type="fixed">
```

```
        <origin xyz="0.006 -0.0016819 0.06302" rpy="0 0 0" />
        <parent link="Link_4" />
        <child link="Link_tool" />
        <axis xyz="0 0 0" />
    </joint>
    <link name="Camera_Link">
        <inertial>
            <origin xyz="-1.5613E-16 -0.0052449 -0.015438" rpy="0
0 0" />
            <mass value="0.0026998" />
            <inertia ixx="2.1472E-07" ixy="1.8526E-21"
ixz="-1.2499E-21" iyy="2.7594E-07" iyz="-3.9771E-08"
izz="1.5735E-07" />
        </inertial>
        <visual>
            <origin xyz="0 0 0" rpy="0 0 0" />
            <geometry>
                <mesh filename="package://robobk/meshes/
Camera_Link.STL" />
            </geometry>
            <material name="">
                <color rgba="0.56471 0.56471 0.56471 1" />
            </material>
        </visual>
        <collision>
            <origin xyz="0 0 0" rpy="0 0 0" />
            <geometry>
                <mesh filename="package://robobk/meshes/
Camera_Link.STL" />
            </geometry>
        </collision>
    </link>
    <joint name="Joint_camera" type="fixed">
        <origin xyz="0.0555 -0.044098 0.11559" rpy="1.5498 0 0" />
        <parent link="Link_3" />
```

```
        <child link="Camera_Link" />
        <axis xyz="0 0 0" />
    </joint>
    <transmission name="trans_Joint_1">
        <type>transmission_interface/SimpleTransmission</type>
        <joint name="Joint_1">
            <hardwareInterface>hardware_interface/
PositionJointInterface</hardwareInterface>
        </joint>
        <actuator name="Joint_1_motor">
            <hardwareInterface>hardware_interface/
PositionJointInterface</hardwareInterface>
            <mechanicalReduction>1</mechanicalReduction>
        </actuator>
    </transmission>
    <transmission name="trans_Joint_2">
        <type>transmission_interface/SimpleTransmission</type>
        <joint name="Joint_2">
            <hardwareInterface>hardware_interface/
PositionJointInterface</hardwareInterface>
        </joint>
        <actuator name="Joint_2_motor">
            <hardwareInterface>hardware_interface/
PositionJointInterface</hardwareInterface>
            <mechanicalReduction>1</mechanicalReduction>
        </actuator>
    </transmission>
    <transmission name="trans_Joint_3">
        <type>transmission_interface/SimpleTransmission</type>
        <joint name="Joint_3">
            <hardwareInterface>hardware_interface/
PositionJointInterface</hardwareInterface>
        </joint>
        <actuator name="Joint_3_motor">
```

```
            <hardwareInterface>hardware_interface/
PositionJointInterface</hardwareInterface>
            <mechanicalReduction>1</mechanicalReduction>
        </actuator>
    </transmission>
    <transmission name="trans_Joint_4">
        <type>transmission_interface/SimpleTransmission</type>
        <joint name="Joint_4">
            <hardwareInterface>hardware_interface/
PositionJointInterface</hardwareInterface>
        </joint>
        <actuator name="Joint_4_motor">
            <hardwareInterface>hardware_interface/
PositionJointInterface</hardwareInterface>
            <mechanicalReduction>1</mechanicalReduction>
        </actuator>
    </transmission>
    <gazebo>
        <plugin name="gazebo_ros_control">
            <robotNamespace>/</robotNamespace>
        </plugin>
    </gazebo>
</robot>
```

# ПРИЛОЖЕНИЕ Ё

## moveit_convert.cpp

```
#include "ros/ros.h"
#include "sensor_msgs/JointState.h"
#include “robobk_moveit/ArmJointState.h"
#include "math.h"
struct JointSteps
{
    int steps[4] = {0, 0, 0, 0};
};
JointSteps current_steps;
JointSteps accumulated_steps;
int steps_per_revolution[4] = {4800, 4000, 4000, 1};
double prev_angle[4] = {0, 0, 0, 0};
int count = 0;
int angleToSteps(double angle, int joint_index) {
    return static_cast<int>(round(angle *
steps_per_revolution[joint_index] / (2.0 * M_PI)));
}
void jointStateCallback(const sensor_msgs::JointState&
joint_state_msg) {
    if (count == 0) {
        for (size_t i = 0; i < joint_state_msg.name.size() && i <
4; ++i) {
            prev_angle[i] = joint_state_msg.position[i];
        }
    }
    for (size_t i = 0; i < joint_state_msg.name.size() && i < 4; +
+i) {
        double delta_angle = joint_state_msg.position[i] -
prev_angle[i];
        if (i == 1) {
```

```
        double compensation_angle = delta_angle * 0.2;
int compensation_steps = angleToSteps(compensation_angle, 2);
accumulated_steps.steps[2] += compensation_steps;
        }
        current_steps.steps[i] = angleToSteps(delta_angle, i);
        accumulated_steps.steps[i] += current_steps.steps[i];
        prev_angle[i] = joint_state_msg.position[i];
    }
    ROS_WARN("Converted joint states to steps.");
    ROS_INFO("Step[0] =  %d",accumulated_steps.steps[0] );
    ROS_INFO("Step[1] =  %d",accumulated_steps.steps[1] );
    ROS_INFO("Step[2] =  %d",accumulated_steps.steps[2] );
    ROS_INFO("position[0] =  %f",joint_state_msg.position[0] );
    ROS_INFO("position[1] =  %f",joint_state_msg.position[1] );
    ROS_INFO("position[2] =  %f",joint_state_msg.position[2] );
    count++;
}
int main(int argc, char **argv) {
    ROS_WARN("Moveit_convert is running");
    ros::init(argc, argv, "robobk_convert");
    ros::NodeHandle nh;
    ros::Subscriber joint_state_sub = nh.subscribe("/move_group/
fake_controller_joint_states", 50, jointStateCallback);
    ros::Publisher joint_step_pub =
nh.advertise<robobk_moveit::ArmJointState>("/joint_steps", 50);
    ros::Rate loop_rate(1);
    while (ros::ok()) {
        robobk_moveit::ArmJointState joint_step_msg;
        joint_step_msg.position1 = accumulated_steps.steps[0];
        joint_step_msg.position2 = accumulated_steps.steps[1];
        joint_step_msg.position3 = accumulated_steps.steps[2];
        joint_step_msg.position4 = accumulated_steps.steps[3];
     // ROS_INFO("---publishposition[1] =
%d",joint_step_msg.position1 );
        joint_step_pub.publish(joint_step_msg);
```

```
        ros::spinOnce();
        loop_rate.sleep();
    }
    return 0;
}
```

# ПРИЛОЖЕНИЕ Ж

## plan_transfer.cpp

```
#include "ros/ros.h"
#include "moveit_msgs/DisplayTrajectory.h"
#include "robobk_moveit/ArmJointState.h"
#include "math.h"
struct JointSteps {
    int steps[4] = {0, 0, 0, 0};};
JointSteps joint_steps;
int steps_per_revolution[4] = {4800, 4000, 4000, 2048};
int command_id = 0 ;
int angleToSteps(double angle, int joint_index)
{
    return static_cast<int>(round(angle *
steps_per_revolution[joint_index] / (2.0 * M_PI)));
}
void displayTrajectoryCallback(const
moveit_msgs::DisplayTrajectory& display_trajectory_msg)
 {
    if (display_trajectory_msg.trajectory.empty() ||
display_trajectory_msg.trajectory[0].joint_trajectory.points.empty
())
     {
        ROS_WARN("Received empty trajectory or joint_trajectory
points.");
        return;
     }
    auto& trajectory =
display_trajectory_msg.trajectory[0].joint_trajectory;
    auto& start_point = trajectory.points.front();
    auto& end_point = trajectory.points.back();
    command_id = command_id + 1 ;
    for (size_t i = 0; i < trajectory.joint_names.size() && i < 4;
++i){
```

```
          double start_angle = start_point.positions[i];
          double end_angle = end_point.positions[i];
          double delta_angle = end_angle - start_angle;
            joint_steps.steps[i] = angleToSteps(delta_angle, i);
     ROS_WARN("Needed Steps:");
     ROS_INFO("Joint1 steps: %d", joint_steps.steps[0]);
     ROS_INFO("Joint2 steps: %d", joint_steps.steps[1]);
     ROS_INFO("Joint3 steps: %d",
int(joint_steps.steps[2]+joint_steps.steps[1]*0.2));
     ROS_INFO("Joint4 steps: %d", joint_steps.steps[3]);
      }
}
int main(int argc, char **argv){
    ROS_WARN("Moveit_transfer is running");
    ros::init(argc, argv, "robobk_transfer");
    ros::NodeHandle nh;
    ros::Subscriber traj_sub = nh.subscribe("/move_group/
display_planned_path", 50, displayTrajectoryCallback);
    ros::Publisher joint_step_pub =
nh.advertise<robobk_moveit::ArmJointState>("/joint_steps", 50);
    ros::Rate loop_rate(20);
    while (ros::ok()) {
        robobk_moveit::ArmJointState joint_step_msg;)
        joint_step_msg.position1 = joint_steps.steps[0];
        joint_step_msg.position2 = joint_steps.steps[1];
        joint_step_msg.position3 =
joint_steps.steps[2]+joint_steps.steps[1]*0.2;
        joint_step_msg.position4 = joint_steps.steps[3];
        joint_step_msg.command_id = command_id;
        joint_step_pub.publish(joint_step_msg);
       ros::spinOnce();
        loop_rate.sleep();
    }
    return 0;}
```

# ПРИЛОЖЕНИЕ 3

## face_detector.py

```
import rospy
import cv2
from cv_bridge import CvBridge, CvBridgeError
from sensor_msgs.msg import Image
from geometry_msgs.msg import PointStamped
from face_tracking.srv import FacePosition, FacePositionResponse
class NoseDetector:
    def __init__(self):
        rospy.init_node('nose_detector', anonymous=True)
        self.image_sub = rospy.Subscriber("/usb_cam/image_raw",
Image, self.image_callback)
        self.service = rospy.Service('get_face_position',
FacePosition, self.handle_face_position)
        self.bridge = CvBridge()
        self.nose_cascade = cv2.CascadeClassifier('/home/bingkun/
robobk_ws/src/face_tracking/haarcascades/
haarcascade_mcs_nose.xml')
        self.last_point = PointStamped()  # Initialize with an
empty PointStamped
    def image_callback(self, data):
        try:
            cv_image = self.bridge.imgmsg_to_cv2(data, "bgr8")
        except CvBridgeError as e:
            rospy.logerr(e)
            return

        height, width = cv_image.shape[:2]
        center_x, center_y = width // 2, height // 2

        cv2.line(cv_image, (center_x, 0), (center_x, height),
(128, 128, 128), 1)
```

```
        cv2.line(cv_image, (0, center_y), (width, center_y), (128, 
128, 128), 1)

        cv2.putText(cv_image, "RoboBK", (10, 30), 
cv2.FONT_HERSHEY_SIMPLEX, 0.6, (255, 100, 100), 2)

        gray = cv2.cvtColor(cv_image, cv2.COLOR_BGR2GRAY)
        noses = self.nose_cascade.detectMultiScale(gray, 
scaleFactor=1.3, minNeighbors=8)
        found = False
        for (nx, ny, nw, nh) in noses:
            found = True
            nose_center_x = nx + nw // 2
            nose_center_y = ny + nh // 2
            transformed_x = nose_center_x - center_x
            transformed_y = center_y - nose_center_y
            distance_estimate = 250000 / (nw * nh)  # Simple 
distance estimation

            self.last_point.header.stamp = rospy.Time.now()
            self.last_point.header.frame_id = "camera_link"
            self.last_point.point.x = transformed_x
            self.last_point.point.y = transformed_y
            self.last_point.point.z = distance_estimate

            rospy.loginfo(f"Detected nose at [x: {transformed_x}, 
y: {transformed_y}, distance: {distance_estimate:.2f}cm]")
            cv2.rectangle(cv_image, (nx, ny), (nx + nw, ny + nh), 
(255, 0, 0), 2)
            cv2.circle(cv_image, (nose_center_x, nose_center_y), 
5, (0, 0, 255), -1)  # Mark the nose center
            cv2.putText(cv_image, f"[X: {transformed_x}] [Y: 
{transformed_y}]", (nx, ny - 5), cv2.FONT_HERSHEY_SIMPLEX, 0.3, 
(255, 255, 255), 1)
```

```
            cv2.putText(cv_image, f"[Z: {distance_estimate:.2f}
cm]", (nx, ny + nh + 10), cv2.FONT_HERSHEY_SIMPLEX, 0.3, (255,
255, 0), 1)

        if not found:
            self.last_point.point.x = 0
            self.last_point.point.y = 0
            self.last_point.point.z = 0

        cv2.imshow("Face Tracking", cv_image)
        cv2.waitKey(3)

    def handle_face_position(self, request):
        if self.last_point.point.x != 0 or self.last_point.point.y
!= 0 or self.last_point.point.z != 0:
            return FacePositionResponse(True, "Position found",
self.last_point)
        else:
            return FacePositionResponse(False, "No position
detected", self.last_point)

if __name__ == '__main__':
    nd = NoseDetector()
    rospy.spin()
```

# ПРИЛОЖЕНИЕ И

## moveit_demo.py

```
import sys
import rospy
import moveit_commander
import math
import time
from geometry_msgs.msg import PointStamped, PoseStamped,
Quaternion, Pose
from face_tracking.srv import FacePosition
import tf2_ros
import tf2_geometry_msgs
from tf.transformations import quaternion_from_euler

class MoveItFkDemo:
    def __init__(self):
        moveit_commander.roscpp_initialize(sys.argv)
        rospy.init_node('moveit_fk_demo', anonymous=True)

        self.robot = moveit_commander.RobotCommander()
        self.arm =
moveit_commander.MoveGroupCommander('manipulator')
        self.arm.set_goal_joint_tolerance(0.001)
        self.arm.set_max_acceleration_scaling_factor(0.1)
        self.arm.set_max_velocity_scaling_factor(0.1)

        self.rotation_factor = 5  # degrees
        self.rotation_direction = -1  # -1 for counterclockwise, 1
for clockwise
        self.face_detected = False  # Initialize the face_detected
attribute here

        self.stable_time_start = None
        self.stable_duration = 3.0
```

```
        self.radius_threshold = 20.0

        self.arm.set_named_target('UP')
        self.arm.go()
        rospy.sleep(3)
        self.move_to_detect()
        rospy.sleep(5)
        self.arm.set_named_target('UP')
        self.arm.go()

    def move_to_detect(self):
        self.arm.set_named_target('detect')
        self.arm.go()
        rospy.sleep(2)
        self.request_face_position()

    def request_face_position(self):
        rospy.loginfo("Requesting face position")
        rospy.wait_for_service('get_face_position')
        try:
            get_face_position =
rospy.ServiceProxy('get_face_position', FacePosition)
            response = get_face_position()
            if response.success:
                rospy.loginfo("Position received.")
                self.process_face_position(response.position)
            else:
                rospy.loginfo("No face position received.")
                if self.face_detected:
                    rospy.loginfo("Continuing to monitor for face
positions.")
                    self.request_face_position()
                if not self.face_detected:
                    self.adjust_and_retry()
        except rospy.ServiceException as e:
```

```
            rospy.logerr("Service call failed: %s", e)
            rospy.sleep(1)
            self.request_face_position()

    def adjust_and_retry(self):
        rospy.loginfo("Adjust and Retry")
        current_joint_values = self.arm.get_current_joint_values()
        joint1 = current_joint_values[0]
        adjustment = self.rotation_direction *
self.rotation_factor * (3.14159 / 180)  # Convert degrees to
radians

        joint1 += adjustment
        self.arm.set_joint_value_target([joint1] +
current_joint_values[1:])
        self.arm.go()
        self.rotation_factor += 10

        self.rotation_direction *= -1
        rospy.sleep(1.5)  # Reduced sleep time
        self.request_face_position()

    def process_face_position(self, position):

        rospy.loginfo(f"[x={position.point.x}]
[y={position.point.y}] [z={position.point.z}]")
        self.face_detected = True
        x_offset = position.point.x
        y_offset = -position.point.y
        distance = math.sqrt(x_offset**2 + y_offset**2)

        if distance <= self.radius_threshold:
            if self.stable_time_start is None:
                self.stable_time_start = time.time()
```

```
            elif time.time() - self.stable_time_start >=
self.stable_duration:
                self.move_to_target_position(position.point)
                return
        else:
            self.stable_time_start = None

        current_joint_values = self.arm.get_current_joint_values()
        x_sensitivity = 0.001
        y_sensitivity = 0.001
        new_joint1 = current_joint_values[0] + x_offset *
x_sensitivity
        new_joint2 = current_joint_values[1] + y_offset *
y_sensitivity
        self.arm.set_joint_value_target([new_joint1, new_joint2] +
current_joint_values[2:])
        self.arm.go()
        self.request_face_position()

    def move_to_target_position(self, face_point):
        #tf_buffer_1 = tf2_ros.Buffer()
        #tf_buffer_2 = tf2_ros.Buffer()

        #listener = tf2_ros.TransformListener(tf_buffer_2)
       # tf_buffer_1 = tf_buffer_2
        #self.arm.set_named_target('ready_to_pick')
        #self.arm.go()
        #self.arm.set_named_target('pick_1')
        #self.arm.go()
        #self.arm.set_named_target('pick_2')
        #self.arm.go()
        #self.arm.set_named_target('pick_3')
        #self.arm.go()
        #self.arm.set_named_target('ready_to_pick')
        #self.arm.go()
```

```
        rospy.loginfo("Moving")
        face_pose = PoseStamped()
        face_pose.header.frame_id = "Camera_Link"
        face_pose.header.stamp = rospy.Time.now()
        initial_y = face_point.z * 0.0080
face_pose.pose.position.x = -(face_point.x) * 0.005
        face_pose.pose.position.y = initial_y
        face_pose.pose.position.z = -(face_point.y+10) * 0.005
        q = quaternion_from_euler(-3.14159/2, 0, 0)
        face_pose.pose.orientation = Quaternion(*q)

        tf_buffer = tf2_ros.Buffer()
        listener = tf2_ros.TransformListener(tf_buffer)
        try:
            transform = tf_buffer.lookup_transform('base_link',
'Camera_Link', rospy.Time(0), rospy.Duration(1.0))
            face_pose_transformed =
tf2_geometry_msgs.do_transform_pose(face_pose, transform)

            self.arm.set_named_target('ready_to_pick')
            self.arm.go()
            self.arm.set_named_target('pick_1')
            self.arm.go()
            self.arm.set_named_target('pick_2')
            self.arm.go()
            self.arm.set_named_target('pick_3')
            self.arm.go()
            self.arm.set_named_target('ready_to_pick')
            self.arm.go()

            attempt = 0
            success = False
            while not success and attempt < 10:
new_pose = Pose()
```

```
                new_pose.position.x =
face_pose_transformed.pose.position.x
                new_pose.position.y =
face_pose_transformed.pose.position.y
                new_pose.position.z =
face_pose_transformed.pose.position.z
                new_pose.orientation =
face_pose_transformed.pose.orientation

                rospy.loginfo(f"Try: x={new_pose.position.x},
y={new_pose.position.y}, z={new_pose.position.z}")

                self.arm.set_goal_position_tolerance(0.1)
                self.arm.set_goal_orientation_tolerance(0.1)
                self.arm.set_pose_target(new_pose)

                plan = self.arm.plan()
                #print(plan)
                if isinstance(plan, tuple) and plan[0] == 1:
                    rospy.loginfo("Trajectory planning successful,
executing trajectory")
                    self.arm.execute(plan[1])
                    success = True
                    rospy.loginfo("End")
                elif isinstance(plan,
moveit_commander.RobotTrajectory):
                    rospy.loginfo("Trajectory planning successful,
executing trajectory")
                    self.arm.execute(plan)
                    success = True
                else:
                    rospy.logerr("Planning failed, adjusting and
retrying...")
                    face_pose_transformed.pose.position.y =
initial_y - (initial_y * 0.1 * (attempt + 1))
```

```
                    attempt += 1
        except (tf2_ros.LookupException,
tf2_ros.ConnectivityException, tf2_ros.ExtrapolationException) as
e:
            rospy.logerr("False: %s", e)
if __name__ == "__main__":
    try:
        MoveItFkDemo()
        rospy.spin()
    except rospy.ROSInterruptException:
        pass
    finally:
        moveit_commander.roscpp_shutdown()
```

# ПРИЛОЖЕНИЕ Й

## robobk_arduino.ino

```
#include <ros.h>
#include <robobk_moveit/ArmJointState.h>
#include <ros/time.h>
#include <std_msgs/Int16MultiArray.h>
#include <AccelStepper.h>
#include <Stepper.h>
const int enablePin = 8;
const int xdirPin = 5;
const int xstepPin = 2;
const int ydirPin = 6;
const int ystepPin = 3;
const int zdirPin = 7;
const int zstepPin = 4;
const int STEPS_PER_ROTOR_REV = 32;
const int GEAR_REDUCTION = 64;
const float STEPS_PER_OUT_REV = STEPS_PER_ROTOR_REV *
GEAR_REDUCTION;

int last_command_id = 0;

int last_steps[4] = {0, 0, 0, 0};

int step_accumulator[4] = {0, 0, 0, 0};
int step_threshold = 10;
AccelStepper stepper1(1, xstepPin, xdirPin);
AccelStepper stepper2(1, ystepPin, ydirPin);
AccelStepper stepper3(1, zstepPin, zdirPin);
AccelStepper stepper4(AccelStepper::FULL4WIRE, 12, 10, 9, 11);
//Stepper stepper4(STEPS_PER_ROTOR_REV,12, 10, 9, 11);
ros::NodeHandle nh;
```

```
void jointStepCallback(const robobk_moveit::ArmJointState&
joint_steps_msg)
{
    if (joint_steps_msg.command_id != last_command_id)
    {
        last_command_id = joint_steps_msg.command_id;
        stepper1.move(joint_steps_msg.position1);
        stepper2.move(joint_steps_msg.position2);
        stepper3.move(joint_steps_msg.position3);
        stepper4.move(-joint_steps_msg.position4);
        //stepper4.step(-joint_steps_msg.position4);

        Serial.print("Commands executed: X=");
        Serial.print(joint_steps_msg.position1);
        Serial.print(", Y=");
        Serial.print(joint_steps_msg.position2);
        Serial.print(", Z=");
        Serial.print(joint_steps_msg.position3);
        Serial.print(", S=");
        Serial.println(joint_steps_msg.position4);
    }
}

ros::Subscriber<robobk_moveit::ArmJointState> sub("joint_steps",
jointStepCallback);

void setup()
{
  Serial.begin(57600);
  nh.initNode();
  nh.subscribe(sub);

  pinMode(enablePin, OUTPUT);
  digitalWrite(enablePin, LOW);
```

```
  stepper1.setMaxSpeed(800.0);
  stepper1.setAcceleration(1000.0);
  stepper2.setMaxSpeed(800.0);
  stepper2.setAcceleration(1000.0);
  stepper3.setMaxSpeed(800.0);
  stepper3.setAcceleration(1000.0);
  stepper4.setMaxSpeed(380.0);
  stepper4.setAcceleration(1000.0);
}

void loop() {
  nh.spinOnce();
  stepper1.run();
  stepper2.run();
  stepper3.run();
  stepper4.run();
}
```